\documentclass{article}

\usepackage[eandd,final]{neurips_2026}

\usepackage[utf8]{inputenc} 
\usepackage[T1]{fontenc}    
\usepackage{hyperref}       
\usepackage{url}            
\usepackage{booktabs}
\usepackage{array}       
\usepackage{amsfonts}       
\usepackage{nicefrac}       
\usepackage{microtype}      
\usepackage{xcolor}         
\usepackage{amsmath}
\usepackage{wrapfig}
\usepackage{float}
\usepackage{afterpage}
\usepackage{graphicx}
\usepackage{caption}
\usepackage{float}
\title{CryoAtlas: A Large Curated Dataset and Unified Benchmark for Cryo-EM Atomic Model Building}

\author{%
  \begin{tabular}{@{}c@{}}
  \bfseries Mingrui Li$^{1}$ \quad Minzhang Li$^{1,2}$ \quad Weichen Qin$^{1}$ \quad Yufan Xie$^{1}$ \quad Zhangzhi Xiong$^{1}$ \\[3pt]
  \bfseries Sixian Shen$^{1}$ \quad Jiakai Zhang$^{1,2*}$ \quad Yuan Pei$^{1*}$ \quad Jingyi Yu$^{1*}$
  \end{tabular}\\[3pt]
  {\normalfont $^{1}$ShanghaiTech University \qquad $^{2}$Cellverse Co., Ltd.}\\[2pt]
  {\normalfont\small $^{*}$Co-corresponding authors: \texttt{\{zhangjk,peiyuan,yujingyi\}@shanghaitech.edu.cn}}
}

\begin{document}

\maketitle

\begin{abstract}
Deep learning has accelerated automated atomic model building from cryo-EM maps, yet progress remains limited by the lack of large, high-quality datasets and standardized evaluation benchmarks. We address this gap by introducing CryoAtlas, a rigorously curated dataset and unified benchmark for density-guided atomic model building of protein chains. By integrating and filtering EMDB and PDB entries (up to March 2026), CryoAtlas provides 17,809 map--model--sequence entries, including a curated 16,657-entry training split. From this resource, we construct a 618-target benchmark with a shared evaluation protocol that jointly assesses structural accuracy, model-to-map fit, and stereochemical validity. Using this framework, we benchmark state-of-the-art learning-based methods against Phenix and AlphaFold~3, comprehensively analyzing the impact of map resolution, sequence length, protein category and sequence identity on modeling accuracy. As a case study, retraining the ModelAngelo CNN component on CryoAtlas improves performance on the benchmark. CryoAtlas provides an open, reproducible foundation for future cryo-EM atomic model building research. Project details and resource release updates are available at \url{https://oulling.github.io/cryoatlas/}.

\end{abstract}
\section{Introduction}

\afterpage{%
\begin{figure}[H]
    \centering
    \includegraphics[width=1.0\linewidth]{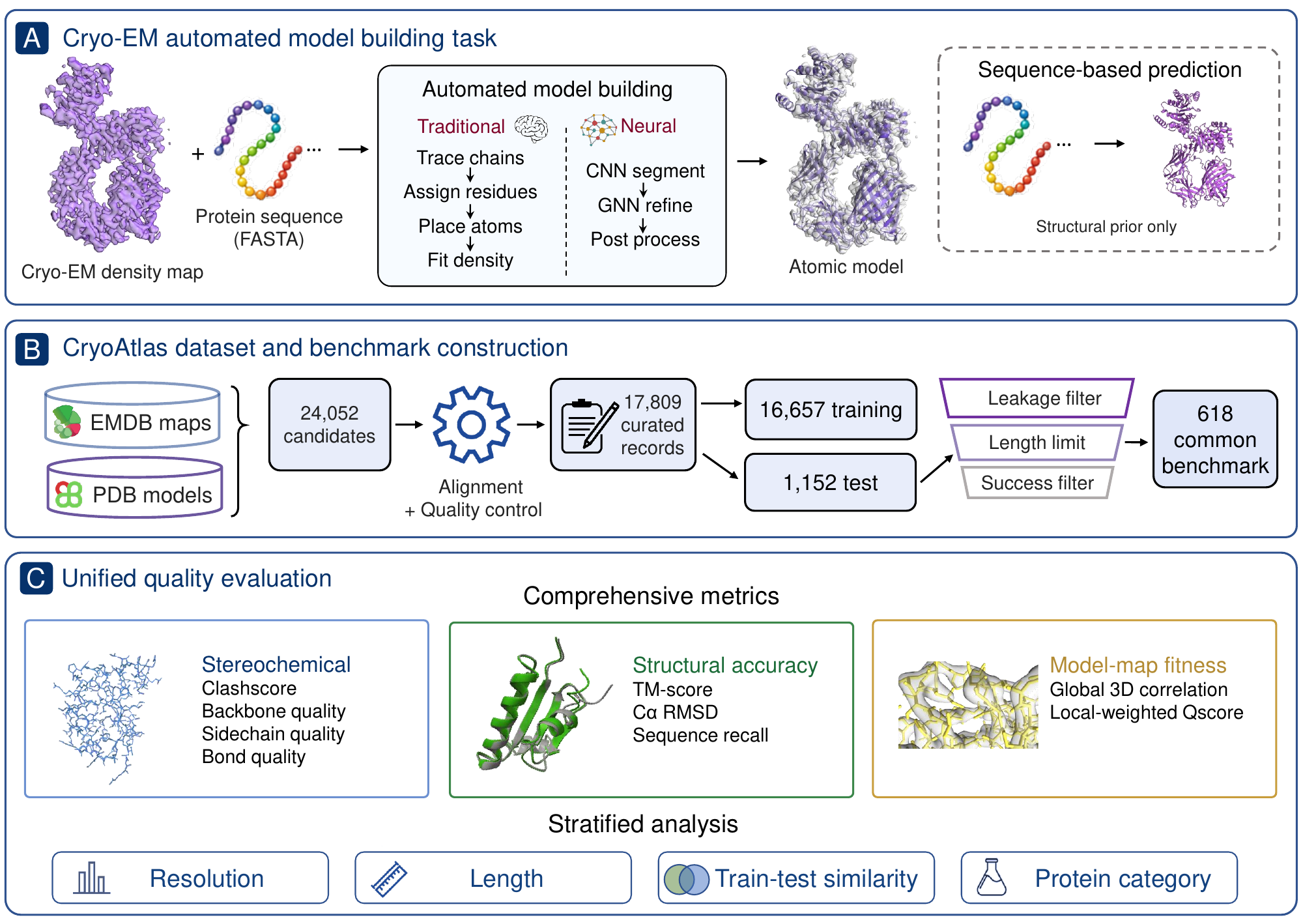}
    \caption{
    \textbf{Overview of CryoAtlas.} \textbf{A.} Task definition: cryo-EM model building uses density maps and protein sequences to produce a map-aligned atomic model, whereas sequence-based prediction uses sequence alone.
    \textbf{B.} Dataset curation workflow, resulting in 16,657 training targets and 1,152 test targets; methods are benchmarked on a shared 618-target subset. 
    \textbf{C.} CryoAtlas evaluates different methods using new composite metrics and performs stratified analysis across four factors.
    }
    \label{fig:dataset_curation}
\end{figure}
}

Cryo-electron microscopy (cryo-EM) has become a central modality for determining large protein complexes and conformational states that are difficult to access by X-ray crystallography~\cite{cryo_revolution1}. From a reconstructed density map, atomic model building must trace polypeptide chains, assign residue identities, and place atoms consistently with both the input sequence and the experimental density~\cite{novo_determine}; the resulting atomic model enables downstream tasks such as mechanistic interpretation, structure-based drug design, and protein engineering. Traditional pipelines combining manual building in interactive tools such as Coot~\cite{Coot} with iterative refinement in Phenix~\cite{Phenix} remain widely used in practice, while recent learning-based methods such as ModelAngelo~\cite{ModelAngelo_2024}, CryoAtom2~\cite{CryoAtom,CryoAtom2}, EMProt~\cite{EMProt}, and E3-CryoFold~\cite{E3-Cryofold} aim to automate atomic model building from cryo-EM maps. CryoBoltz~\cite{CryoBoltz}, EmbedOpt~\cite{EmbedOpt_2026}, and CryoACE~\cite{CryoACE_2026} further combine AF3-like predictors with cryo-EM maps through inference-time guidance or learned density conditioning. 



Despite recent progress, automated cryo-EM model building still lacks a unified evaluation framework~\cite{zhang2025comprehensive}. Existing evaluations rely on small, disjoint test sets with narrow resolution and protein-class coverage, and use inconsistent metric implementations, preventing systematic cross-method comparison~\cite{ModelAngelo_2024,Cryo2StructData_2024}. 
Such fragmented evaluation makes it difficult to attribute reported gains, since any apparent improvement may reflect differences in training data, target distribution, or evaluation protocol rather than the modeling method itself.
The literature consequently provides limited guidance on a fundamental practical question: which method is most reliable for a given cryo-EM map and modeling scenario?

We introduce CryoAtlas (Fig. ~\ref{fig:dataset_curation}), a large-scale curated dataset and unified benchmark for cryo-EM atomic model building. Specifically, the dataset is obtained through a rigorous curation pipeline that integrates EMDB~\cite{EMDB_2024} and PDB~\cite{PDB_2019} entries up to March 4, 2026 with strict cross-modal alignment, Q-score-based quality control, and removal of densities with large unmodeled regions, yielding 17{,}809 map--model--sequence entries with coordinate files, FASTA sequences, and density maps. From this resource, we construct a 618-target benchmark under a shared evaluation protocol designed around a central principle: cryo-EM model quality cannot be reduced to a single global similarity score against the deposited reference, but must combine multiple complementary aspects of model quality with stratification across target conditions. Our benchmark therefore separately scores structural accuracy, fit to the experimental density map, and stereochemical validity, while recording per-target sequence identity, map resolution, protein category, and sequence length to support stratified analysis.

Using CryoAtlas, we evaluate state-of-the-art learning-based methods including ModelAngelo, CryoAtom2, EMProt, and E3-CryoFold. Additionally, we include Phenix as a  conditional traditional reference, and AlphaFold~3~\cite{AF3} as a sequence-based reference to evaluate the role of density maps as inputs. Our evaluation yields three main findings. First, map resolution emerges as the dominant performance factor; in contrast, sequence length has a real but minor effect on the structural accuracy of density-guided methods, while AlphaFold~3 performance degrades sharply on long sequences. Second, AlphaFold~3 produces higher-quality stereochemistry, yet its predicted structures do not align tightly with the experimental density map and reference structure. Third, score distributions overlap across train--test sequence-identity groups, without a consistent monotonic trend across methods. Together, these findings show that no single axis is sufficient to characterize a method, which is why CryoAtlas reports structural accuracy, map agreement, and stereochemical validity jointly.



For large-scale training, CryoAtlas also provides a curated 16{,}657-entry training split for learning-based model building research. As a case study, we retrain the ModelAngelo CNN component on this split and observe improved performance on the CryoAtlas benchmark. Together, these results establish CryoAtlas as both a reproducible benchmark for cryo-EM atomic model building and a practical training resource for future learning-based method development.


\section{Background and Related Work}

\subsection{Traditional Atomic Modeling from Cryo-EM Density Maps}


Cryo-electron microscopy (cryo-EM) has become a central approach for resolving large macromolecular assemblies~\cite{cryo_development,cryo_revolution2}. Public archives such as EMDB~\cite{EMDB_2024} and PDB~\cite{PDB_2019} provide paired experimental density maps and deposited atomic coordinates. After a density map is reconstructed, however, atomic model building remains a difficult challenge: models must be placed in the map frame, traced through local density, assigned residues from sequence, and constrained by stereochemistry~\cite{rosettaes,MAINMAST}. Conventional tools~\cite{Buccaneer_2006,Pathwalking_2016,novo_determine} such as Phenix~\cite{Phenix}, together with visualization environments such as ChimeraX~\cite{Chimerax_2021}, remain widely used, but full automation is hindered by noise, local resolution variation, missing or ambiguous density, and manual correction, especially for heterogeneous complexes and lower-resolution regions~\cite{automated_ros}. Thus, cryo-EM automated model building differs from protein structure prediction: the goal is not merely a plausible fold, but an atomic model that explains a specific experimental density map.

\subsection{Sequence-based Prediction and Automated Model Building}
\label{subsec:sequence_and_model_building}

Sequence-based structure prediction~\cite{Rosettafold1,Rosettafold2,Boltz_1,Openfold,ESM2,ESM3,Seedfold} learns structural priors from large collections of experimentally determined structures. AlphaFold~2~\cite{AF2} and AlphaFold~3~\cite{AF3} use sequences, multiple-sequence alignments, templates, and biomolecular context to predict atomic structures. In their sequence-based setting, these models do not use the cryo-EM map, so their predictions need not match its coordinate frame, local features, or observed conformational state~\cite{terwilliger2024alphafold}. They can nevertheless provide starting models and structural priors for reconstruction from density~\cite{CryoBoltz}.

Density-driven automodeling~\cite{si2020deep,zhang2022cr,giri2024novo,Secondary_detection,wang2023cryoread} instead learns to build structures directly from experimental maps. DeepTracer~\cite{Deeptracer_2021} infers backbone traces from density, while DeepMainmast~\cite{terashi2024deepmainmast}, ModelAngelo~\cite{ModelAngelo_2024}, CryoAtom2~\cite{CryoAtom2}, EMProt~\cite{EMProt}, and E3-CryoFold~\cite{E3-Cryofold} combine density-derived features with sequence information and structural constraints. These methods must assign residues and maintain chain connectivity despite incomplete or poorly resolved density, making the integration of pretrained structural information particularly useful in ambiguous regions.

Density-guided structure prediction methods combine pretrained structure predictors with experimental density. ROCKET~\cite{ROCKET_2026} optimizes OpenFold MSA profiles against experimental likelihoods, while MICA~\cite{MICA_2025} fuses AF3 structures with maps for backbone modeling and refinement. Experiment-guided AF3~\cite{GuidedAF3_2026} and CryoBoltz~\cite{CryoBoltz} guide diffusion sampling with experimental constraints; EmbedOpt~\cite{EmbedOpt_2026} instead optimizes conditional embeddings. CryoACE~\cite{CryoACE_2026} learns atom-level density features within a Boltz-based diffusion model, with local-resolution-guided inference. Together, these methods integrate learned structural priors with experimental density.

\subsection{Open Resources for Learning-Based Automated Model Building}

Learning-based automated model building requires curated multimodal entries. Although EMDB and PDB provide public archives, building ML-ready datasets requires careful standardization of maps, map--structure alignment, sequence--structure consistency checks, and leakage-controlled train--test splits. These choices directly affect both training stability and benchmark reliability.
 
ModelAngelo~\cite{ModelAngelo_2024} and CryoAtom2~\cite{CryoAtom,CryoAtom2} describe curated training resources without releasing their complete training corpora. Cryo2StructData~\cite{Cryo2StructData_2024} releases processed maps, structures, and sequences, while CryoDataBot~\cite{CryoDataBot_2025} provides customizable dataset generation with map--model quality control. CryoAtlas combines explicit alignment and filtering rules with a fixed corpus and common benchmark; Appendix~\ref{app:curation_comparison} compares these workflows.

Evaluation protocols are also fragmented: existing benchmarks are small and non-unified, such as ModelAngelo's 177 targets and Cryo2StructData's 390 targets, both restricted to pre-March-2023 structures. Metrics are inconsistent across studies and often omit important aspects such as stereochemical quality. As a result, reported gains are difficult to attribute to modeling improvements rather than differences in data curation, test difficulty, or evaluation protocol.

\section{Dataset Construction and Curation}
\label{sec:dataset}




We curate a large, multimodal cryo-EM dataset that links EMDB density maps, PDB atomic models, and matched protein sequences for entries released up to March 4, 2026, with a $6\,\text{\AA}$ resolution cutoff to ensure interpretable structural signals~\cite{EMDB_2024,PDB_2019}. Maps better than $4\,\text{\AA}$ typically provide side-chain and backbone densities suitable for residue-level modeling~\cite{ModelAngelo_2024,Deeptracer_2021}, while $4$--$6\,\text{\AA}$ maps still retain traceable backbone trajectories despite unresolved side chains~\cite{Visualization_alpha_2006,Identification_2007}.

\subsection{Multi-Modal Alignment and Curation}

Starting from 24,052 EMDB entries, we applied a sequential filtering pipeline. We excluded maps without deposited structures and sequences containing any chain composed entirely of unknown residues. This yielded 21,229 map--structure--sequence triplets, which still contain three multi-modal issues: structure-map misalignment, sequence-structure mismatch, and experimental maps with large unmodeled regions.

We addressed these through a multi-stage curation workflow. We used UCSF ChimeraX's fitmap~\cite{Chimerax_2021} to align atomic structures to experimental maps, improving the average Q-score~\cite{Qscore_2020} (ranging from -1 to 1, reflecting local atom resolvability) by 0.103 across 989 entries requiring realignment. An example is shown in Appendix Fig.~\ref{fig:c2s_examples}. In parallel, Clustal Omega~\cite{Clustal_2014} was used to align full-length FASTA sequences to chains in the structure, marking unmodeled residues as gap tokens and additionally producing terminal-trimmed FASTA files that remove flexible N- and C-terminal tails. Quality control then removed 994 entries with aligned Q-score below 0.3 and further excluded severely incomplete models (Appendix Fig.~\ref{fig:c2s_examples}) using an unmodeled-density ratio computed from experimental and simulated density (Appendix~\ref{sec:appendix_unmodeled}). The pipeline yielded 17,809 refined triplets. Dataset composition and filtering effects are documented in Appendix~\ref{app:dataset_composition}.

\subsection{Train--Test Split and Benchmark Pool}

For train--test splitting, we clustered chains with MMseqs2~\cite{MMseqs2_2017} at 40\% sequence identity and grouped maps sharing identical chain-cluster signatures into 6,062 non-redundant complex-level clusters. Size-stratified sampling drew a 1,152-map held-out test pool by selecting one representative from 10\% of large clusters (size $>5$) and all maps from 10\% of small clusters (size $\leq 5$), leaving 16,657 maps for training. For the cross-method evaluations reported in this paper, we focus on a 618-target benchmark subset, detailed in Section~\ref{sec:benchmark-test-set}. The full 1,152-map pool is released as a broader candidate set for any model trained on our train--test split.

To characterize sequence-level novelty relative to our training split, we performed an all-versus-all MMseqs2 similarity analysis (query and target coverage $\geq50\%$, E-value $\leq10^{-3}$, sensitivity 7.5, and no minimum identity cutoff) between the 1,152 test pool and the 16,657 training samples. We summarize each map by the maximum observed identity across its qualifying alignments. Of 1,028 maps with a qualifying hit, 30, 89, 37, and 872 fall into the $<30\%$, $30$--$50\%$, $50$--$70\%$, and $\geq70\%$ bins, respectively, while the remaining 124 maps have no qualifying hit. The full 1,152 set measures performance under realistic data distributions, whereas the sequence-novel maps with observed identity $<30\%$ or no qualifying hit specifically assess generalization to low-homology targets.


\section{Evaluation Framework}
\label{sec:evaluate}

We benchmarked six representative methods across three paradigms: traditional density-guided tools (Phenix~\cite{Phenix}), automodeling methods (ModelAngelo~\cite{ModelAngelo_2024}, E3-CryoFold~\cite{E3-Cryofold}, CryoAtom2~\cite{CryoAtom2}, EMProt~\cite{EMProt}), and sequence-based prediction (AF3~\cite{AF3} via the official server). All models were run using default parameters (EMProt in de novo mode) without post-refinement to assess their intrinsic performance. Phenix is included as a conventional reference baseline. For AF3 and E3-CryoFold, the predicted structures were aligned to the experimental maps because their output coordinate frames differ from the ground truth. For AF3, we used the highest-confidence model among the five server predictions. Density-guided structure prediction methods were excluded because they were insufficiently robust on complex targets and prone to failure in our preliminary tests.

\subsection{Test Set Composition and Diversity}

\noindent\begin{minipage}[t]{0.55\textwidth}
\vspace{0pt}
Our test set comprises 1{,}152 EMDB-PDB pairs, substantially exceeding prior benchmarks such as ModelAngelo (177 cases) and Cryo2StructData (390 cases). Resolution distributions and dataset composition are reported in Appendix~\ref{app:dataset_composition}.

Functionally, targets are grouped into nine PDB-keyword classes (Table~\ref{tab:test_categories}). Rare multi-label cases follow fixed priority: membrane/transport first, viral second, then other function-specific labels. Membrane \& Transport is largest (35.9\%), followed by Enzymes (16.1\%) and Viral assemblies (14.1\%). Overall, the test set covers major cryo-EM modeling classes, making it diverse and representative of real-world targets.
\end{minipage}\hfill
\begin{minipage}[t]{0.42\textwidth}
\vspace{0pt}
  \centering
  \small
  \setlength{\tabcolsep}{3pt}
  \begin{tabular}{@{}lrr@{}}
    \toprule
    \textbf{Protein Category} & \textbf{N} & \textbf{Share} \\
    \midrule
    Membrane \& Transport    & 413   & 35.9\% \\
    Enzymes                  & 185   & 16.1\% \\
    Viral                    & 163   & 14.1\% \\
    Others                   & 115   & 10.0\% \\
    Nucleic Acid Binding     & 75    & 6.5\%  \\
    Signaling \& Receptor    & 61    & 5.3\%  \\
    Structural \& Motor      & 61    & 5.3\%  \\
    Immune System            & 56    & 4.9\%  \\
    Photosynthesis           & 23    & 2.0\%  \\
    \midrule
    \textbf{Total}           & \textbf{1{,}152} & \textbf{100\%} \\
    \bottomrule
  \end{tabular}
  \captionof{table}{\textbf{Protein category of the test set.}}
  \label{tab:test_categories}
\end{minipage}
\par\medskip

\subsection{Benchmark Test Set Construction}\label{sec:benchmark-test-set}

All test inputs use raw EMDB~\cite{EMDB_2024} and PDB~\cite{PDB_2019} data. To avoid leakage for density-guided baselines, we excluded entries overlapping with their training sets, yielding 721 eligible targets. AF3 was evaluated on targets within the 5,000-residue limit of its official server. The final benchmark contains the 618 targets successfully modeled by all five learning-based methods. We additionally evaluate Phenix on the same 618 targets. Only learning-based automodeling methods are ranked; AF3 and Phenix serve as references.

\subsection{Metrics}
We evaluate stereochemical quality, structural accuracy, and model-to-map fitness using both individual metrics and dimension-specific composite scores. The composites provide an overview; the raw metrics identify the strengths and failure modes behind each score.



\paragraph{Stereochemical Quality}

Stereochemical quality measures the physical plausibility of predicted atomic structures. Using phenix.molprobity~\cite{Phenix}, we extract seven raw geometry indicators together with the official MolProbity score~\cite{Molprobity}. Since the MolProbity composite uses only three indicators and was calibrated mainly on X-ray reference structures, it is less directly suited to our cryo-EM benchmark. We therefore define Geom Score $\in (0,1)$, a bounded, benchmark-standardized composite that integrates all seven raw indicators.

Each indicator is mapped to a standardized value $q_i \in (0,1)$. The seven values are grouped into four sub-categories: steric, backbone, sidechain, and bonded geometry. We aggregate them by geometric mean within each group and arithmetic mean across groups:
\[
S_g = \left( \prod_{k \in g} q_k \right)^{1/|g|}, 
\qquad 
\text{Geom Score} = \frac{1}{|\mathcal{G}_{\mathrm{valid}}|} \sum_{g \in \mathcal{G}_{\mathrm{valid}}} S_g .
\]
Here $\mathcal{G}_{\mathrm{valid}}$ contains groups with all required indicators available for the evaluated structure. Equal weighting of available groups avoids giving more influence to groups with more indicators. Standardization and robustness analyses are provided in Appendix~\ref{subsec:appendix_stereo}; agreement with MolProbity gives Spearman $|\rho|>0.87$ across methods.

\paragraph{Structural Accuracy}
Structural accuracy measures agreement with the deposited reference. We compute TM-score with USalign~\cite{USalign}, and local C$\alpha$ RMSD, residue coverage, sequence match, and sequence recall with phenix.chain\_comparison~\cite{Phenix}. These metrics capture global topology, coordinate accuracy on aligned residues, and recovery of correctly assigned reference residues, respectively.



Since individual metrics introduce different biases, especially for partial-coverage models, we define GT Score $\in (0,1]$ as a balanced composite of topology, geometric accuracy, and sequence-level recovery. In particular, TM-score can penalize incomplete models, whereas C$\alpha$ RMSD may favor models that build only well-resolved regions~\cite{EMProt}; sequence-only measures, meanwhile, lack 3D geometric context. We therefore combine the three complementary terms as
\[
\text{GT Score}
= 0.15 \cdot \text{TM-score}
+ 0.45 \cdot \exp\left(-\text{C}_{\alpha}\text{ RMSD}/3\right)
+ 0.40 \cdot \text{sequence\_recall}.
\]
The fixed weights prioritize local placement and sequence recovery; component selection and weighting are detailed in Appendix~\ref{subsubsec:gt_formulation}.

\paragraph{Model-to-Map Fit}

Model-to-map fit measures how well predicted atomic coordinates are supported by the corresponding cryo-EM density. We evaluate this dimension at two scales: four real-space correlation metrics, CC\_mask, CC\_volume, CC\_peaks, and CC\_box, computed with phenix.real\_space\_correlation~\cite{Phenix}, assess global density agreement, while Q-score~\cite{Qscore_2020} measures local atomic resolvability.


We define Map Score $\in [-1,1]$ by combining global and local density agreement. The global term, cc\_mean, averages the four CC variants. For the local term, raw Q-score can be inflated by partial models that build only well-resolved regions; we therefore use recall\_qscore, defined as $\text{sequence\_recall} \cdot \text{Q-score}$, to reduce the local contribution of incomplete reconstructions when Q-score is nonnegative. For negative Q-scores, recall weighting moves the local term toward zero, so it is not an incompleteness penalty in that regime. Negative values are retained. The final score weights equally:
\[
\text{Map Score}
= 0.5 \cdot \underbrace{\text{cc\_mean}}_{\text{global}}
+ 0.5 \cdot \underbrace{\text{recall\_qscore}}_{\substack{\text{\scriptsize coverage-} \\ \text{\scriptsize weighted local}}}
\]
\vspace{-1cm}

\section{Results}

\begin{figure}[t]
    \centering
    \vspace{-0.3cm}
    \includegraphics[width=1.0\linewidth]{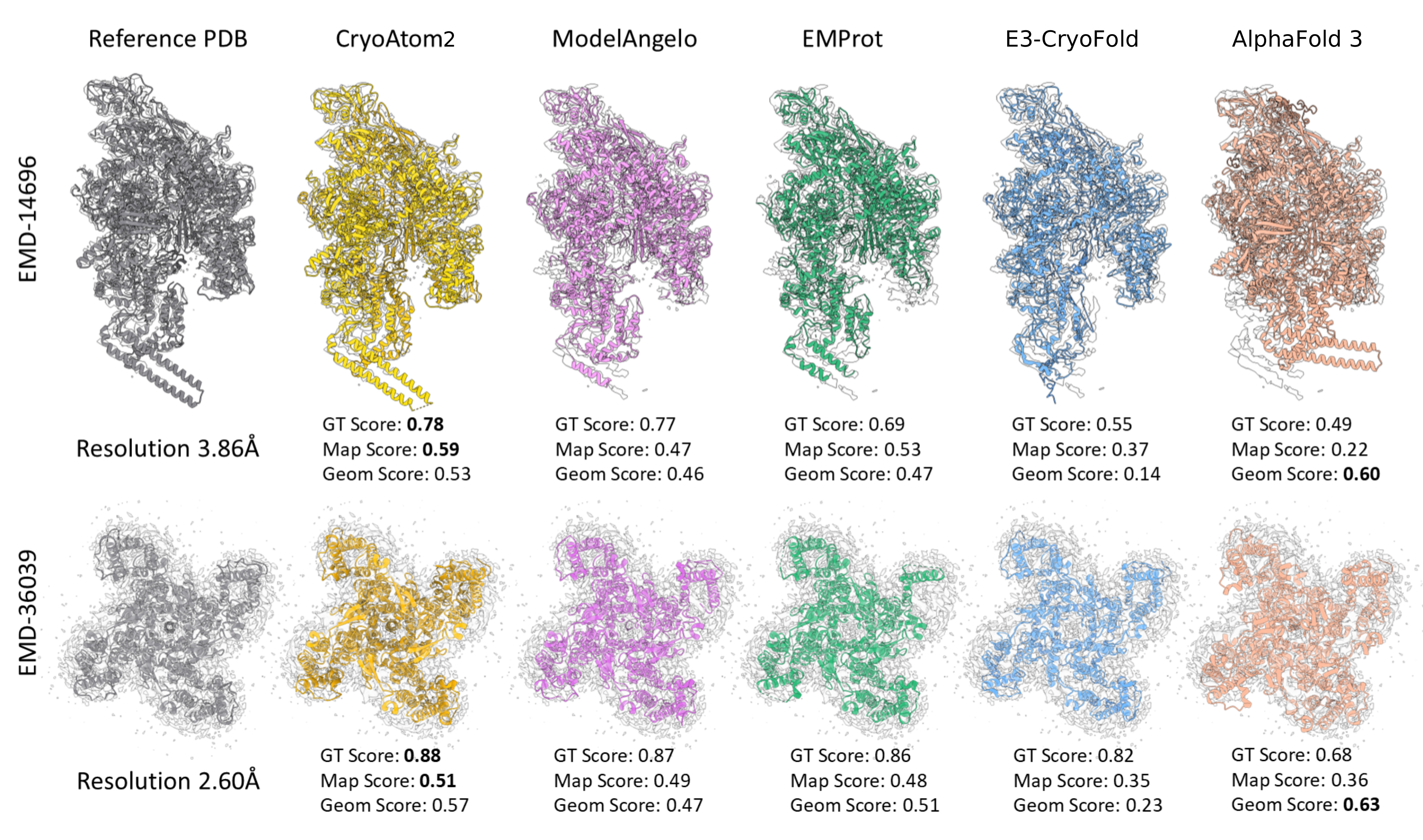}
    \caption{
    \textbf{Visual comparison of different methods.} Results are shown on EMD-14696 (length 3640; Mtb RNA polymerase--$\sigma$B protomer)~\cite{morichaud2023structural} and EMD-36039 (length 1176; human ion channel)~\cite{chen2024structural}.
    }
    \vspace{-0.5cm}
    \label{fig:Exhibition}
\end{figure}

\subsection{Completion Rates}

We define the completion rate as a method's ability to produce a valid structural output under our evaluation protocol (Table~\ref{tab:success_rate}).

\noindent
\begin{minipage}[t]{0.52\textwidth}
\vspace{0pt}
For density-guided learning-based methods, completion requires a modeled structure with completeness $>0.05$, defined as the fraction of modeled residues relative to the reference length. For AlphaFold~3, completion requires a valid server prediction with a positive ranking score (see Appendix \ref{sec:AF3_failure} for analysis of failure cases). For Phenix, completion requires phenix.trace\_and\_build~\cite{Phenix} to generate a model within the 24-hour budget.
\end{minipage}\hfill
\begin{minipage}[t]{0.45\textwidth}
  \vspace{0pt}
  \centering
  \small
  \setlength{\tabcolsep}{3pt}
  \begin{tabular}{lcc}
    \toprule
    \textbf{Method} & \textbf{Rate} & \textbf{Completed / Eligible} \\
    \midrule
    ModelAngelo & 0.9695 & 699 / 721 \\
    CryoAtom2 & \textbf{0.9972} & 719 / 721 \\
    E3-CryoFold & 0.9390 & 677 / 721 \\
    EMProt & 0.9723 & 701 / 721 \\
    AlphaFold~3 & 0.9873 & 935 / 947 \\
    Phenix (24h) & 0.5922 & 366 / 618 \\
    \bottomrule
  \end{tabular}
\captionof{table}{\textbf{Completion rate for each method.}}
  \label{tab:success_rate}
\end{minipage}
\par\medskip

Completion rate reflects algorithmic robustness and practical reliability across diverse real-world inputs. This is especially relevant for CPU-based traditional pipelines such as Phenix, whose runtime can be prohibitive at benchmark scale. Subsequent analyses use the 618-target common subset with valid outputs from all learning-based methods.

\subsection{Benchmark Results}
Table~\ref{tab:main_results} summarizes the overall performance of the evaluated methods across 618 targets. Figure~\ref{fig:Exhibition} provides two representative visual comparisons of the models predicted by different methods. To provide a comprehensive assessment, we analyze the results from the following three perspectives.

\textbf{Stereochemical Quality.}
AlphaFold~3 achieves the highest Geom Score, reflecting strong structural priors that produce idealized protein geometry. Among density-guided methods, CryoAtom2 performs best, while Phenix reaches a comparable geometry on its completed subset, consistent with its mature stereochemical and density-fitting restraints. Standard refinement can further improve the geometry of most automodeling outputs to AF3-level quality~\cite{EMProt}. Details are provided in Appendix~\ref{subsubsec:stereo_analysis}.

\begin{table}[!b]
\centering
\small
\setlength{\tabcolsep}{2pt}
\resizebox{\textwidth}{!}{%
\begin{tabular}{lccccccc}
\toprule
\textbf{Method} & \textbf{Geom Score ($\uparrow$)}  & \textbf{C$\alpha$ RMSD ($\downarrow$)} & \textbf{TM-score ($\uparrow$)} & \textbf{Seq. Recall ($\uparrow$)} & \textbf{GT Score ($\uparrow$)} & \textbf{Q-score ($\uparrow$)} & \textbf{Map Score ($\uparrow$)} \\
\midrule
Phenix (24h)$^\dagger$ & 0.462 & 3.562 & 0.075 & 0.063 & 0.194 & 0.539 & 0.193 \\
ModelAngelo & 0.462 & \textbf{1.014} & 0.648 & 0.739 & 0.723 & \textbf{0.600} & 0.520 \\
CryoAtom2 & 0.537 & 1.198 & 0.763 & \textbf{0.899} & \textbf{0.787} & 0.585 & \textbf{0.568} \\
EMProt & 0.484  & 1.224 & 0.639 & 0.734 & 0.702 & 0.588 & 0.517 \\
E3-CryoFold & 0.165  & 3.278 & 0.739 & 0.622 & 0.539 & 0.332 & 0.305 \\
\midrule
AlphaFold~3 & \textbf{0.598} & 3.655 & \textbf{0.813} & 0.647 & 0.552 & 0.202 & 0.220 \\
\bottomrule
\end{tabular}
}
\vspace{1mm}
\caption{\textbf{Main performance evaluation.} All methods except Phenix are evaluated on the 618-target common benchmark; means use available metric values. $^\dagger$Phenix is averaged over the 366 targets completed within 24 hours. Target-level sample SDs are reported in Appendix Table~\ref{tab:main_results_sd}. Additional detailed results are provided in Appendix Tables~\ref{tab:supp_stereo}, \ref{tab:supp_stereo_subscores}, \ref{tab:supp_gt}, and \ref{tab:supp_map}. RMSD values are in \AA{}.}
\label{tab:main_results}
\end{table}

\textbf{Structural Accuracy.}
While low RMSD indicates high positional accuracy on aligned fragments, chain breaks and incomplete structures degrade TM-scores in density-guided methods. Thus, sequence recall is necessary to capture the true alignable proportion. Integrating these metrics, GT Score confirms that CryoAtom2 achieves the highest accuracy among the density-guided methods. Phenix attains a lower GT Score despite plausible local geometry, mainly because it often reconstructs only limited portions of the reference structure. Details are provided in Appendix~\ref{subsubsec:gt_analysis}.

\textbf{Model-to-Map Fitness.}
Q-score only evaluates successfully modeled atoms. ModelAngelo achieves the highest Q-score by modeling regions with high density. However, CryoAtom2 covers substantially more residues, resulting in the highest overall Map Score. Phenix obtains a moderate Q-score, but its low sequence recall drives a much lower Map Score. Details are provided in Appendix~\ref{subsec:Model-Map fitness}.

\textbf{Summary.}
Overall, CryoAtom2 performs best among the density-guided methods, with ModelAngelo and EMProt comparable and E3-CryoFold substantially lower. Its advantage likely reflects architectural extensions to ModelAngelo and a larger similarly curated corpus, whereas EMProt remains competitive with less training data. E3-CryoFold shows that data scale alone is insufficient without high-fidelity data and robust model design. AlphaFold~3 has stronger geometry but lower agreement with density-constrained ground truth, and Phenix reconstructs fewer residues with lower structural accuracy. Details are provided in Appendix~\ref{sec:appendix_metrics}.

\subsection{Stratified Performance Analysis}
We stratify the 618 test maps by map resolution, sequence length, protein category, and sequence identity to each model's training set. Phenix is included only for targets completed within the 24-hour wall-clock budget.

\noindent
\begin{minipage}[t]{0.43\textwidth}
  \vspace{0pt}
  \centering
  \includegraphics[width=\linewidth]{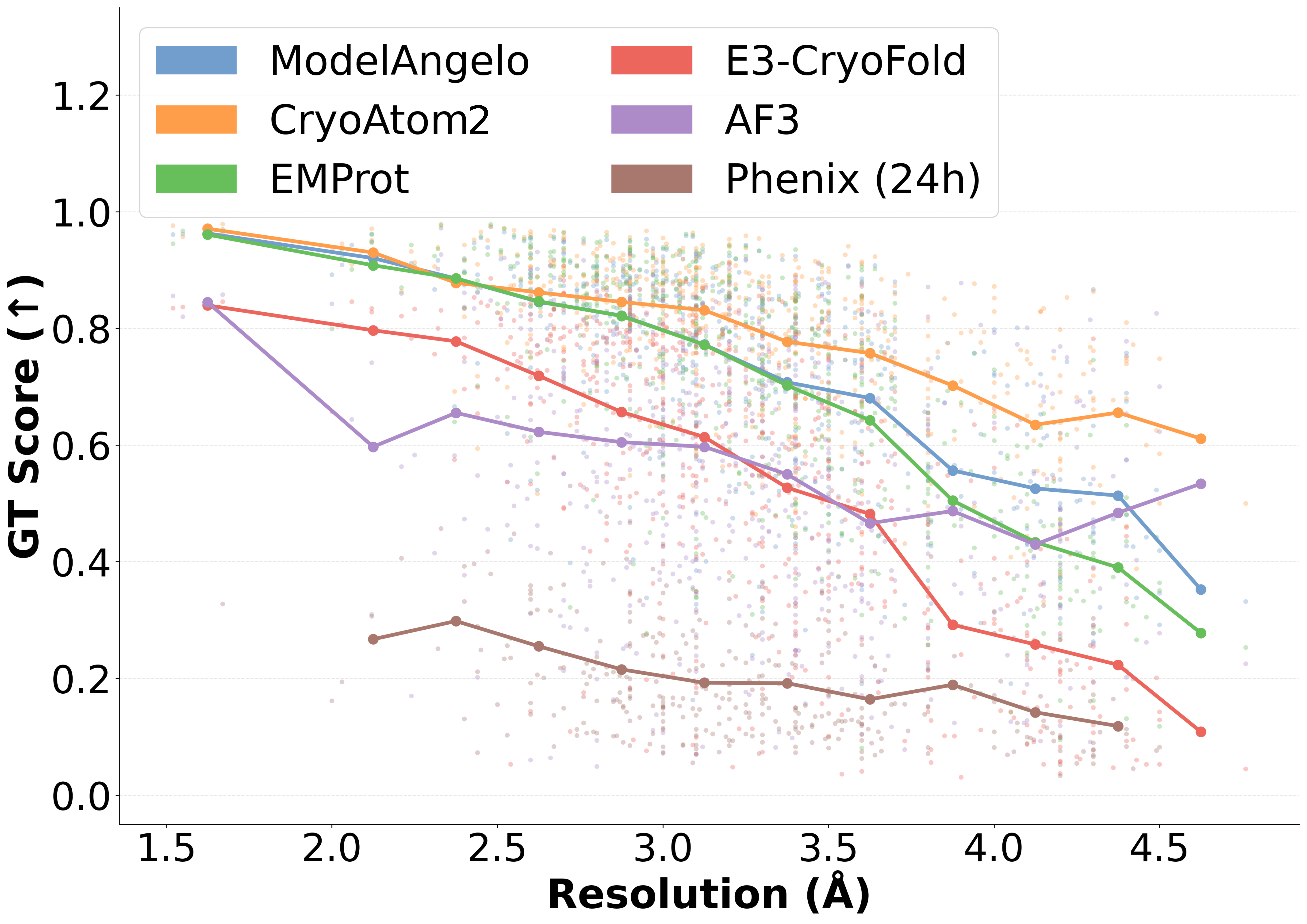}
  \captionof{figure}{GT Score by map resolution; companion metrics are in Appendix~\ref{app:Res}.}
  \label{fig:Res_GT}
\end{minipage}\hfill
\begin{minipage}[t]{0.545\textwidth}
\vspace{0pt}
\paragraph{Resolution-dependent Performance}
GT Score declines with decreasing map resolution across all models (Fig.~\ref{fig:Res_GT}). Map Score follows the same trend, whereas Geom Score remains largely stable (Appendix Fig.~\ref{fig:res_map_geom}). This indicates that resolution mainly affects structural accuracy and density fitting rather than stereochemical validity. Among density-guided methods, CryoAtom2 is the most robust to resolution degradation: its performance is similar to EMProt and ModelAngelo at resolutions better than $3\,\text{\AA}$, but its advantage increases at lower resolutions, suggesting stronger extraction of useful signal from sparse and noisy density.

\end{minipage}
\par\smallskip

AF3 shows a similar decline despite not using density maps, which should be interpreted as dataset difficulty rather than direct resolution dependence: lower-resolution entries tend to be larger, more heterogeneous, or structurally more complex, making sequence-based prediction harder and increasing the discrepancy between the predictions and the ground truth.

\paragraph{Length-dependent Performance}

\begin{wrapfigure}[15]{R}{0.48\linewidth}
  \vspace{-4mm}
  \centering
  \includegraphics[width=\linewidth]{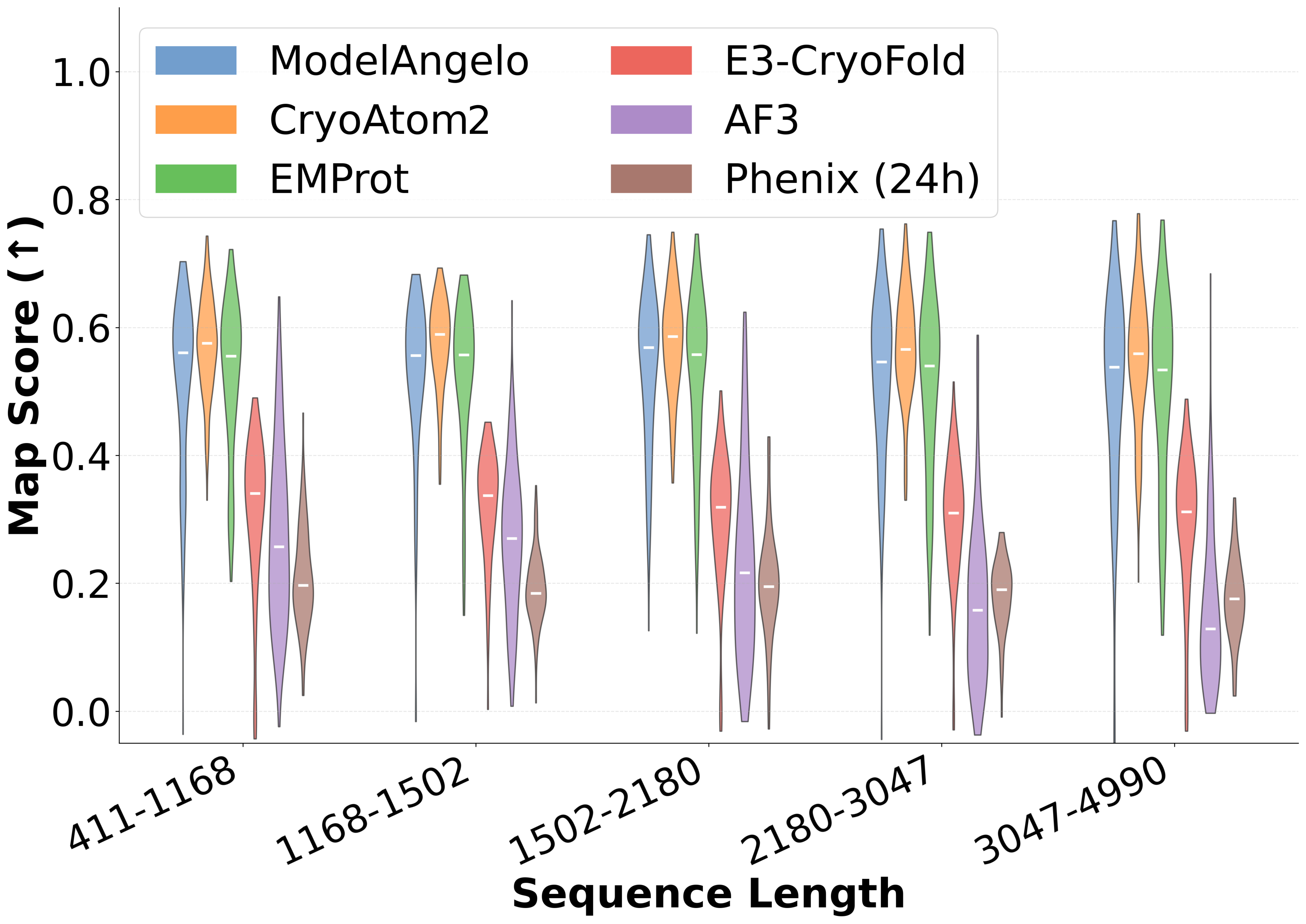}
  \caption{Map Score by sequence length. GT Score and Geom Score are in Appendix~\ref{app:length}.}
  \label{fig:Length_Map}
\end{wrapfigure}

For density-guided methods, Map Score remains stable across length bins (Fig.~\ref{fig:Length_Map}), GT Score decreases only mildly, and Geom Score is largely insensitive to length (Appendix~\ref{app:length}). This pattern is consistent with their architectures: CNN encoders extract local voxel features and GNN decoders propagate residue-neighborhood information, making predictions slightly dependent on total chain length. The mild GT Score decline reflects the modest length sensitivity of global structure-comparison metrics, where accumulated local deviations penalize longer chains.

AF3 shows a stronger length-dependent decline in both GT Score and Map Score. Without experimental density, AF3 must infer global topology and inter-domain organization from sequence-derived priors alone, allowing errors to accumulate in longer chains. These results suggest that long cryo-EM targets, often involving larger proteins, multi-domain organization, or complex assemblies, remain challenging for sequence-based prediction, consistent with prior reports that AlphaFold-family reliability varies with protein size and structural complexity.

\paragraph{Protein-category Effects}

We grouped the 618 benchmark targets into nine functional categories using PDB keywords; Appendix Fig.~\ref{fig:category_scores} shows the corresponding category-level results. The score distributions overlap across categories, and the plots do not show a consistent category-dependent ordering. Limited per-category sample sizes constrain this descriptive comparison; the figures alone do not establish the absence of category effects.

\paragraph{Test-Train Sequence Identity}

For each density-guided learning-based method, we used MMseqs2 to compute the maximum sequence identity between each benchmark chain and the method's training set, and grouped targets into three bins ($\geq70\%$, $30\%\leq\mathrm{identity}<70\%$, and $<30\%$; Appendix~\ref{app:identity}). The plotted score distributions overlap across identity bins, with no consistent monotonic trend shared by all four methods. AlphaFold~3 is excluded because its training distribution cannot be matched to our targets under the same protocol. These stratified comparisons characterize performance across sequence-similarity groups, but do not isolate sequence identity from map resolution or other target properties.

\section{Training Utility}

We evaluate the training utility of CryoAtlas through controlled retraining of ModelAngelo's CNN segmentation module, assessing both optimization stability and downstream atomic model-building accuracy. CryoAtlas and Cryo2StructData are compared under a shared training recipe, with additional preprocessing and filtering controls to assess the contribution of data curation. We reimplement CNN training from the published description and retain the official GNN checkpoint for inference, keeping the downstream refinement stage fixed across retrained variants. The released ModelAngelo model serves as an external reference.

\paragraph{Reimplementation.}
Our reimplementation follows the published ModelAngelo description. The architecture follows the published specification, while minor details such as loss scheduling may differ from the original recipe; full settings are provided in Appendix~\ref{app:reimpl-v21}.

\paragraph{Training setup.}
ModelAngelo reports training for $400{,}000$ steps with batch size~$16$, taking approximately $4.6$ days. Our reimplementation trains for $50{,}000$ steps with effective batch size~$96$ on $8\times$ NVIDIA H20 GPUs (96\,GB each), completing in approximately $39$ hours. We assess early training stability over the first $5{,}000$ steps and evaluate downstream performance using the final checkpoints.

\paragraph{Controlled experiments.}
For downstream evaluation, we train CNNs on CryoAtlas and raw maps obtained from the Cryo2StructData entry list using the same preprocessing and training recipe, then evaluate their $50{,}000$-step checkpoints on the 618-target benchmark. For early training dynamics, we run three independent CryoAtlas seeds for $5{,}000$ steps and compare four single-run controls: raw Cryo2StructData, its released normalized maps, CryoAtlas with discarded entries restored, and raw Cryo2StructData with our filters applied. These controls assess normalization and filtering effects under a shared training protocol; implementation details and stability measures are provided in Appendices~\ref{app:reimpl-v21} and~\ref{subsec:c2s_train_stability}.

\begin{table}[t]
\centering
\small
\setlength{\tabcolsep}{2pt}
\resizebox{\textwidth}{!}{%
\begin{tabular}{lccccccc}
\toprule
\textbf{Method} & \textbf{Geom Score}($\uparrow$)  & \textbf{C$\alpha$ RMSD ($\downarrow$)} & \textbf{TM-score ($\uparrow$)} & \textbf{Seq. Recall ($\uparrow$)} & \textbf{GT Score ($\uparrow$)} & \textbf{Q-score ($\uparrow$)} & \textbf{Map Score ($\uparrow$)} \\
\midrule
MA (Official) & 0.462 & 1.014 & 0.648 & 0.739 & 0.723 & 0.600 & 0.520 \\
MA (Cryo2Struct) & 0.455 & 1.108 & 0.604 & 0.708  & 0.697 & 0.599 & 0.505 \\
MA (CryoAtlas) & \textbf{0.470} & \textbf{0.769} & \textbf{0.680} & \textbf{0.753}  & \textbf{0.755} & \textbf{0.603} & \textbf{0.532} \\
\bottomrule
\end{tabular}
}
\vspace{3mm} 
\caption{
Performance of three ModelAngelo (MA) variants on the benchmark test set: the official checkpoint, MA trained on Cryo2StructData, and MA trained on CryoAtlas. Means are reported here; target-level sample SDs are provided in Appendix Table~\ref{tab:training_utility_sd}. Full metric breakdowns are provided in Appendix Tables~\ref{tab:supp_stereo_train}, \ref{tab:supp_geom_train}, \ref{tab:supp_gt_train}, and \ref{tab:supp_map_train}.
 RMSD values are in \AA{}.}
\label{tab:training_utility}
\end{table}

\begin{table}[t]
\centering
\small
\setlength{\tabcolsep}{4pt}
\renewcommand{\arraystretch}{1.15}
\begin{tabular*}{\textwidth}{@{\extracolsep{\fill}}lccc@{}}
\toprule
\textbf{Training data} & \textbf{EMA loss ($\downarrow$)} & $\boldsymbol{r_{\mathrm{RMS}}}$ ($\downarrow$) & $\boldsymbol{r_{\mathrm{MAD}}}$ ($\downarrow$) \\
\midrule
CryoAtlas (3 seeds) & $0.0898\pm0.0009$ & $0.2240\pm0.0074$ & $0.1853\pm0.0109$ \\
Cryo2StructData (raw) & 0.1127 & 0.2836 & 0.2069 \\
Cryo2StructData (normalized) & 0.1475 & 0.2299 & 0.1565 \\
CryoAtlas + discarded & 0.1140 & 0.2841 & 0.1956 \\
Cryo2StructData + filters & 0.0865 & 0.2261 & 0.1911 \\
\bottomrule
\end{tabular*}
\vspace{2mm}
\caption{Training dynamics over $5{,}000$ steps. CryoAtlas reports mean $\pm$ SD over three seeds; other rows are single runs. EMA uses decay 0.99. ``Normalized'' uses the released Cryo2StructData maps; ``+ filters'' applies CryoAtlas filtering to raw Cryo2StructData.}
\label{tab:rebuttal_training_stability}
\end{table}

\paragraph{Results.}
At $50{,}000$ steps, CryoAtlas training reduces C$\alpha$ RMSD from $1.108$ to $0.769$\,\AA{} and improves GT Score from $0.697$ to $0.755$ relative to Cryo2StructData under matched settings (Table~\ref{tab:training_utility}). It also slightly exceeds the released ModelAngelo checkpoint, which serves as an external reference because its original training data and code are unavailable. Across three seeds, CryoAtlas reaches an EMA loss of $0.0898\pm0.0009$ at $5{,}000$ steps. Restoring discarded entries raises the loss to $0.1140$, while filtering raw Cryo2StructData lowers it from $0.1127$ to $0.0865$ (Table~\ref{tab:rebuttal_training_stability}). The released normalized maps have lower volatility but a higher loss plateau than raw maps. Together, these controls support the effectiveness of our filters in improving training dynamics; differing sample counts limit causal comparisons between datasets.

\section{Discussion}
\paragraph{Conclusion.}

We curated a large-scale, rigorously cleaned multimodal dataset and established a standardized benchmark for cryo-EM density-guided atomic modeling, evaluating existing methods alongside AlphaFold~3 and the traditional Phenix pipeline. Beyond method comparison, the benchmark quantifies how map resolution, sequence length, protein category, and test--train sequence identity affect model accuracy. To validate the quality of our dataset, we trained the ModelAngelo CNN segmentation model on both the CryoAtlas and Cryo2StructData training sets, demonstrating that our curated dataset improves training stability and yields higher downstream task accuracy. This work provides a foundation for reproducible progress in cryo-EM automated model building.

\paragraph{Limitations and future work.}
Four directions remain open. First, our entries retain only protein chains; excluding nucleic acids prevents fully all-atom modeling, and extending curation to preserve them is a natural next step. Second, although we adopted comprehensive raw metrics and designed composite scores to reduce single-metric biases, these biases are not fully eliminated; developing more robust evaluation algorithms remains important. Third, current whole-structure scores do not capture class-specific functional regions, such as enzyme active sites, transmembrane interfaces, antibody CDRs, or nucleic-acid contact surfaces; region-aware evaluation remains an important direction for future benchmarks. Fourth, while our dataset contains many heterogeneous and dynamic assemblies, our analysis treats each density as a single static target; multi-state, ensemble-aware modeling and evaluation remain future work.

\paragraph{Data and Code Availability.}
\label{para:data_code_availability}
The dataset, benchmark resources, and documentation are publicly available on \href{https://huggingface.co/datasets/markfaa/CryoAtlas}{Hugging Face}, and code in \href{https://github.com/oulling/CryoAtlas-Code}{CryoAtlas-Code}. The dataset and original CryoAtlas code are released under CC BY-NC 4.0; third-party assets retain their original licenses and terms.

\section*{Impact Statement}
\phantomsection\label{sec:impact_statement}
CryoAtlas supports reproducible development of cryo-EM atomic model-building methods through curated map--model--sequence data and a common benchmark. More consistent training data and joint evaluation of structural accuracy, map agreement, and stereochemistry can advance structural biology, disease-mechanism studies, and structure-based drug discovery.

Incorrect or weakly supported atomic models may mislead downstream biological interpretation. Benchmark scores should be considered alongside experimental density and model validation, particularly for targets underrepresented after curation and benchmark selection. Users should validate structures before drawing biological conclusions. We have not identified specific additional dual-use risks from this resource beyond those generally associated with structural biology tools.

{\renewcommand{\acksection}{\section*{Acknowledgments}}
\begin{ack}
This work was supported in part by the National Key R\&D Program of China (2025YFA1309603), the National Natural Science Foundation of China (Grant W2431046), the Central Guided Local Science and Technology Foundation of China (YDZX20253100001001), the MoE Key Lab of Intelligent Perception and Human-Machine Collaboration (ShanghaiTech University), and the Shanghai Frontiers Science Center of Human-centered Artificial Intelligence. The experiments were supported by the SIST Computing Platform and HPC at ShanghaiTech University. There are no additional competing interests.
\end{ack}
}
\clearpage

{
\bibliographystyle{unsrt}
\bibliography{reference}

@article{ModelAngelo_2024,
  title={Automated model building and protein identification in cryo-{EM} maps},
  author={Jamali, Kiarash and K{\"a}ll, Lukas and Zhang, Rui and Brown, Alan and Kimanius, Dari and Scheres, Sjors HW},
  journal={Nature},
  volume={628},
  number={8007},
  pages={450--457},
  year={2024},
  publisher={Nature Publishing Group UK London}
}

@article{Deeptracer_2021,
  title={{DeepTracer} for fast de novo cryo-{EM} protein structure modeling and special studies on {CoV}-related complexes},
  author={Pfab, Jonas and Phan, Nhut Minh and Si, Dong},
  journal={Proceedings of the National Academy of Sciences},
  volume={118},
  number={2},
  pages={e2017525118},
  year={2021},
  publisher={National Academy of Sciences}
}

@article{Visualization_alpha_2006,
  title={Visualization of $\alpha$-helices in a 6-angstrom resolution cryoelectron microscopy structure of adenovirus allows refinement of capsid protein assignments},
  author={Saban, Susan D and Silvestry, Mariena and Nemerow, Glen R and Stewart, Phoebe L},
  journal={Journal of Virology},
  volume={80},
  number={24},
  pages={12049--12059},
  year={2006},
  publisher={American Society for Microbiology}
}

@article{Identification_2007,
  title={Identification of secondary structure elements in intermediate-resolution density maps},
  author={Baker, Matthew L and Ju, Tao and Chiu, Wah},
  journal={Structure},
  volume={15},
  number={1},
  pages={7--19},
  year={2007},
  publisher={Elsevier}
}

@article{Chimerax_2021,
  title={{UCSF} {ChimeraX}: Structure visualization for researchers, educators, and developers},
  author={Pettersen, Eric F and Goddard, Thomas D and Huang, Conrad C and Meng, Elaine C and Couch, Gregory S and Croll, Tristan I and Morris, John H and Ferrin, Thomas E},
  journal={Protein Science},
  volume={30},
  number={1},
  pages={70--82},
  year={2021},
  publisher={Wiley Online Library}
}

@article{Qscore_2020,
  title={Measurement of atom resolvability in cryo-{EM} maps with {Q-scores}},
  author={Pintilie, Grigore and Zhang, Kaiming and Su, Zhaoming and Li, Shanshan and Schmid, Michael F and Chiu, Wah},
  journal={Nature Methods},
  volume={17},
  number={3},
  pages={328--334},
  year={2020},
  publisher={Nature Publishing Group US New York}
}

@article{Clustal_2014,
  title={{Clustal Omega}},
  author={Sievers, Fabian and Higgins, Desmond G},
  journal={Current protocols in bioinformatics},
  volume={48},
  number={1},
  pages={3.13.1--3.13.16},
  year={2014},
  publisher={Wiley Online Library},
  doi={10.1002/0471250953.bi0313s48}
}

@article{MMseqs2_2017,
  title={{MMseqs2} enables sensitive protein sequence searching for the analysis of massive data sets},
  author={Steinegger, Martin and S{\"o}ding, Johannes},
  journal={Nature Biotechnology},
  volume={35},
  number={11},
  pages={1026--1028},
  year={2017},
  publisher={Nature Publishing Group US New York}
}

@article{CryoAtom,
  title={{CryoAtom} improves model building for cryo-{EM}},
  author={Su, Baoquan and Huang, Kun and Peng, Zhenling and Amunts, Alexey and Yang, Jianyi},
  journal={Nature Structural \& Molecular Biology},
  pages={351--361},
  year={2026},
  publisher={Nature Publishing Group US New York},
  volume={33},
  number={2},
  doi={10.1038/s41594-025-01713-3}
}

@article{CryoAtom2,
  title={{CryoAtom2}: inverse folding-inspired deep learning enables accurate model building of protein-nucleic acid complexes in cryo-{EM}},
  author={Su, Baoquan and Zhu, Lingfan and Li, Wenfei and Wang, Wenkai and Liang, Jiahui and Zhao, Kailong and Peng, Zhenling and Yang, Jianyi},
  journal={LangTaoSha Preprint Server},
  year={2026},
  doi={10.65215/LTSpreprints.2026.04.14.000185},
  url={https://langtaosha.org.cn/lts/en/preprint/view/185},
  note={Preprint. \url{https://doi.org/10.65215/LTSpreprints.2026.04.14.000185}}
}

@article{E3-Cryofold,
  title={End-to-end Cryo-{EM} complex structure determination with high accuracy and ultra-fast speed},
  author={Wang, Jue and Tan, Cheng and Gao, Zhangyang and Zhang, Guijun and Zhang, Yang and Li, Stan Z},
  journal={Nature Machine Intelligence},
  pages={1091--1103},
  year={2025},
  publisher={Nature Publishing Group UK London},
  volume={7},
  number={7},
  doi={10.1038/s42256-025-01056-0}
}

@article{EMProt,
  title={{EMProt} improves structure determination from cryo-{EM} maps},
  author={Li, Tao and Chen, Ji and Li, Hao and Cao, Hong and Huang, Sheng-You},
  journal={Nature Structural \& Molecular Biology},
  volume={33},
  number={2},
  pages={341--350},
  year={2026},
  publisher={Nature Publishing Group US New York}
}

@article{AF2,
  title={Highly accurate protein structure prediction with {AlphaFold}},
  author={Jumper, John and Evans, Richard and Pritzel, Alexander and Green, Tim and Figurnov, Michael and Ronneberger, Olaf and Tunyasuvunakool, Kathryn and Bates, Russ and {\v{Z}}{\'\i}dek, Augustin and Potapenko, Anna and others},
  journal={Nature},
  volume={596},
  number={7873},
  pages={583--589},
  year={2021},
  publisher={Nature Publishing Group UK London}
}

@article{AF3,
  title={Accurate structure prediction of biomolecular interactions with {AlphaFold} 3},
  author={Abramson, Josh and Adler, Jonas and Dunger, Jack and Evans, Richard and Green, Tim and Pritzel, Alexander and Ronneberger, Olaf and Willmore, Lindsay and Ballard, Andrew J and Bambrick, Joshua and others},
  journal={Nature},
  volume={630},
  number={8016},
  pages={493--500},
  year={2024},
  publisher={Nature Publishing Group UK London}
}

@article{Phenix,
  title={{PHENIX}: a comprehensive {Python}-based system for macromolecular structure solution},
  author={Adams, Paul D and Afonine, Pavel V and Bunk{\'o}czi, G{\'a}bor and Chen, Vincent B and Davis, Ian W and Echols, Nathaniel and Headd, Jeffrey J and Hung, L-W and Kapral, Gary J and Grosse-Kunstleve, Ralf W and others},
  journal={Acta Crystallographica Section D: Biological Crystallography},
  volume={66},
  number={2},
  pages={213--221},
  year={2010},
  publisher={International Union of Crystallography},
  doi={10.1107/S0907444909052925}
}

@article{Molprobity,
  title={{MolProbity}: more and better reference data for improved all-atom structure validation},
  author={Williams, Christopher J and Headd, Jeffrey J and Moriarty, Nigel W and Prisant, Michael G and Videau, Lizbeth L and Deis, Lindsay N and Verma, Vishal and Keedy, Daniel A and Hintze, Bradley J and Chen, Vincent B and others},
  journal={Protein Science},
  volume={27},
  number={1},
  pages={293--315},
  year={2018},
  publisher={Wiley Online Library}
}

@article{USalign,
  title={{US-align}: universal structure alignments of proteins, nucleic acids, and macromolecular complexes},
  author={Zhang, Chengxin and Shine, Morgan and Pyle, Anna Marie and Zhang, Yang},
  journal={Nature Methods},
  volume={19},
  number={9},
  pages={1109--1115},
  year={2022},
  publisher={Nature Publishing Group US New York}
}

@article{PDB_2019,
  title={{Protein Data Bank}: the single global archive for {3D} macromolecular structure data},
  author={{wwPDB Consortium}},
  journal={Nucleic Acids Research},
  volume={47},
  number={D1},
  pages={D520--D528},
  year={2019},
  doi={10.1093/nar/gky949}
}

@article{EMDB_2024,
  title={{EMDB}--the {Electron Microscopy Data Bank}},
  author={{wwPDB Consortium}},
  journal={Nucleic Acids Research},
  volume={52},
  number={D1},
  pages={D456--D465},
  year={2024},
  doi={10.1093/nar/gkad1019}
}

@article{Cryo2StructData_2024,
  title={{Cryo2StructData}: A Large Labeled Cryo-{EM} Density Map Dataset for {AI}-based Modeling of Protein Structures},
  author={Giri, Nabin and Wang, Liguo and Cheng, Jianlin},
  journal={Scientific Data},
  volume={11},
  number={1},
  pages={458},
  year={2024},
  doi={10.1038/s41597-024-03299-9}
}

@article{Coot,
  title={{Coot}: model-building tools for molecular graphics},
  author={Emsley, Paul and Cowtan, Kevin},
  journal={Acta Crystallographica Section D: Biological Crystallography},
  volume={60},
  number={12},
  pages={2126--2132},
  year={2004},
  publisher={International Union of Crystallography},
  doi={10.1107/S0907444904019158}
}

@article{cryo_development,
  title={The development of cryo-{EM} into a mainstream structural biology technique},
  author={Nogales, Eva},
  journal={Nature Methods},
  volume={13},
  number={1},
  pages={24--27},
  year={2016},
  publisher={Nature Publishing Group US New York}
}

@article{cryo_revolution1,
  title={How cryo-{EM} is revolutionizing structural biology},
  author={Bai, Xiao-Chen and McMullan, Greg and Scheres, Sjors HW},
  journal={Trends in Biochemical Sciences},
  volume={40},
  number={1},
  pages={49--57},
  year={2015},
  publisher={Elsevier}
}

@article{cryo_revolution2,
  title={Cryo-electron microscopy for structural analysis of dynamic biological macromolecules},
  author={Murata, Kazuyoshi and Wolf, Matthias},
  journal={Biochimica et Biophysica Acta (BBA)-General Subjects},
  volume={1862},
  number={2},
  pages={324--334},
  year={2018},
  publisher={Elsevier}
}

@article{novo_determine,
  title={De novo protein structure determination from near-atomic-resolution cryo-{EM} maps},
  author={Wang, Ray Yu-Ruei and Kudryashev, Mikhail and Li, Xueming and Egelman, Edward H and Basler, Marek and Cheng, Yifan and Baker, David and DiMaio, Frank},
  journal={Nature Methods},
  volume={12},
  number={4},
  pages={335--338},
  year={2015},
  publisher={Nature Publishing Group US New York}
}

@article{rosettaes,
  title={{RosettaES}: a sampling strategy enabling automated interpretation of difficult cryo-{EM} maps},
  author={Frenz, Brandon and Walls, Alexandra C and Egelman, Edward H and Veesler, David and DiMaio, Frank},
  journal={Nature Methods},
  volume={14},
  number={8},
  pages={797--800},
  year={2017},
  publisher={Nature Publishing Group US New York}
}

@article{MAINMAST,
  title={De novo main-chain modeling for {EM} maps using {MAINMAST}},
  author={Terashi, Genki and Kihara, Daisuke},
  journal={Nature Communications},
  volume={9},
  number={1},
  pages={1618},
  year={2018},
  publisher={Nature Publishing Group UK London}
}

@article{automated_ros,
  title={Automated structure refinement of macromolecular assemblies from cryo-{EM} maps using {Rosetta}},
  author={Wang, Ray Yu-Ruei and Song, Yifan and Barad, Benjamin A and Cheng, Yifan and Fraser, James S and DiMaio, Frank},
  journal={eLife},
  volume={5},
  pages={e17219},
  year={2016},
  publisher={eLife Sciences Publications, Ltd}
}

@article{zhang2025comprehensive,
  title={A comprehensive survey and benchmark of deep learning-based methods for atomic model building from cryo-electron microscopy density maps},
  author={Zhang, Chenwei and Condon, Anne and Dao Duc, Khanh},
  journal={Briefings in Bioinformatics},
  volume={26},
  number={4},
  pages={bbaf322},
  year={2025},
  publisher={Oxford University Press}
}

@article{terashi2024deepmainmast,
  title={{DeepMainmast}: integrated protocol of protein structure modeling for cryo-{EM} with deep learning and structure prediction},
  author={Terashi, Genki and Wang, Xiao and Prasad, Devashish and Nakamura, Tsukasa and Kihara, Daisuke},
  journal={Nature Methods},
  volume={21},
  number={1},
  pages={122--131},
  year={2024},
  publisher={Nature Publishing Group US New York}
}

@article{Boltz_1,
  title={{Boltz-1}: Democratizing Biomolecular Interaction Modeling},
  author={Wohlwend, Jeremy and Corso, Gabriele and Passaro, Saro and Getz, Noah and Reveiz, Mateo and Leidal, Ken and Swiderski, Wojtek and Atkinson, Liam and Portnoi, Tally and Chinn, Itamar and Silterra, Jacob and Jaakkola, Tommi and Barzilay, Regina},
  journal={bioRxiv},
  year={2024},
  doi={10.1101/2024.11.19.624167}
}

@article{Openfold,
  title={{OpenFold}: Retraining {AlphaFold2} yields new insights into its learning mechanisms and capacity for generalization},
  author={Ahdritz, Gustaf and Bouatta, Nazim and Floristean, Christina and Kadyan, Sachin and Xia, Qinghui and Gerecke, William and O'Donnell, Timothy J and Berenberg, Daniel and Fisk, Ian and Zanichelli, Niccol{\`o} and others},
  journal={Nature Methods},
  volume={21},
  number={8},
  pages={1514--1524},
  year={2024},
  publisher={Nature Publishing Group US New York}
}

@article{Seedfold,
  title={{SeedFold}: Scaling Biomolecular Structure Prediction},
  author={Zhou, Yi and Lu, Chan and Ma, Yiming and Qu, Wei and Ye, Fei and Zhang, Kexin and Wang, Lan and Gui, Minrui and Gu, Quanquan},
  journal={arXiv preprint arXiv:2512.24354},
  year={2025},
  doi={10.48550/arXiv.2512.24354}
}

@article{Rosettafold1,
  title={Accurate prediction of protein structures and interactions using a three-track neural network},
  author={Baek, Minkyung and DiMaio, Frank and Anishchenko, Ivan and Dauparas, Justas and Ovchinnikov, Sergey and Lee, Gyu Rie and Wang, Jue and Cong, Qian and Kinch, Lisa N and Schaeffer, R Dustin and others},
  journal={Science},
  volume={373},
  number={6557},
  pages={871--876},
  year={2021},
  publisher={American Association for the Advancement of Science}
}

@article{Rosettafold2,
  title={Accurate prediction of protein--nucleic acid complexes using {RoseTTAFoldNA}},
  author={Baek, Minkyung and McHugh, Ryan and Anishchenko, Ivan and Jiang, Hanlun and Baker, David and DiMaio, Frank},
  journal={Nature Methods},
  volume={21},
  number={1},
  pages={117--121},
  year={2024},
  publisher={Nature Publishing Group US New York}
}

@article{ESM2,
  title={Evolutionary-scale prediction of atomic-level protein structure with a language model},
  author={Lin, Zeming and Akin, Halil and Rao, Roshan and Hie, Brian and Zhu, Zhongkai and Lu, Wenting and Smetanin, Nikita and Verkuil, Robert and Kabeli, Ori and Shmueli, Yaniv and others},
  journal={Science},
  volume={379},
  number={6637},
  pages={1123--1130},
  year={2023},
  publisher={American Association for the Advancement of Science}
}

@article{ESM3,
  title={Simulating 500 million years of evolution with a language model},
  author={Hayes, Thomas and Rao, Roshan and Akin, Halil and Sofroniew, Nicholas J and Oktay, Deniz and Lin, Zeming and Verkuil, Robert and Tran, Vincent Q and Deaton, Jonathan and Wiggert, Marius and others},
  journal={Science},
  volume={387},
  number={6736},
  pages={850--858},
  year={2025},
  publisher={American Association for the Advancement of Science}
}

@article{terwilliger2024alphafold,
  title={{AlphaFold} predictions are valuable hypotheses and accelerate but do not replace experimental structure determination},
  author={Terwilliger, Thomas C and Liebschner, Dorothee and Croll, Tristan I and Williams, Christopher J and McCoy, Airlie J and Poon, Billy K and Afonine, Pavel V and Oeffner, Robert D and Richardson, Jane S and Read, Randy J and others},
  journal={Nature Methods},
  volume={21},
  number={1},
  pages={110--116},
  year={2024},
  publisher={Nature Publishing Group US New York}
}

@article{morichaud2023structural,
  title={Structural basis of the mycobacterial stress-response {RNA} polymerase auto-inhibition via oligomerization},
  author={Morichaud, Zakia and Trapani, Stefano and Vishwakarma, Rishi K and Chaloin, Laurent and Lionne, Corinne and Lai-Kee-Him, Jos{\'e}phine and Bron, Patrick and Brodolin, Konstantin},
  journal={Nature Communications},
  volume={14},
  number={1},
  pages={484},
  year={2023},
  publisher={Nature Publishing Group UK London}
}

@article{chen2024structural,
  title={Structural mechanism of voltage-gated sodium channel slow inactivation},
  author={Chen, Huiwen and Xia, Zhanyi and Dong, Jie and Huang, Bo and Zhang, Jiangtao and Zhou, Feng and Yan, Rui and Shi, Yiqiang and Gong, Jianke and Jiang, Juquan and others},
  journal={Nature Communications},
  volume={15},
  number={1},
  pages={3691},
  year={2024},
  publisher={Nature Publishing Group UK London}
}

@inproceedings{CryoBoltz,
  title={Multiscale guidance of protein structure prediction with heterogeneous cryo-{EM} data},
  author={Raghu, Rishwanth and Levy, Axel and Wetzstein, Gordon and Zhong, Ellen D},
  booktitle={Advances in Neural Information Processing Systems},
  volume={38},
  pages={10175--10203},
  year={2025},
  doi={10.52202/085713-0344}
}

@article{Buccaneer_2006,
  title={The Buccaneer software for automated model building. 1. Tracing protein chains},
  author={Cowtan, Kevin},
  journal={Acta Crystallographica Section D: Biological Crystallography},
  volume={62},
  number={9},
  pages={1002--1011},
  year={2006},
  publisher={International Union of Crystallography},
  doi={10.1107/S0907444906022116}
}

@article{Pathwalking_2016,
  title={De Novo modeling in cryo-{EM} density maps with {Pathwalking}},
  author={Chen, Muyuan and Baldwin, Philip R and Ludtke, Steven J and Baker, Matthew L},
  journal={Journal of Structural Biology},
  volume={196},
  number={3},
  pages={289--298},
  year={2016},
  publisher={Elsevier}
}

@article{si2020deep,
  title={Deep learning to predict protein backbone structure from high-resolution cryo-{EM} density maps},
  author={Si, Dong and Moritz, Spencer A and Pfab, Jonas and Hou, Jie and Cao, Renzhi and Wang, Liguo and Wu, Tianqi and Cheng, Jianlin},
  journal={Scientific Reports},
  volume={10},
  number={1},
  pages={4282},
  year={2020},
  publisher={Nature Publishing Group UK London}
}

@article{zhang2022cr,
  title={{CR-I-TASSER}: assemble protein structures from cryo-{EM} density maps using deep convolutional neural networks},
  author={Zhang, Xi and Zhang, Biao and Freddolino, Lydia and Zhang, Yang},
  journal={Nature Methods},
  volume={19},
  number={2},
  pages={195--204},
  year={2022},
  publisher={Nature Publishing Group US New York}
}

@article{giri2024novo,
  title={De novo atomic protein structure modeling for {cryoEM} density maps using {3D} transformer and {HMM}},
  author={Giri, Nabin and Cheng, Jianlin},
  journal={Nature Communications},
  volume={15},
  number={1},
  pages={5511},
  year={2024},
  publisher={Nature Publishing Group UK London}
}

@article{Secondary_detection,
  title={Protein secondary structure detection in intermediate-resolution cryo-{EM} maps using deep learning},
  author={Maddhuri Venkata Subramaniya, Sai Raghavendra and Terashi, Genki and Kihara, Daisuke},
  journal={Nature Methods},
  volume={16},
  number={9},
  pages={911--917},
  year={2019},
  publisher={Nature Publishing Group US New York}
}

@article{wang2023cryoread,
  title={{CryoREAD}: de novo structure modeling for nucleic acids in cryo-{EM} maps using deep learning},
  author={Wang, Xiao and Terashi, Genki and Kihara, Daisuke},
  journal={Nature Methods},
  volume={20},
  number={11},
  pages={1739--1747},
  year={2023},
  publisher={Nature Publishing Group US New York}
}

@article{EmbedOpt_2026,
  title={Robust Inference-Time Steering of Protein Diffusion Models via Embedding Optimization},
  author={Li, Minhuan and Han, Jiequn and Cossio, Pilar and Wu, Luhuan},
  journal={arXiv preprint arXiv:2602.05285},
  year={2026},
  doi={10.48550/arXiv.2602.05285},
  url={https://arxiv.org/abs/2602.05285}
}

@inproceedings{CryoACE_2026,
  title={{CryoACE}: An Atom-centric Framework for Accurate and Automated Model Building in Cryo-{EM}},
  author={Li, Minzhang and Li, Mingrui and Qin, Weichen and Chen, Qihe and Shen, Sixian and Pei, Yuan and Zhang, Jiakai and Yu, Jingyi},
  booktitle={Proceedings of the 43rd International Conference on Machine Learning},
  series={Proceedings of Machine Learning Research},
  volume={306},
  pages={69214--69230},
  year={2026},
  publisher={PMLR},
  url={https://proceedings.mlr.press/v306/li26eh.html}
}

@article{ROCKET_2026,
  title={{AlphaFold} as a prior: experimental structure determination conditioned on a pretrained neural network},
  author={Fadini, Alisia and Li, Minhuan and McCoy, Airlie J. and Banjara, Suresh and Okumura, Hiroki and Napier, Eve and Fontana, Pietro and Khan, Amir R. and Jovine, Luca and Terwilliger, Thomas C. and Read, Randy J. and Hekstra, Doeke R. and AlQuraishi, Mohammed},
  journal={Nature Methods},
  volume={23},
  pages={785--795},
  year={2026},
  doi={10.1038/s41592-026-03047-4}
}

@article{MICA_2025,
  title={Multimodal deep learning integration of cryo-{EM} and {AlphaFold3} for high-accuracy protein structure determination},
  author={Gyawali, Rajan and Dhakal, Ashwin and Cheng, Jianlin},
  journal={Communications Chemistry},
  volume={8},
  number={1},
  pages={320},
  year={2025},
  doi={10.1038/s42004-025-01718-5}
}

@article{GuidedAF3_2026,
  title={Experiment-guided {AlphaFold3} resolves measurement-consistent protein ensembles},
  author={Maddipatla, Advaith and {Sellam Bojan}, Nadav and Bojan, Meital and Masalitin, Volodymyr and Vedula, Sanketh and Schanda, Paul and Marx, Ailie and Bronstein, Alex M.},
  journal={Nature Biotechnology},
  year={2026},
  doi={10.1038/s41587-026-03166-5},
  note={\url{https://doi.org/10.1038/s41587-026-03166-5}}
}

@article{CryoDataBot_2025,
  title={{CryoDataBot}: a pipeline to curate cryo{EM} datasets for {AI}-driven structural biology},
  author={Xu, Qibo and Wu, Leon and Rebelo, Michael and Feng, Shi and Yu, Xinye and Farheen, Farhanaz and Kihara, Daisuke and Zhou, Z. Hong},
  journal={GigaScience},
  volume={14},
  pages={giaf127},
  year={2025},
  doi={10.1093/gigascience/giaf127}
}

@inproceedings{ModelAngelo_ICLR2023,
  title={A Graph Neural Network Approach to Automated Model Building in Cryo-{EM} Maps},
  author={Jamali, Kiarash and Kimanius, Dari and Scheres, Sjors HW},
  booktitle={International Conference on Learning Representations},
  year={2023},
  url={https://openreview.net/forum?id=65XDF_nwI61}
}
}

\clearpage

\appendix
\begingroup
\hypersetup{hidelinks}
\setlength{\parindent}{0pt}
\setlength{\parskip}{0pt}
\newcommand{\appcontentsdots}{%
  \nobreak\hspace{0.5em}\leaders\hbox to 0.8em{\hss.\hss}\hfill
  \nobreak\hspace{0.5em}}
\newcommand{\appcontentssection}[2]{%
  \par\vspace{8pt}\noindent
  {\rmfamily\bfseries\fontsize{11}{13}\selectfont
  \hyperref[#1]{\makebox[1.8em][l]{\ref*{#1}}#2}\hfill
  \hyperref[#1]{\pageref*{#1}}\par}\nobreak\vspace{2pt}}
\newcommand{\appcontentssubsection}[2]{%
  \noindent{\rmfamily\fontsize{10.5}{13}\selectfont
  \hspace*{1.8em}\hyperref[#1]{\makebox[2.6em][l]{\ref*{#1}}#2}%
  \appcontentsdots\hyperref[#1]{\pageref*{#1}}\par}}
\newcommand{\appcontentssubsubsection}[2]{%
  \noindent{\rmfamily\fontsize{9.5}{11.5}\selectfont
  \hspace*{4.4em}\hyperref[#1]{\makebox[3.3em][l]{\ref*{#1}}#2}%
  \appcontentsdots\hyperref[#1]{\pageref*{#1}}\par}}

{\centering\rmfamily\bfseries\fontsize{18}{22}\selectfont
Supplementary Material\par}
\vspace{25pt}
{\rmfamily\bfseries\fontsize{14}{18}\selectfont
Contents of the Supplementary Material\par}
\vspace{5pt}

\appcontentssection{sec:appendix_metrics}{Metric Formulations and Analysis}
\appcontentssubsection{subsec:appendix_stereo}{Stereochemical Quality}
\appcontentssubsubsection{subsubsec:geom_formulation}{Definition of Geom Score}
\appcontentssubsubsection{subsubsec:geom_robustness}{Robustness analysis: agreement with MolProbity}
\appcontentssubsubsection{subsubsec:stereo_table}{Per-Method Stereochemical Metrics}
\appcontentssubsubsection{subsubsec:stereo_analysis}{Analysis of raw metrics}
\appcontentssubsection{subsec:appendix_gt}{Structural Accuracy}
\appcontentssubsubsection{subsubsec:gt_table}{Per-Method Structural Metrics}
\appcontentssubsubsection{subsubsec:gt_formulation}{Definition, biases, and weighting rationale}
\appcontentssubsubsection{subsubsec:gt_analysis}{Analysis of raw metrics}
\appcontentssubsection{subsec:Model-Map fitness}{Model-Map fitness}

\appcontentssection{sec:appendix_unmodeled}{Unmodeled-Density Ratio Formula}

\appcontentssection{app:reimpl-v21}{ModelAngelo CNN Reimplementation}
\appcontentssubsection{app:ma_differences}{Differences from the Paper}
\appcontentssubsection{app:ma_training_recipe}{Training Recipe}
\appcontentssubsection{app:ma_reproduction_notes}{Reproduction Notes}
\appcontentssubsection{app:ma_variant_breakdown}{Detailed metric breakdown for training utility}
\appcontentssubsubsection{app:ma_stereochemistry}{Stereochemistry}
\appcontentssubsubsection{app:ma_structural_accuracy}{Structural Accuracy}
\appcontentssubsubsection{app:ma_map_fitness}{Model-to-map fitness}

\appcontentssection{sec:c2s_issues}{Potential Issues with Cryo2StructData}
\appcontentssubsection{subsec:c2s_train_stability}{Training-time Instability}
\appcontentssubsection{subsec:c2s_cleanliness}{Label and Sample Cleanliness}

\appcontentssection{app:stratified_figures}{Supplementary Stratified Performance Figures}
\appcontentssubsection{app:Res}{Resolution-dependent Performance}
\appcontentssubsection{app:length}{Length-dependent Performance}
\appcontentssubsection{app:category}{Protein-category Effects}
\appcontentssubsection{app:identity}{Test--Train Sequence Similarity}

\appcontentssection{sec:AF3_failure}{Failure Cases of AlphaFold~3}

\appcontentssection{app:dataset_composition}{Dataset Composition and Curation Analysis}
\appcontentssubsection{app:dataset_groups}{Dataset groups and composition}
\appcontentssubsection{app:filtering_selection}{Filtering and benchmark selection}
\appcontentssubsection{app:curation_comparison}{Comparison of curation workflows}
\endgroup
\clearpage

\section{Metric Formulations And Analysis}
\label{sec:appendix_metrics}

The three composites separately summarize stereochemistry, reference agreement, and map fit; individual metrics diagnose incomplete models and other failures. No cross-dimension total is defined. Component weights are fixed design choices; Geom Score standardization is estimated from benchmark outputs as specified below.

\subsection*{Benchmark overview and variability}
Table~\ref{tab:main_results_sd} extends the main-text comparison in Table~\ref{tab:main_results} with sample standard deviations across targets, characterizing variation in performance across the benchmark. The following sections define the three composite scores and examine their constituent metrics to explain the method-level differences summarized here.

\begin{table}[H]
\centering
\footnotesize
\setlength{\tabcolsep}{0.5pt}
\renewcommand{\arraystretch}{1.05}
\begin{tabular*}{\textwidth}{@{\extracolsep{\fill}}lccccccc@{}}
\toprule
\textbf{Method} & \shortstack{\textbf{Geom}\\\textbf{Score} ($\uparrow$)} & \shortstack{\textbf{C$\alpha$ RMSD}\\($\downarrow$)} & \shortstack{\textbf{TM-score}\\($\uparrow$)} & \shortstack{\textbf{Seq. recall}\\($\uparrow$)} & \shortstack{\textbf{GT Score}\\($\uparrow$)} & \shortstack{\textbf{Q-score}\\($\uparrow$)} & \shortstack{\textbf{Map Score}\\($\uparrow$)} \\
\midrule
Phenix (24h)$^\dagger$ & $0.462$ & $3.562$ & $0.075$ & $0.063$ & $0.194$ & $0.539$ & $0.193$ \\
ModelAngelo & $0.462\!\pm\!0.036$ & $\mathbf{1.014}\!\pm\!0.837$ & $0.648\!\pm\!0.277$ & $0.739\!\pm\!0.251$ & $0.723\!\pm\!0.178$ & $\mathbf{0.600}\!\pm\!0.102$ & $0.520\!\pm\!0.133$ \\
CryoAtom2 & $0.537\!\pm\!0.040$ & $1.198\!\pm\!0.890$ & $0.763\!\pm\!0.207$ & $\mathbf{0.899}\!\pm\!0.123$ & $\mathbf{0.787}\!\pm\!0.131$ & $0.585\!\pm\!0.097$ & $\mathbf{0.568}\!\pm\!0.085$ \\
EMProt & $0.484\!\pm\!0.027$ & $1.224\!\pm\!0.987$ & $0.639\!\pm\!0.283$ & $0.734\!\pm\!0.258$ & $0.702\!\pm\!0.200$ & $0.588\!\pm\!0.086$ & $0.517\!\pm\!0.129$ \\
E3-CryoFold & $0.165\!\pm\!0.028$ & $3.278\!\pm\!1.988$ & $0.739\!\pm\!0.268$ & $0.622\!\pm\!0.288$ & $0.539\!\pm\!0.236$ & $0.332\!\pm\!0.092$ & $0.305\!\pm\!0.105$ \\
\midrule
AlphaFold~3 & $\mathbf{0.598}\!\pm\!0.029$ & $3.655\!\pm\!2.483$ & $\mathbf{0.813}\!\pm\!0.199$ & $0.647\!\pm\!0.266$ & $0.552\!\pm\!0.219$ & $0.202\!\pm\!0.147$ & $0.220\!\pm\!0.140$ \\
\bottomrule
\end{tabular*}
\vspace{2mm}
\caption{\textbf{Main performance evaluation.} Entries are mean $\pm$ target-level sample SD over available metric values. All methods except Phenix use the 618-target common benchmark. $^\dagger$Phenix means use the 366 targets completed within 24 hours; only means are reported because SDs are unavailable. Detailed metrics are in Appendix Tables~\ref{tab:supp_stereo}, \ref{tab:supp_stereo_subscores}, \ref{tab:supp_gt}, and \ref{tab:supp_map}. RMSD values are in \AA{}.}
\label{tab:main_results_sd}
\end{table}

\subsection{Stereochemical Quality}
\label{subsec:appendix_stereo}

\subsubsection{Definition of Geom Score}
\label{subsubsec:geom_formulation}

Seven raw \texttt{phenix.molprobity} indicators are aggregated into a bounded composite $\text{Geom Score}$ in three steps.

\textbf{Step 1: Direction-adjusted transform.} Each indicator $i$ is oriented so that larger denotes better quality, with heavy-tailed counts log-transformed:
\begin{equation}
\tilde{y}_i = s_i \cdot \phi_i(x_i),
\qquad
\phi_i(x) =
\begin{cases}
\log\!\bigl(1+\max(x,0)\bigr) & i \in \mathcal{I}_{\log}, \\
x & i \in \mathcal{I}_{\text{lin}},
\end{cases}
\qquad s_i \in \{+1,-1\}.
\end{equation}
$\mathcal{I}_{\log} = \{$clashscore, rama\_outlier, rotamer\_outlier, c$\beta$\_deviations$\}$; $\mathcal{I}_{\text{lin}} = \{$rama\_favored, rms\_bonds, rms\_angles$\}$. We use $s_i=+1$ for rama\_favored and $s_i=-1$ for all other indicators.

\textbf{Step 2: Pooled robust standardization.} Statistics are computed from the available values pooled across the five methods and 618 targets and mapped to the unit interval via a logistic sigmoid:
\begin{equation}
q_i = \sigma\!\left( \frac{\tilde{y}_i - \widehat{m}_i}{\widehat{\sigma}_i + \epsilon} \right),
\qquad
\widehat{\sigma}_i = \max\!\Bigl(\text{IQR}_i,\, \tfrac{1}{2}(P_{90} - P_{10})\Bigr).
\end{equation}
Here $\widehat{m}_i$ is the median of the direction-adjusted indicator over valid pooled observations; $\mathrm{IQR}_i$ and $P_{10},P_{90}$ are computed over the same observations. We use $\epsilon=10^{-6}$ to prevent division by zero. Missing values are excluded when estimating these statistics.

\textbf{Step 3: Hierarchical aggregation.} The seven $q$-values are partitioned into four sub-categories: Steric $= \{$clashscore$\}$; Backbone $= \{$rama\_favored, rama\_outlier$\}$; Sidechain $= \{$rotamer\_outlier, c$\beta$\_deviations$\}$; Bonded $= \{$rms\_bonds, rms\_angles$\}$. Let $\mathcal{G}$ denote these four groups, and define $\mathcal{G}_{\mathrm{valid}}=\{g\in\mathcal{G}:q_i\text{ is available for every }i\in g\}$ for each evaluated structure. A group is omitted if any of its indicators is missing; we do not average the remaining indicators within that group. The composite averages the valid group scores:
\begin{equation}
S_g = \Bigl(\textstyle\prod_{i \in g} q_i\Bigr)^{1/|g|},
\qquad
\text{Geom Score} = \frac{1}{|\mathcal{G}_{\mathrm{valid}}|}\sum_{g \in \mathcal{G}_{\mathrm{valid}}} S_g.
\end{equation}

If $\mathcal{G}_{\mathrm{valid}}$ is empty, Geom Score is unavailable. Geometric averaging within each valid group penalizes imbalance among related indicators, reducing the influence of a favorable value when another component is poor. Equal arithmetic weighting across valid groups balances steric, backbone, side-chain, and bonded geometry without overweighting groups that contain more indicators.

\subsubsection{Robustness analysis: agreement with MolProbity}
\label{subsubsec:geom_robustness}

To verify alignment with the established MolProbity composite~\citep{Molprobity}, we compute per-method Spearman rank correlation on the 618-target benchmark (Table~\ref{tab:supp_geom_robustness}). Since Geom Score is higher-is-better and MolProbity is lower-is-better, agreement appears as strongly negative $\rho$: $|\rho| > 0.87$ for every method and $> 0.94$ for four of the five, with pooled $|\rho| = 0.986$ over 2,887 valid method--target pairs.

\begin{table}[h]
\centering
\small
\begin{tabular}{lcc}
\toprule
\textbf{Method} & \textbf{Spearman $\rho$} & \textbf{$n$} \\
\midrule
ModelAngelo  & $-0.980$ & 617 \\
CryoAtom2     & $-0.991$ & 618 \\
EMProt       & $-0.982$ & 618 \\
E3-CryoFold  & $-0.872$ & 416 \\
AlphaFold 3  & $-0.947$ & 618 \\
\midrule
Pooled (all five) & $-0.986$ & 2887 \\
\bottomrule
\end{tabular}
\vspace{3mm}
\caption{Per-method Spearman correlation between Geom Score and MolProbity. E3-CryoFold has fewer pairs because some of its outputs cause MolProbity sub-metrics to be uncomputable.}
\label{tab:supp_geom_robustness}
\end{table}

Geom Score is bounded to $(0,1)$, includes all seven raw indicators rather than MolProbity's three, and is calibrated on cryo-EM benchmark outputs rather than X-ray reference distributions.

\subsubsection{Per-Method Stereochemical Metrics}
\label{subsubsec:stereo_table}

\begin{table}[H]
\centering
\small
\setlength{\tabcolsep}{3pt}
\begin{tabular}{lcccccccc}
\toprule
\textbf{Method} & \textbf{Rama-F (\%)} & \textbf{Rama-O (\%)} & \textbf{Rot-O (\%)} & \textbf{C$\beta$ Dev} & \textbf{Clash} & \textbf{Bonds} & \textbf{Angles} & \textbf{MolProb} \\
\midrule
ModelAngelo & 91.262 & 3.074 & 10.388 & \textbf{0.000} & 69.207 & 0.067 & 5.481 & 3.530 \\
CryoAtom2 & \textbf{95.629} & 1.210 & 6.278 & \textbf{0.000} & 39.856 & 0.030 & 2.324 & 2.860 \\
EMProt & 92.244 & 2.399 & 13.227 & \textbf{0.000} & 69.703 & 0.031 & 2.369 & 3.583 \\
E3-CryoFold & 54.632 & 29.198 & 28.939 & 1136.202 & 431.612 & 0.347 & 20.753 & 5.125 \\
AlphaFold 3 & 94.821 & \textbf{1.200} & \textbf{1.442} & 1.068 & \textbf{12.511} & \textbf{0.013} & \textbf{1.648} & \textbf{1.978} \\
\bottomrule
\end{tabular}%
\vspace{2mm}
\caption{Per-method raw stereochemical indicators on the 618-target benchmark. Rama-F/O: Ramachandran favored/outlier residues; Rot-O: rotamer outliers; C$\beta$ Dev: mean number of C$\beta$ deviations per structure; Clash: all-atom clashscore; Bonds/Angles: RMS deviations of bond lengths (\AA) and angles ($^\circ$) from ideal geometry; MolProb: composite MolProbity score. Rama-F is higher-is-better; all other columns are lower-is-better. Best per column is shown in \textbf{bold}.}
\label{tab:supp_stereo}
\end{table}

\begin{table}[H]
\centering
\small
\setlength{\tabcolsep}{6pt}
\begin{tabular}{lccccc}
\toprule
\textbf{Method} & \textbf{$S_{\text{Steric}}$} & \textbf{$S_{\text{Backbone}}$} & \textbf{$S_{\text{Sidechain}}$} & \textbf{$S_{\text{Bonded}}$} & \textbf{Geom Score} \\
\midrule
ModelAngelo & 0.477 & 0.464 & 0.491 & 0.417 & 0.462 \\
CryoAtom2 & 0.561 & \textbf{0.544} & 0.536 & 0.506 & 0.537 \\
EMProt & 0.474 & 0.488 & 0.467 & 0.505 & 0.484 \\
E3-CryoFold & 0.242 & 0.143 & 0.182 & 0.094 & 0.165 \\
AlphaFold 3 & \textbf{0.705} & 0.542 & \textbf{0.605} & \textbf{0.539} & \textbf{0.598} \\
\bottomrule
\end{tabular}
\vspace{2mm}
\caption{Per-method sub-category quality scores $S_g \in (0, 1)$ and the aggregated $\text{Geom Score} \in (0, 1)$. Each $S_g$ is the geometric mean of standardized indicators within sub-category $g$ as defined in Section~\ref{subsubsec:geom_formulation}; Geom Score is the arithmetic mean of the four sub-scores. Best per column is shown in \textbf{bold}.}
\label{tab:supp_stereo_subscores}
\end{table}

\subsubsection{Analysis of raw metrics}
\label{subsubsec:stereo_analysis}

\paragraph{CryoAtom2 leads among density-guided methods.}
Among density-guided methods, CryoAtom2 is clearly best (Geom Score 0.537 vs.\ 0.484/0.462 for EMProt/ModelAngelo), with systematically tighter raw indicators: 95.6\% Ramachandran-favored (vs.\ 91.3--92.2\%), 1.21\% outliers (vs.\ 2.40--3.07\%), clashscore 39.9 (vs.\ 69.2--69.7), and lower RMS bond/angle deviations, indicating stronger geometric regularity than the other density-guided baselines.

\paragraph{E3-CryoFold's pronounced geometric deficits.}
E3-CryoFold shows substantially poorer stereochemical quality across the reported indicators (e.g., 54.6\% Ramachandran-favored vs.\ 91.3--95.6\% for the other methods; clashscore 431.6 vs.\ 12--70 for the others), and \texttt{phenix.molprobity} fails to compute on about one third of its outputs. These deficits can complicate standard real-space refinement, which presumes a near-correct starting backbone. They likely reflect combined architectural, training, and data issues; a finer attribution is beyond this benchmark.

\paragraph{Sequence-based AlphaFold~3 dominates stereochemistry.}
AlphaFold~3 leads three sub-categories (Steric 0.705, Sidechain 0.605, Bonded 0.539) and nearly ties CryoAtom2 on Backbone (0.542 vs.\ 0.544), with lower clashscore (12.5 vs.\ 39.9--69.7) and the smallest RMS bond/angle deviations (0.013/1.65; Table~\ref{tab:supp_stereo}). Its large training corpus embeds extensive structural priors in the model weights, whereas density-guided methods have substantially smaller training datasets. Incorporating pretrained structural knowledge into density-guided pipelines is therefore a promising direction.

\paragraph{Implications.}
Geom Score measures pre-refinement stereochemistry. Real-space refinement (e.g., \texttt{phenix.real\_space\_refine}) can bring local geometry broadly in line with AlphaFold~3's pretrained quality, but typically requires tens of minutes to hours of CPU time per target and depends on map resolution and the starting model. CryoAtom2's stronger baseline (0.537) therefore suggests lower downstream refinement costs than ModelAngelo or EMProt.

\subsection{Structural Accuracy}
\label{subsec:appendix_gt}

\subsubsection{Per-Method Structural Metrics}
\label{subsubsec:gt_table}

\begin{table}[H]
\centering
\small
\setlength{\tabcolsep}{3pt}
\begin{tabular}{lcccccccc}
\toprule
\textbf{Method} & \textbf{TM} & \textbf{RMSD-L} & \textbf{RMSD-G} & \textbf{Res. Cov.} & \textbf{Seq. Match} & \textbf{Seq. Recall} & \textbf{Compl.} & \textbf{GT Score} \\
\midrule
ModelAngelo & 0.648 & \textbf{1.014} & \textbf{4.436} & 0.760 & 0.952 & 0.739 & 0.779 & 0.723 \\
CryoAtom2 & 0.763 & 1.198 & 6.418 & \textbf{0.932} & \textbf{0.961} & \textbf{0.899} & 1.010 & \textbf{0.787} \\
EMProt & 0.639 & 1.224 & 5.472 & 0.766 & 0.937 & 0.734 & 0.779 & 0.702 \\
E3-CryoFold & 0.739 & 3.278 & 12.700 & 0.788 & 0.727 & 0.622 & 0.936 & 0.539 \\
AlphaFold 3 & \textbf{0.813} & 3.655 & 15.307 & 0.752 & 0.820 & 0.647 & 1.348 & 0.552 \\
\bottomrule
\end{tabular}
\vspace{2mm}
\caption{Per-method structural accuracy metrics on the 618-target benchmark. TM: TM-score; RMSD-L: local C$\alpha$ RMSD on aligned residues; RMSD-G: global C$\alpha$ RMSD on all paired C$\alpha$ atoms; Cov.: residue coverage; Seq. Match: sequence match; Seq. Recall: sequence recall; Compl.: completeness (model residues / GT residues). TM, Cov., Match, Recall, and GT Score are higher-is-better; RMSD-L and RMSD-G are lower-is-better. Best per column in \textbf{bold}. RMSD values are in \AA{}.}
\label{tab:supp_gt}
\end{table}

\subsubsection{Definition, biases, and weighting rationale}
\label{subsubsec:gt_formulation}

We assess agreement with the ground-truth atomic structure along three complementary axes: global topology (TM-score), local geometric accuracy (C$\alpha$ RMSD), and sequence-level coverage (sequence\_recall). They are combined into a bounded composite $\text{GT Score} \in (0,1]$:
\begin{equation}
\text{GT Score}
= 0.15 \cdot \text{TM-score}
+ 0.45 \cdot \exp\left(-\text{C}_{\alpha}\text{ RMSD}/3\right)
+ 0.40 \cdot \text{sequence\_recall}.
\label{eq:gt_score}
\end{equation}

\paragraph{Component selection.}
We considered TM-score, local C$\alpha$ RMSD, residue coverage, sequence match, sequence recall, and completeness. Sequence recall combines coverage with correct residue identification and is strongly correlated with coverage and sequence match (Spearman $\rho=0.967$ and $0.843$, respectively). Retaining recall without separately weighting its two factors avoids counting the same recovery information multiple times. Completeness is a predicted-to-reference length ratio, can exceed one, and does not establish correctness; it is therefore reported separately. The retained components describe global topology, local atomic placement, and correctly recovered sequence information.

\paragraph{Raw indicators and their biases.}
\textbf{TM-score} is normalized by GT length, ignores unmodeled residues, and is sensitive to chain breaks, disadvantaging partial or fragmented density-guided models. \textbf{Local C$\alpha$ RMSD} (RMSD-L, from \texttt{phenix.chain\_comparison}) averages residue pairs aligned within a distance cutoff and favors models restricted to well-resolved regions: ModelAngelo achieves the lowest RMSD-L (1.014\,\AA{}) at only 0.76 coverage. \textbf{Sequence\_recall} penalizes both omitted and misidentified GT residues. Table~\ref{tab:supp_gt} also reports \textbf{Coverage} (fraction of GT residues aligned to model residues), \textbf{Completeness} (model/GT length), and in-house \textbf{global C$\alpha$ RMSD} (RMSD-G) over all paired C$\alpha$ atoms after global superposition. RMSD-G is less robust but indicates the displacement of misaligned residues. Coverage and Completeness distinguish reference recovery from total model extent; these raw indicators should be interpreted alongside the three GT Score components.

\paragraph{Weighting rationale.}
The weights $(0.15,\,0.45,\,0.40)$ were specified as fixed design choices, not optimized to maximize benchmark performance. TM-score receives weight $0.15$ because reference-length normalization can disadvantage partial but accurately built models. Local C$\alpha$ RMSD receives weight $0.45$, slightly above sequence recall ($0.40$), to emphasize accurate atomic placement while retaining a substantial penalty for missing or incorrectly assigned residues. The transform $\exp(-\mathrm{RMSD}/3)$ maps local RMSD to $(0,1]$ with a $3$\,\AA{} sensitivity scale, appropriate to the typical $0.5$--$5$\,\AA{} range of aligned-residue errors. These weights balance complementary failure modes and are not claimed to be uniquely optimal.
 
\subsubsection{Analysis of raw metrics}
\label{subsubsec:gt_analysis}

As illustrated in Table~\ref{tab:supp_gt}, ModelAngelo, CryoAtom2, and AlphaFold~3 differ markedly in how much they build and how accurately. ModelAngelo's Completeness (0.779) is essentially equal to its Coverage (0.760), indicating that it commits only to residues it can confidently place; this restraint yields the lowest RMSD-L (1.014\,\AA{}). CryoAtom2's Completeness (1.010) is noticeably above its Coverage (0.932), meaning it additionally builds atoms in less certain regions: some of these placements are inaccurate (raising RMSD-L to 1.198\,\AA{}), but the excess also extends modeling into regions the GT did not annotate, potentially recovering protein content absent from the deposited structure. ModelAngelo is therefore locally more precise, but CryoAtom2 delivers a more complete reconstruction whose Sequence Recall (0.899) is 22\% higher than ModelAngelo's (0.739). Both methods achieve very high Sequence Match (CryoAtom2 0.961, ModelAngelo 0.952), showing comparable local residue-type identification. AlphaFold~3, by design, predicts the entire input FASTA regardless of map evidence, yielding the highest Completeness (1.348) and TM-score (0.813); the elevated Completeness primarily reflects that GT structures typically omit flexible or unresolved regions, so AlphaFold~3 outputs are on average ${\sim}35\%$ longer than the GT-modeled chains. Its low Coverage (0.752) and the largest RMSD-L (3.655\,\AA{}) show that the predicted topology, although plausible globally, does not match the GT atoms locally, a consequence of sequence-only prediction without map constraints.

\subsection{Model-Map fitness}
\label{subsec:Model-Map fitness}
\begin{table}[H]
\centering
\small
\setlength{\tabcolsep}{3pt} 
\begin{tabular}{lccccccccc}
\toprule
\textbf{Method} & \textbf{CC$_\text{mask}$} & \textbf{CC$_\text{vol}$} & \textbf{CC$_\text{peaks}$} & \textbf{CC$_\text{box}$} & \textbf{cc\_mean} & \textbf{Q-score} & \textbf{Recall Q} & \textbf{Map Score} \\
\midrule
ModelAngelo & 0.680 & 0.678 & 0.404 & 0.553 & 0.579 & \textbf{0.600} & 0.462 & 0.520 \\
CryoAtom2 & 0.660 & 0.665 & \textbf{0.486} & \textbf{0.602} & \textbf{0.603} & 0.585 & \textbf{0.533} & \textbf{0.568} \\
EMProt & \textbf{0.692} & \textbf{0.689} & 0.402 & 0.555 & 0.584 & 0.588 & 0.449 & 0.517 \\
E3-CryoFold & 0.399 & 0.439 & 0.270 & 0.432 & 0.385 & 0.332 & 0.224 & 0.305 \\
AlphaFold 3 & 0.298 & 0.299 & 0.148 & 0.374 & 0.280 & 0.202 & 0.161 & 0.220 \\
\bottomrule
\end{tabular}%
\vspace{2mm}
\caption{Per-method model-to-map fitness metrics on the 618-target benchmark. CC subscripts denote \texttt{phenix.real\_space\_correlation} variants; cc\_mean is the average of the four CC variants; Q-score is per-atom local resolvability; Recall Qscore is the coverage-weighted local term. All columns are higher-is-better. Best per column in \textbf{bold}.}
\label{tab:supp_map}
\end{table}

\paragraph{Q-score and CC variants.}
Q-score measures the local resolvability of modeled atoms in the cryo-EM density and generally decreases at lower local resolution as atomic peaks become broader and less defined. ModelAngelo obtains the highest average Q-score (0.600), partly because it tends to build well-resolved, high-density regions; skipping peripheral or ambiguous residues biases the per-atom average toward easier regions. The four \texttt{phenix} CC variants assess map--model agreement at complementary scales. CC$_\text{box}$ measures overall similarity over the full map box, CC$_\text{mask}$ evaluates fit around modeled atoms, CC$_\text{vol}$ captures agreement with the high-density molecular envelope, and CC$_\text{peaks}$ compares the strongest density peaks. The lower CC$_\text{peaks}$ values show that peak-level agreement remains challenging, as this metric is sensitive to missing or misplaced atoms at sharp density maxima. CryoAtom2's advantage on CC$_\text{peaks}$ is consistent with its more complete coverage and stronger local placement.

\paragraph{Component selection and weighting.}
Map Score weights global density agreement (the CC family) and local atomic resolvability (Q-score) equally. Averaging the four CC variants prevents their number from outweighing the local term. For nonnegative Q-score, multiplying by sequence recall reduces the local contribution of incomplete recovery. For negative Q-score, the product is instead shifted toward zero as recall decreases; the term is therefore coverage-weighted local resolvability, not a universal incompleteness penalty. Because this correction uses reference information and shares recall with GT Score, Map Score is reference-assisted and the composites are not statistically independent. Raw CC values and Q-score remain necessary for interpretation.

\section{Unmodeled-Density Ratio Formula}
\label{sec:appendix_unmodeled}

This appendix defines the per-entry unmodeled-density ratio: the fraction of valid experimental signal not covered by the deposited atomic model. It serves as an indicator of dataset cleanliness. See Fig.~\ref{fig:c2s_examples} for an example of an entry with a large unmodeled-density ratio.

Let $M$ be the cryo-EM density on a voxel grid $\mathcal{V}$, and let $S$ be the model-implied density on the same grid, obtained from \texttt{chimerax.molmap}. With a molmap threshold $\tau_S = 0.1$ and an adaptively chosen map threshold $\tau_M$ (specified below), the model-support and map-support sets are
\begin{align}
\mathcal{V}_{\text{model}} &= \{v \in \mathcal{V} : S(v) \geq \tau_S\}, \\
\mathcal{V}_{\text{supp}}(\tau_M) &= \{v \in \mathcal{V} : M(v) \geq \tau_M\}.
\end{align}
We calibrate $\tau_M$ relative to the typical map intensity inside the model region: $\tau_M^\star$ is the largest threshold at which at least a fraction $p$ of model-support voxels still satisfy $M(v) \geq \tau_M^\star$:
\begin{equation}
\tau_M^\star = \sup\!\left\{\tau_M \;\middle|\; \frac{\bigl|\mathcal{V}_{\text{model}} \cap \mathcal{V}_{\text{supp}}(\tau_M)\bigr|}{|\mathcal{V}_{\text{model}}|} \;\geq\; p \right\}, \qquad p = 0.5.
\label{eq:tau_M_star}
\end{equation}
With $p = 0.5$, $\tau_M^\star$ is the median of map intensities within the model region; since this distribution is approximately Gaussian, the median captures the main body of the valid signal. $\tau_M^\star$ is found by binary search ($20$ iterations) over $[0.01,\, \max_v M(v)]$. The per-entry unmodeled-density ratio is
\begin{equation}
r \;=\; \frac{\bigl|\mathcal{V}_{\text{supp}}(\tau_M^\star) \setminus \mathcal{V}_{\text{model}}\bigr|}{\bigl|\mathcal{V}_{\text{supp}}(\tau_M^\star)\bigr|}\;\in\;[0,\,1),
\label{eq:r_unmodeled}
\end{equation}

\section{ModelAngelo CNN Reimplementation}
\label{app:reimpl-v21}

ModelAngelo's training pipeline is not open-sourced; neither are the training pipelines of the other recent automated cryo-EM model-building systems we surveyed. To enable controlled re-training experiments, we therefore re-implemented one method from its published description. We chose ModelAngelo because it is among the most widely used and cited recent end-to-end systems.

We re-trained only ModelAngelo's graph-initialisation CNN, while reusing the graph-refinement GNN unchanged from the official inference bundle~\cite{ModelAngelo_2024}. The architecture and loss family follow the ICLR 2023 description~\cite{ModelAngelo_ICLR2023}: a 3D residual/FPN backbone with instance normalization and ReLU, $1.5$\,\AA{} voxel resampling, weighted focal binary cross-entropy with per-voxel class-imbalance weighting, and a Tversky region loss introduced in the later stage of training.

\subsection{Differences from the Paper}
\label{app:ma_differences}

The following procedural details are either unspecified in the ICLR 2023 version of ModelAngelo~\cite{ModelAngelo_ICLR2023} or differ from our implementation.

For the \textbf{Tversky loss}, the paper states that the asymmetric formulation favors recall but does not specify the exponents $(\alpha,\beta)$. We use $(\alpha,\beta)=(0.7,0.3)$. Since the paper only states that Tversky loss is added ``in the later parts of training'', we delay it until step $13{,}500$ after a one-epoch focal-only warmup, and then ramp its weight from $0$ to $1$ with a cosine schedule. At the reported step-$50$k checkpoint, the Tversky weight reaches approximately $0.98$ of its maximum.

For \textbf{optimization}, the paper does not specify the optimiser or learning-rate schedule. We use AdamW with $\beta_1=0.9$, $\beta_2=0.999$, weight decay $10^{-2}$, cosine learning-rate annealing from a peak of $7.4{\times}10^{-4}$, and gradient-norm clipping at $0.5$. The peak learning rate follows square-root batch-size scaling: $3\times10^{-4}\sqrt{96/16}=7.35\times10^{-4}$, rounded to $7.4\times10^{-4}$. We performed no dataset-specific hyperparameter search; architecture, optimizer, learning-rate schedule, augmentation, crop sampling, and training budget were shared across the dataset comparisons.

For \textbf{augmentation}, the paper samples a B-factor sharpening parameter uniformly from $[-30,+30]$\,\AA$^2$, whereas we use $[-25,+25]$\,\AA$^2$. Our training set covers a broad resolution range, including both high-resolution ($<3$\,\AA) and low-resolution ($4$--$6$\,\AA) maps; for the latter, stronger B-factor perturbations tended to degrade already-blurred features, so the tighter range gave more stable behaviour across the dataset. We also omit two augmentations described in the paper: random $90^\circ$ rotations of the map and target, and additive colored noise on the cryo-EM map.

\subsection{Training Recipe}
\label{app:ma_training_recipe}

We use ModelAngelo's preprocessing code. Training uses an effective batch size of $96$ ($12$ per GPU on $8$ H20 GPUs) with \texttt{bf16} mixed precision. We extract subvolumes of $64\times64\times64$ voxels from each map: with probability $0.25$ the crop is sampled uniformly at random, and otherwise it is centered at a randomly selected modeled atom. Random crops expose the network to background-dominated regions that occupy most of the map volume, improving robustness and suppressing false positives. Atom-centered crops ensure that most updates contain protein signal, allowing the network to learn the true structural-density distribution rather than being overwhelmed by empty background. We train for $50{,}000$ steps and use the resulting checkpoint for downstream benchmark evaluation. Early training stability is assessed separately over the first $5{,}000$ steps. The per-voxel focal weighting and the Tversky add-in follow the published loss formulation.

\subsection{Reproduction Notes}
\label{app:ma_reproduction_notes}

Because the official ModelAngelo training hyperparameters are unavailable, we cannot perform a head-to-head reproduction with identical settings. Instead, we verify internal consistency: the same training recipe is applied to both the Cryo2StructData training set and our CryoAtlas training set, and the resulting checkpoints are evaluated under the same protocol (Sec.~\ref{sec:evaluate}). The CryoAtlas-trained EMDB checkpoint also exceeds the official ModelAngelo inference bundle on the reported test metrics. Thus, while we cannot claim global optimality of our hyperparameters, the listed choices are competitive with the published configuration under our evaluation and provide a concrete reference for future reproduction.

\paragraph{Subsequent implementation audit.}
Subsequent audits identified inconsistent handling of START-derived origins for nonstandard MRC axis orders. Explicit MRC ORIGIN values could also be incorrectly scaled by voxel size despite already being in angstroms. In our later colored-noise implementation, full-map and crop Fourier-shell indices were incorrectly matched; this augmentation was disabled in the reported training experiments. All reported training results retain the original implementation and settings, and the released reproduction code incorporates these corrections.

\subsection{Detailed metric breakdown for training utility}
\label{app:ma_variant_breakdown}

This section expands Table~\ref{tab:training_utility}. We evaluate three MA-architecture checkpoints on the same $618$-target benchmark used in the main text: (i)~the \emph{official} ModelAngelo checkpoint released by the authors; (ii)~\textbf{MA (Cryo2Struct)}, retrained on the public Cryo2StructData training set with our pipeline; (iii)~\textbf{MA (CryoAtlas)}, retrained on our proposed CryoAtlas training set with the same pipeline. 

\noindent\begin{minipage}{\textwidth}
\subsubsection*{Checkpoint comparison and variability}
Table~\ref{tab:training_utility_sd} supplements the training-utility results in Table~\ref{tab:training_utility} with sample standard deviations across the 618 targets. It summarizes the three checkpoints across the same evaluation dimensions; the metric-specific breakdowns below examine the stereochemical, structural, and map-fitting differences underlying this comparison.

\begin{table}[H]
\centering
\footnotesize
\setlength{\tabcolsep}{0.5pt}
\renewcommand{\arraystretch}{1.05}
\begin{tabular*}{\textwidth}{@{\extracolsep{\fill}}lccccccc@{}}
\toprule
\textbf{MA variant} & \shortstack{\textbf{Geom}\\\textbf{Score} ($\uparrow$)} & \shortstack{\textbf{C$\alpha$ RMSD}\\($\downarrow$)} & \shortstack{\textbf{TM-score}\\($\uparrow$)} & \shortstack{\textbf{Seq. recall}\\($\uparrow$)} & \shortstack{\textbf{GT Score}\\($\uparrow$)} & \shortstack{\textbf{Q-score}\\($\uparrow$)} & \shortstack{\textbf{Map Score}\\($\uparrow$)} \\
\midrule
Official & $0.462\!\pm\!0.036$ & $1.014\!\pm\!0.837$ & $0.648\!\pm\!0.277$ & $0.739\!\pm\!0.251$ & $0.723\!\pm\!0.178$ & $0.600\!\pm\!0.102$ & $0.520\!\pm\!0.133$ \\
C2S & $0.455\!\pm\!0.038$ & $1.108\!\pm\!0.948$ & $0.604\!\pm\!0.284$ & $0.708\!\pm\!0.264$ & $0.697\!\pm\!0.190$ & $0.599\!\pm\!0.102$ & $0.505\!\pm\!0.139$ \\
CryoAtlas & $\mathbf{0.470}\!\pm\!0.038$ & $\mathbf{0.769}\!\pm\!0.454$ & $\mathbf{0.680}\!\pm\!0.271$ & $\mathbf{0.753}\!\pm\!0.247$ & $\mathbf{0.755}\!\pm\!0.165$ & $\mathbf{0.603}\!\pm\!0.102$ & $\mathbf{0.532}\!\pm\!0.129$ \\
\bottomrule
\end{tabular*}
\vspace{2mm}
\caption{Performance of three ModelAngelo (MA) variants on the 618-target benchmark: the official checkpoint and CNNs trained on Cryo2StructData (C2S) or CryoAtlas. Entries are mean $\pm$ sample SD across targets. Full metric breakdowns are in Appendix Tables~\ref{tab:supp_stereo_train}, \ref{tab:supp_geom_train}, \ref{tab:supp_gt_train}, and \ref{tab:supp_map_train}. RMSD values are in \AA{}.}
\label{tab:training_utility_sd}
\end{table}
\end{minipage}

\subsubsection{Stereochemistry}
\label{app:ma_stereochemistry}

\begin{table}[H]
\centering
\small
\setlength{\tabcolsep}{3pt}
\begin{tabular}{lcccccccc}
\toprule
\textbf{Method} & \textbf{Rama-F (\%)} & \textbf{Rama-O (\%)} & \textbf{Rot-O (\%)} & \textbf{C$\beta$ Dev} & \textbf{Clash} & \textbf{Bonds} & \textbf{Angles} & \textbf{MolProb} \\
\midrule
ModelAngelo          & 91.262 & 3.074 & 10.388 & \textbf{0.000} & 69.207 & 0.067 & 5.481 & 3.530 \\
MA (Cryo2Struct)     & 90.854 & 3.297 & 10.410 & \textbf{0.000} & 71.524 & 0.077 & 5.823 & 3.553 \\
MA (CryoAtlas)       & \textbf{91.399} & \textbf{2.860} & \textbf{10.110} & \textbf{0.000} & \textbf{66.189} & \textbf{0.059} & \textbf{5.164} & \textbf{3.489} \\
\bottomrule
\end{tabular}
\vspace{2mm}
\caption{Raw stereochemical indicators on the 618-target benchmark. Rama-F/O: Ramachandran favored / outlier residues; Rot-O: rotamer outliers; C$\beta$ Dev: mean number of C$\beta$ deviations per structure; Clash: all-atom clashscore; Bonds / Angles: RMS deviations of bond lengths (\AA) and angles ($^\circ$) from ideal geometry; MolProb: composite MolProbity score. Rama-F is higher-is-better; all other columns are lower-is-better.}
\label{tab:supp_stereo_train}
\end{table}

\vspace{-4mm}

\begin{table}[H]
\centering
\small
\setlength{\tabcolsep}{3pt}
\begin{tabular}{lccccc}
\toprule
\textbf{Method} & \textbf{$S_{\text{Steric}}$} & \textbf{$S_{\text{Backbone}}$} & \textbf{$S_{\text{Sidechain}}$} & \textbf{$S_{\text{Bonded}}$} & \textbf{Geom Score} \\
\midrule
ModelAngelo          & 0.477 & 0.464 & 0.491 & 0.417 & 0.462 \\
MA (Cryo2Struct)     & 0.472 & 0.458 & 0.492 & 0.400 & 0.455 \\
MA (CryoAtlas)       & \textbf{0.485} & \textbf{0.470} & \textbf{0.494} & \textbf{0.430} & \textbf{0.470} \\
\bottomrule
\end{tabular}
\vspace{2mm}
\caption{Stereochemistry sub-scores on the 618-target benchmark. $S_{\text{Steric}}$, $S_{\text{Backbone}}$, $S_{\text{Sidechain}}$ and $S_{\text{Bonded}}$ are the four normalized geometry components, and \texttt{Geom Score} is their composite. All columns are higher-is-better.}
\label{tab:supp_geom_train}
\end{table}

\subsubsection{Structural Accuracy}
\label{app:ma_structural_accuracy}

\begin{table}[H]
\centering
\small
\setlength{\tabcolsep}{3pt}
\begin{tabular}{lccccccc}
\toprule
\textbf{Method} & \textbf{TM} & \textbf{RMSD} & \textbf{Res. Cov.} & \textbf{Seq. Match} & \textbf{Seq. Recall} & \textbf{Compl.} & \textbf{GT Score} \\
\midrule
ModelAngelo          & 0.648 & 1.014 & 0.760 & 0.952 & 0.739 & 0.779 & 0.723 \\
MA (Cryo2Struct)     & 0.604 & 1.108 & 0.728 & 0.947 & 0.708 & 0.748 & 0.697 \\
MA (CryoAtlas)       & \textbf{0.680} & \textbf{0.769} & \textbf{0.771} & \textbf{0.955} & \textbf{0.753} & \textbf{0.787} & \textbf{0.755} \\
\bottomrule
\end{tabular}
\vspace{2mm}
\caption{Structural accuracy metrics on the 618-target benchmark. TM: TM-score; RMSD: C$\alpha$ RMSD on aligned residues; Res.\ Cov.: residue coverage; Seq.\ Match: sequence match; Seq.\ Recall: sequence recall; Compl.: completeness (model residues / GT residues). TM, Res.\ Cov., Match, Recall, and GT Score are higher-is-better; RMSD is lower-is-better. RMSD values are in \AA{}.}
\label{tab:supp_gt_train}
\end{table}

\subsubsection{Model-to-map fitness}
\label{app:ma_map_fitness}

\begin{table}[H]
\centering
\small
\setlength{\tabcolsep}{3pt}
\begin{tabular}{lcccccccc}
\toprule
\textbf{Method} & \textbf{CC$_\text{mask}$} & \textbf{CC$_\text{vol}$} & \textbf{CC$_\text{peaks}$} & \textbf{CC$_\text{box}$} & \textbf{cc\_mean} & \textbf{Q-score} & \textbf{Recall Q} & \textbf{Map Score} \\
\midrule
ModelAngelo          & 0.680 & 0.678 & 0.404 & 0.553 & 0.579 & 0.600 & 0.462 & 0.520 \\
MA (Cryo2Struct)     & 0.671 & 0.669 & 0.381 & 0.539 & 0.565 & 0.599 & 0.445 & 0.505 \\
MA (CryoAtlas)       & \textbf{0.693} & \textbf{0.690} & \textbf{0.418} & \textbf{0.562} & \textbf{0.591} & \textbf{0.603} & \textbf{0.473} & \textbf{0.532} \\
\bottomrule
\end{tabular}
\vspace{2mm}
\caption{Model-to-map fitness metrics on the 618-target benchmark. CC subscripts denote \texttt{phenix.real\_space\_correlation} variants; cc\_mean is the average of the four CC variants; Q-score is per-atom local resolvability; Recall Q is the coverage-weighted local term. All columns are higher-is-better. }
\label{tab:supp_map_train}
\end{table}

\section{Potential Issues with Cryo2StructData}
\label{sec:c2s_issues}

Cryo2StructData is, to our knowledge, the only publicly available training dataset tailored to end-to-end \emph{de novo} cryo-EM atomic model building. This resource supports reproducible research and enabled the controlled retraining experiments in this work.

Our experiments identify two practical limitations relevant to its use as an off-the-shelf training resource: training instability and non-negligible label and sample contamination. We examine these limitations to characterize the additional stabilization and filtering that may be required for downstream applications.

\subsection{Training-time Instability}
\label{subsec:c2s_train_stability}

\begin{figure}[H]
\centering
\begin{minipage}[t]{0.48\linewidth}
\centering
\includegraphics[width=\linewidth]{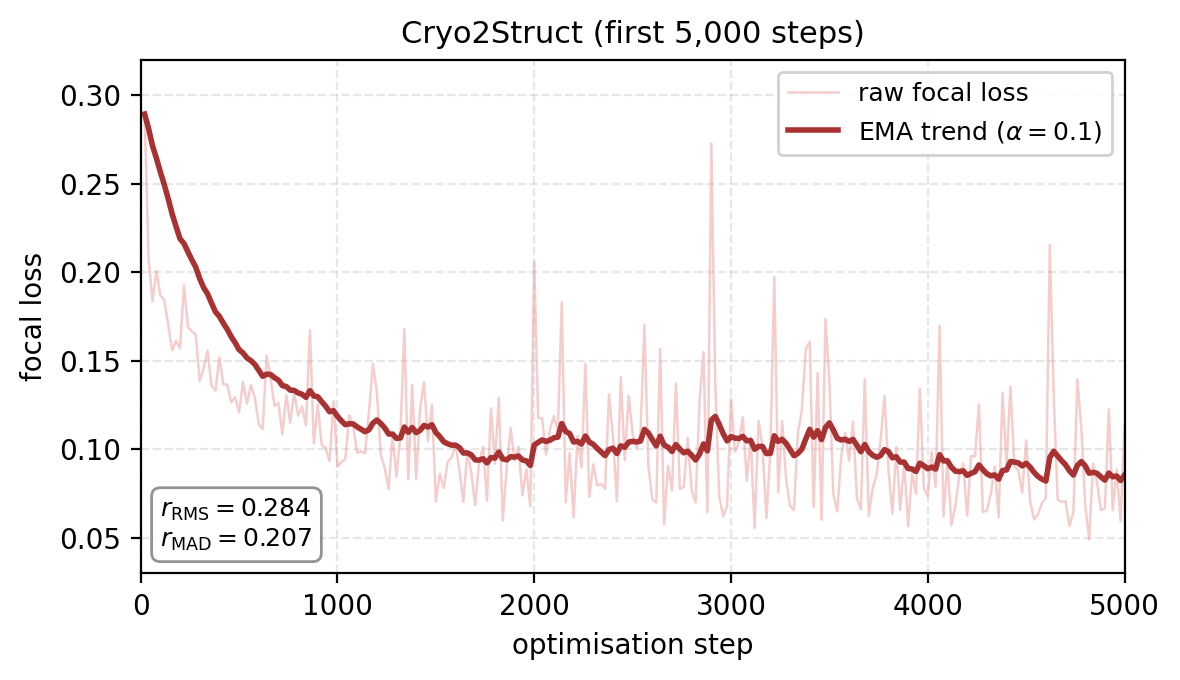}
\end{minipage}
\hfill
\begin{minipage}[t]{0.48\linewidth}
\centering
\includegraphics[width=\linewidth]{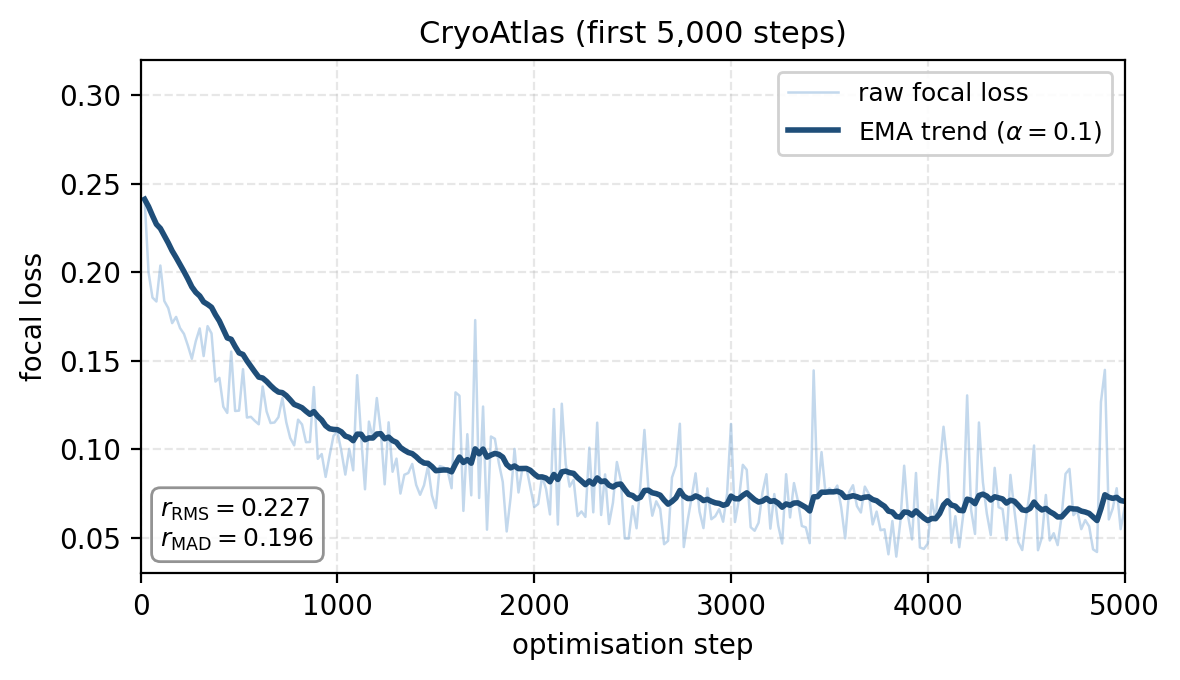}
\end{minipage}
\caption{
Per-step \texttt{train/focal} over the first $5{,}000$ optimization steps for a single-seed run on each dataset, with the EMA trend ($\alpha=0.1$) overlaid. Cryo2StructData exhibits larger short-term fluctuations and occasional sharp loss spikes, whereas CryoAtlas follows a smoother training trajectory.
}
\label{fig:focal_curves}
\end{figure}

Under the same reimplemented training pipeline, Cryo2StructData exhibits larger short-term focal-loss fluctuations than CryoAtlas, most prominently during the first $\sim\!5{,}000$ optimization steps. Both runs remain in the focal-only warm-up stage throughout this window, before the introduction of Tversky loss, allowing early training stability to be compared without confounding from changes in the loss schedule. The analysis uses logged per-step \texttt{train/focal} values from a single-seed run for each dataset (Figure~\ref{fig:focal_curves}).

We detrend each loss curve by subtracting an exponential moving average (EMA) and quantify the normalized residual fluctuations using two complementary statistics. Residual RMS captures large loss spikes, whereas median absolute deviation (MAD) characterizes typical step-to-step variability and is less sensitive to outliers. The relative ordering of the datasets is unchanged across EMA coefficients $\alpha\in[0.05,0.30]$. In Table~\ref{tab:rebuttal_training_stability}, EMA loss is the endpoint of a separate focal-loss EMA with decay 0.99; the residual statistics summarize fluctuations over the full $5{,}000$-step window.

\paragraph{Results.}
Cryo2StructData exhibits greater spike-sensitive volatility during early training. This difference is robust to the EMA coefficient: over $\alpha\in[0.05,0.30]$, the Cryo2StructData-to-CryoAtlas RMS ratio is $1.19$--$1.29\times$, compared with $1.06$--$1.09\times$ for MAD. These observations alone do not establish label contamination, but are consistent with a small number of noisy or inconsistent samples, including map--structure misalignments, unmodeled density regions, abnormal labels, and preprocessing mismatches. We therefore examine sample-level quality in Section~\ref{subsec:c2s_cleanliness}.

\subsection{Label and Sample Cleanliness}
\label{subsec:c2s_cleanliness}

We assess sample-level quality using the per-EMD mean Q-score~\citep{Qscore_2020}, computed using UCSF ChimeraX~\citep{Chimerax_2021}. Q-score measures local agreement between deposited atomic coordinates and the experimental cryo-EM density: higher values indicate stronger density support, whereas low or negative values indicate weak or inconsistent support. Because CNN segmentation models use deposited coordinates as voxel-space supervision, poor map--model agreement can introduce label noise. We use Q-score $<0.3$ to flag weak supervision and Q-score $<0$ to flag severe map--structure inconsistency.

\paragraph{Q-score audit.}
The audit compares four datasets: the full CryoAtlas training set, its subset with deposited resolution $\leq 4\,\text{\AA}$, the normalized maps released by Cryo2StructData, and raw EMDB full maps re-downloaded for the same Cryo2StructData entries. Since Q-score tends to decrease at lower resolution, the CryoAtlas $\leq 4\,\text{\AA}$ subset provides a resolution-matched reference. Table~\ref{tab:c2s_qscore_stats} summarizes the per-EMD mean Q-score distributions, and Table~\ref{tab:c2s_qscore_thresholds} reports the numbers of entries with observed Q-scores below selected quality thresholds.

\begin{table}[H]
\centering
\small
\setlength{\tabcolsep}{3pt}
\resizebox{\linewidth}{!}{
\begin{tabular}{lcccc}
\toprule
\textbf{Statistic} 
& \textbf{CryoAtlas Full}
& \textbf{CryoAtlas $\leq 4\,\text{\AA}$}
& \textbf{Cryo2StructData Normalized}
& \textbf{Cryo2StructData Raw EMDB} \\
\midrule
Mean   & 0.596 & 0.611 & 0.416 & 0.595 \\
Q1     & 0.520 & 0.540 & 0.290 & 0.540 \\
Median & 0.610 & 0.620 & 0.440 & 0.610 \\
Q3     & 0.680 & 0.680 & 0.560 & 0.670 \\
Min    & 0.300 & 0.300 & $-0.660$ & $-0.110$ \\
Max    & 0.950 & 0.950 & 0.910 & 0.950 \\
\bottomrule
\end{tabular}
}
\vspace{2mm}
\caption{
Per-EMD mean Q-score distributions computed with UCSF ChimeraX. The full CryoAtlas training set contains 16,657 entries, its $\leq 4\,\text{\AA}$ subset contains 15,402 entries, and both Cryo2StructData variants comprise the same 7,390 entries. These are cohort sizes; Q-score summary statistics use available measurements.
}
\label{tab:c2s_qscore_stats}
\end{table}

\begin{table}[H]
\centering
\small
\setlength{\tabcolsep}{3pt}
\resizebox{\linewidth}{!}{
\begin{tabular}{lcccc}
\toprule
\textbf{Threshold} 
& \textbf{CryoAtlas Full}
& \textbf{CryoAtlas $\leq 4\,\text{\AA}$}
& \textbf{Cryo2StructData Normalized}
& \textbf{Cryo2StructData Raw EMDB} \\
\midrule
Q-score $<0$   & 0       & 0       & 181     & 25 \\
Q-score $<0.1$ & 0       & 0       & 472     & 50 \\
Q-score $<0.2$ & 0       & 0       & 962   & 67 \\
Q-score $<0.3$ & 0       & 0       & 1{,}845 & 136 \\
Q-score $<0.5$ & 3{,}239 & 2{,}182 & 4{,}542 & 1{,}194 \\
\bottomrule
\end{tabular}
}
\vspace{2mm}
\caption{
Counts of entries with available per-EMD mean Q-scores below each threshold. Unavailable Q-scores are excluded from these counts and are not treated as passing the thresholds. Low Q-scores indicate weak map--model agreement and potentially noisy voxel-space supervision.
}
\label{tab:c2s_qscore_thresholds}
\end{table}

\begin{figure}[t]
\centering

\begin{minipage}[t]{0.48\linewidth}
    \centering
    \includegraphics[width=\linewidth]{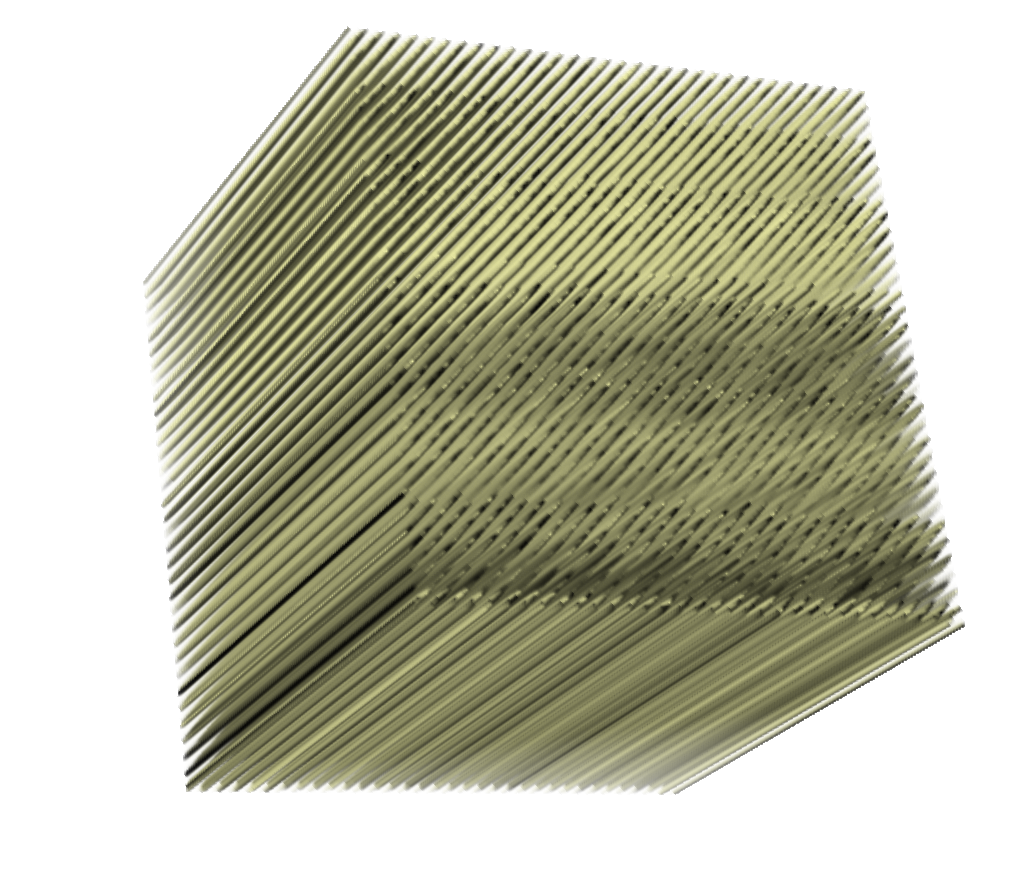}
\end{minipage}
\hfill
\begin{minipage}[t]{0.48\linewidth}
    \centering
    \includegraphics[width=\linewidth]{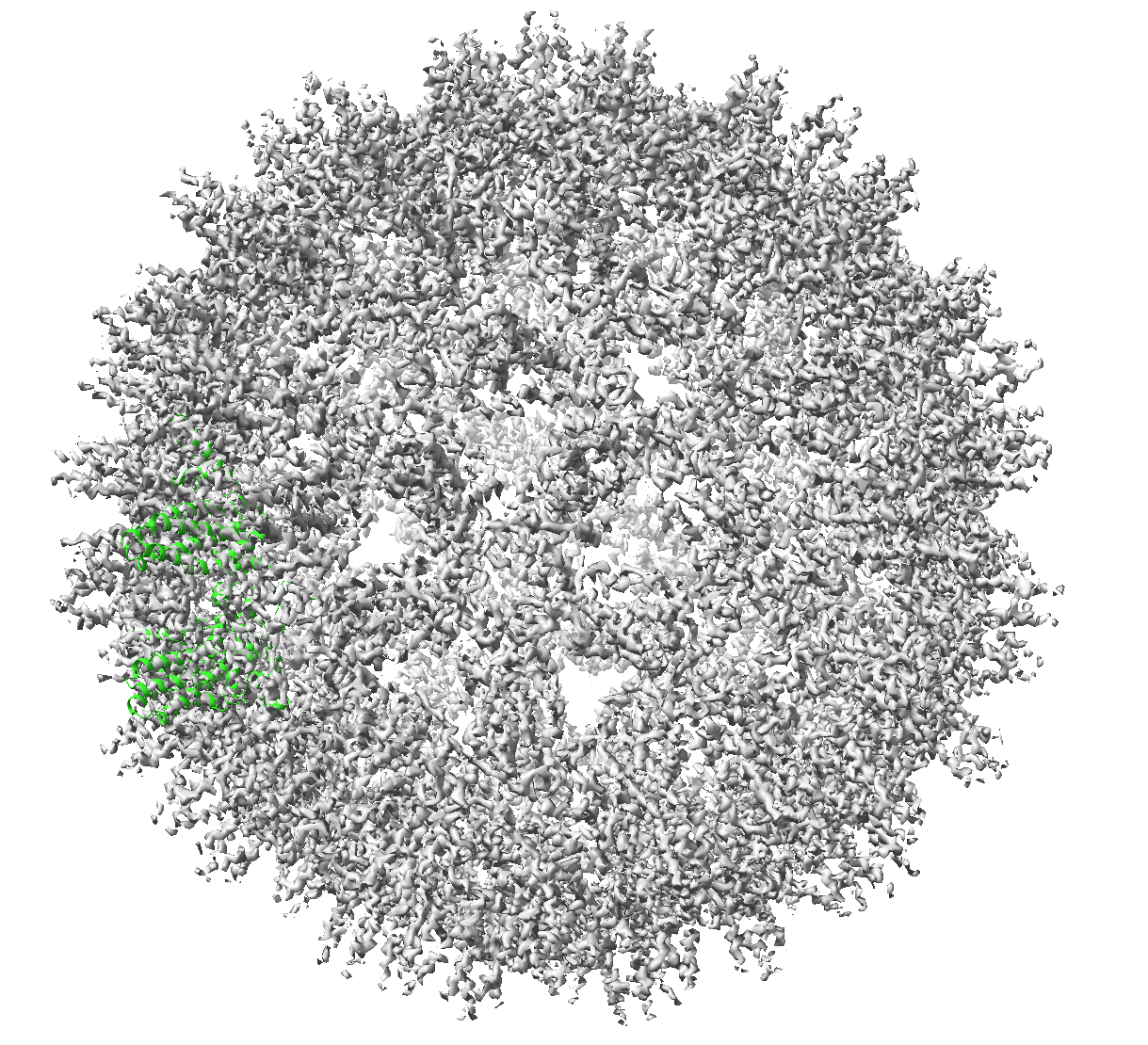}
\end{minipage}

\vspace{2mm}

\begin{minipage}[t]{0.48\linewidth}
    \centering
    \includegraphics[width=\linewidth]{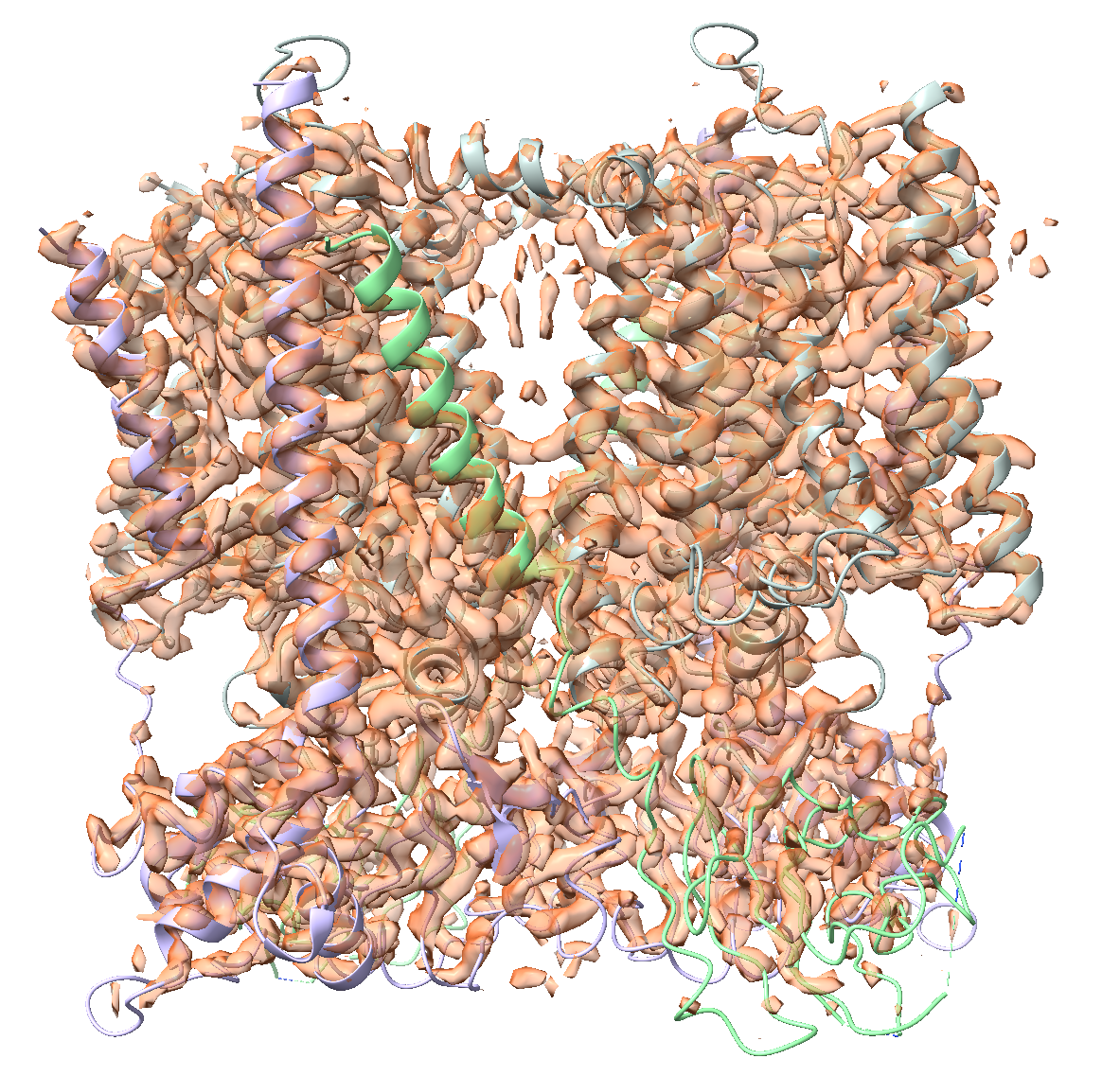}
\end{minipage}
\hfill
\begin{minipage}[t]{0.48\linewidth}
    \centering
    \includegraphics[width=\linewidth]{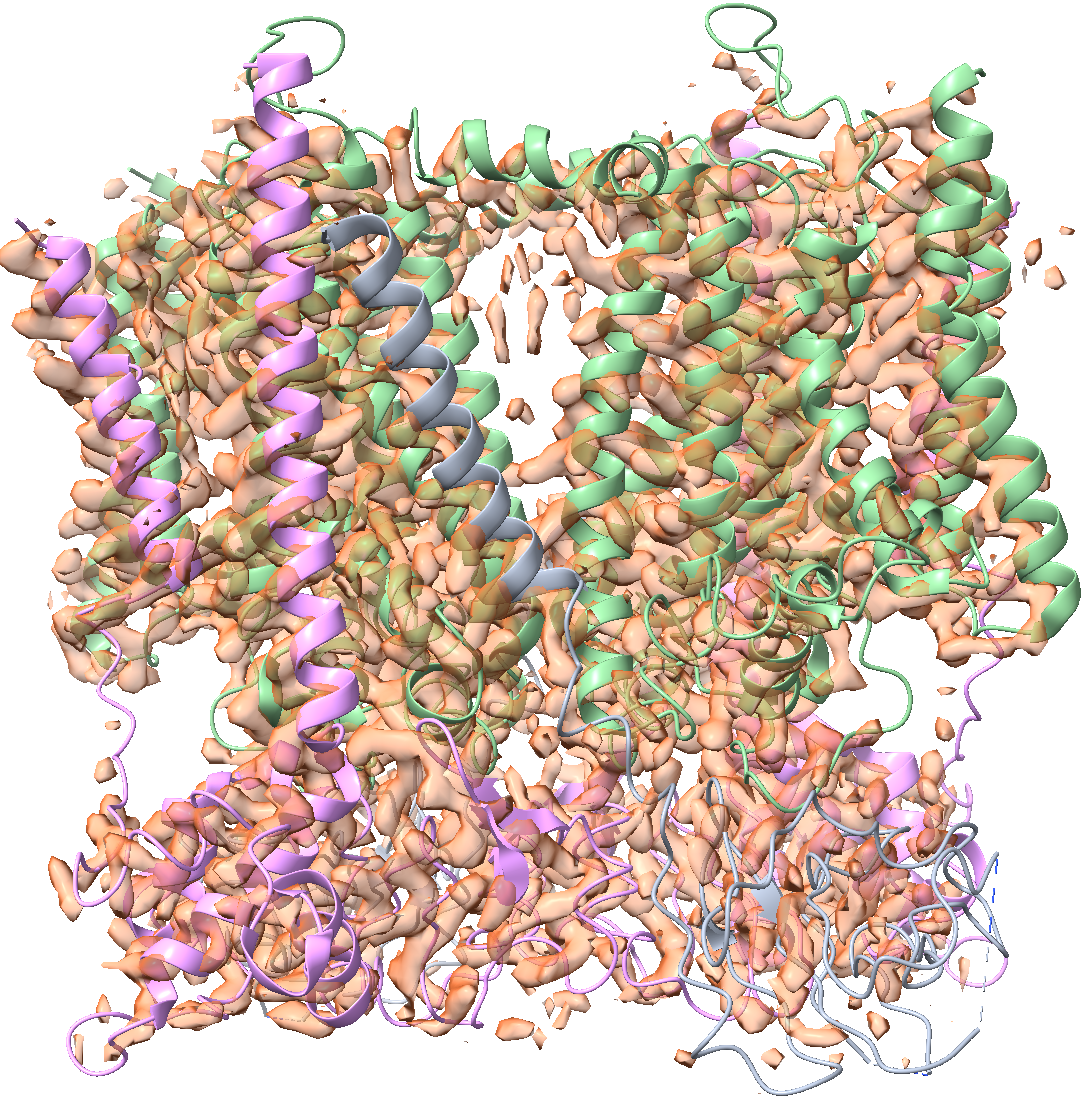}
\end{minipage}

\caption{
Representative Cryo2StructData issues and the effects of curation. Top left: the released normalized map for EMD-0271, visualized in ChimeraX. Top right: the raw density map and deposited structure, PDB 6HTX, for the same entry, showing substantial unmodeled density. These views illustrate the potential brittleness of hard-threshold normalization and the mismatch between density inputs and coordinate supervision in large unmodeled regions, particularly when training crops include density unsupported by the deposited model. Bottom row: EMD-0719/PDB 6KLV, an entry shared by Cryo2StructData and CryoAtlas. The original map--structure pair is poorly aligned; our fitmap correction improves spatial agreement with the experimental density and raises Q-score from 0.10 to 0.65.
}
\label{fig:c2s_examples}
\end{figure}

\paragraph{Findings.}
CryoAtlas and raw EMDB Cryo2StructData exhibit similar overall Q-score distributions, particularly when CryoAtlas is restricted to deposited resolutions $\leq 4\,\text{\AA}$. The normalized Cryo2StructData release instead shifts toward lower Q-scores and has a heavier low-quality tail. In particular, 181 normalized entries have observed Q-scores $<0$, compared with 25 raw-map entries (Table~\ref{tab:c2s_qscore_thresholds}). We also observe map--structure misalignments in which deposited coordinates are spatially inconsistent with the experimental density (Figure~\ref{fig:c2s_examples}).

These findings indicate reduced robustness of the normalized Cryo2StructData release under the ChimeraX Q-score audit. Hard-threshold normalization may remain effective with a specifically adapted training strategy, but is less suitable for density-guided segmentation pipelines such as the ModelAngelo-style setup used here, where coordinates provide voxel-space supervision. CryoAtlas, in comparison, aligns structures to maps and excludes entries with per-EMD mean Q-score below 0.3 under the same audit.

We additionally identify sequence-alignment inconsistencies in several FASTA files, primarily misplaced gap tokens. These appear to arise from chain-pairing or chain-order mismatches during Clustal Omega alignment of full-length FASTA sequences to PDB-extracted sequences. In EMD-0004, for example, gap positions do not correspond to the modeled residues, yielding an inconsistent sequence--structure alignment and potentially noisy residue-level supervision.

\paragraph{Connection to training-time instability.}
Low-Q-score entries, map--structure misalignments, and large unmodeled regions provide a plausible explanation for the loss spikes described in Section~\ref{subsec:c2s_train_stability}. In low-Q-score or misaligned entries, supervised atoms can occupy locations with weak, shifted, or below-background density, creating a direct conflict between the input map and the target labels. Large unmodeled regions introduce the converse inconsistency: the map contains density that is absent from the deposited coordinate model (Figure~\ref{fig:c2s_examples}). These cases affect approximately 5--10\% of entries in our audit and can generate unusually large losses, increasing spike-sensitive RMS more strongly than robust MAD. This pattern is consistent with the loss curves in Figure~\ref{fig:focal_curves} and the smoother training trajectory observed for CryoAtlas.

\clearpage

\section{Supplementary Stratified Performance Figures}
\label{app:stratified_figures}

All plots in this section use the composite scores defined in Appendix~\ref{sec:appendix_metrics}. Phenix is included only as a conditional 24-hour traditional baseline on targets for which it produced a scorable model. Train-set identity analyses include only the four density-guided learning-based methods, because AlphaFold~3 and Phenix do not have comparable method-specific training-set identity bins.

\subsection{Resolution-dependent Performance}
\label{app:Res}

\begin{figure}[H]
\centering
\begin{minipage}[t]{0.48\linewidth}
\centering
\includegraphics[width=\linewidth]{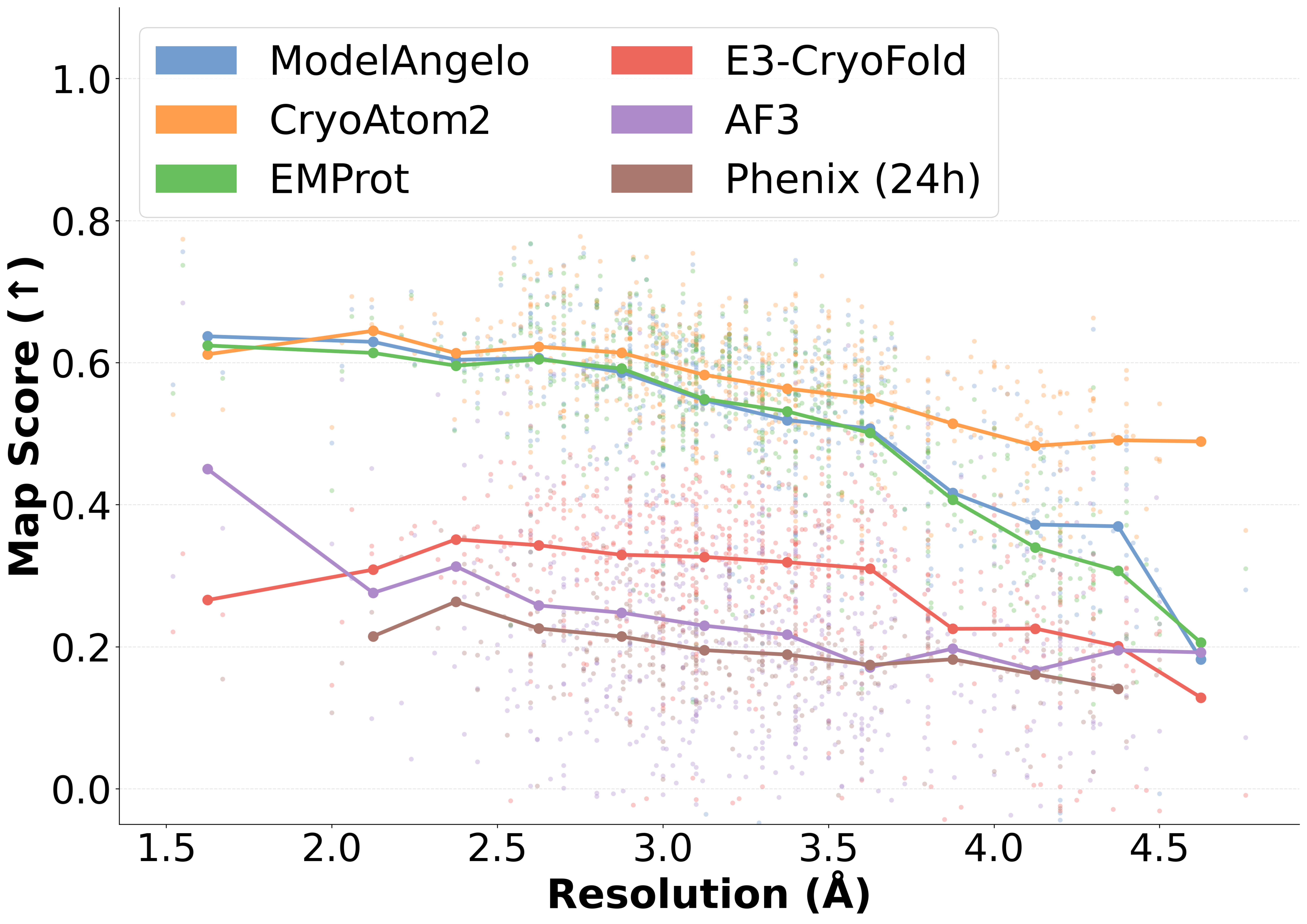}
\end{minipage}
\hfill
\begin{minipage}[t]{0.48\linewidth}
\centering
\includegraphics[width=\linewidth]{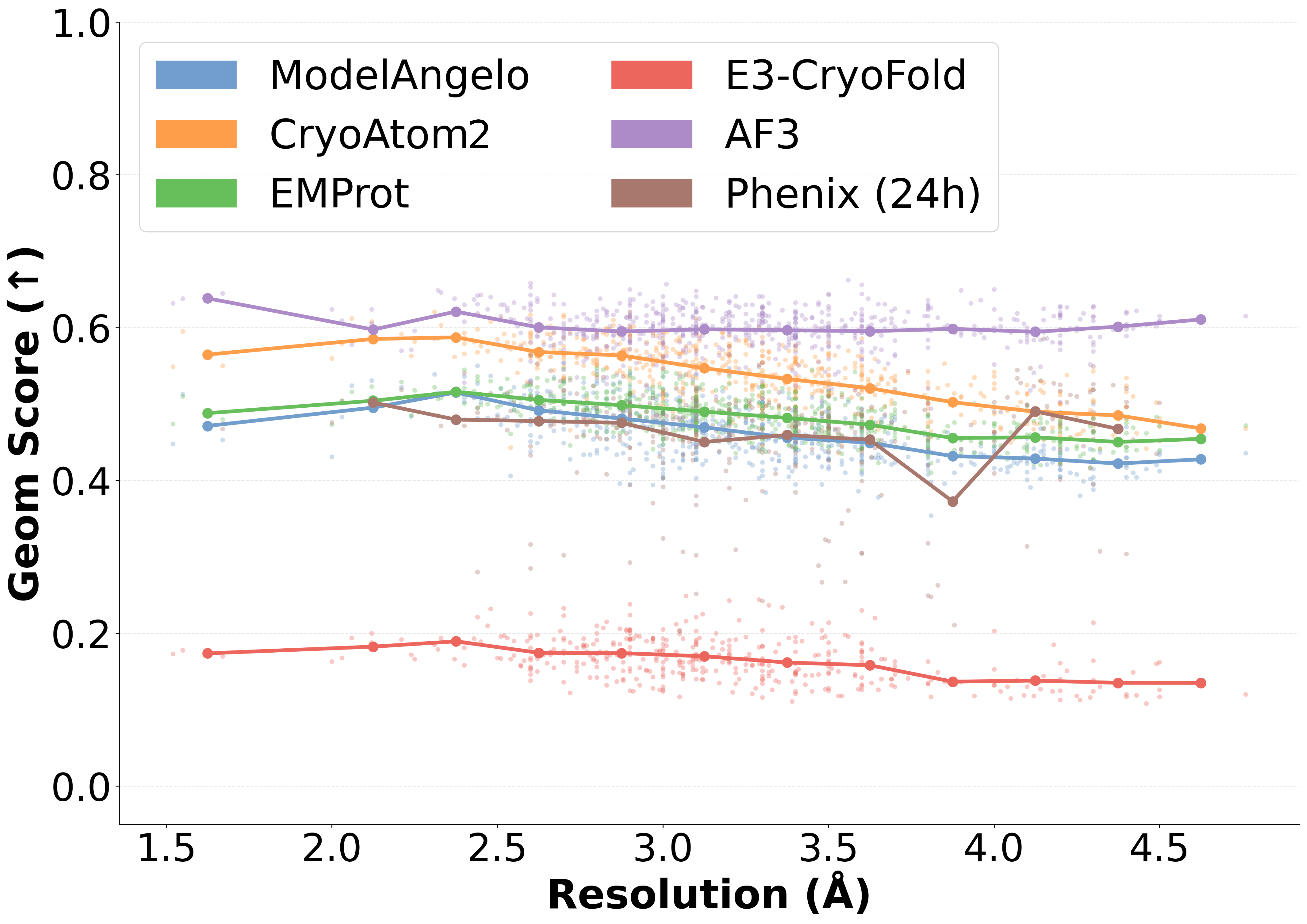}
\end{minipage}
\caption{Map Score and Geom Score by map resolution. Phenix is shown only for completed 24-hour targets.}
\label{fig:res_map_geom}
\end{figure}

\subsection{Length-dependent Performance}
\label{app:length}

\begin{figure}[H]
\centering
\begin{minipage}[t]{0.48\linewidth}
\centering
\includegraphics[width=\linewidth]{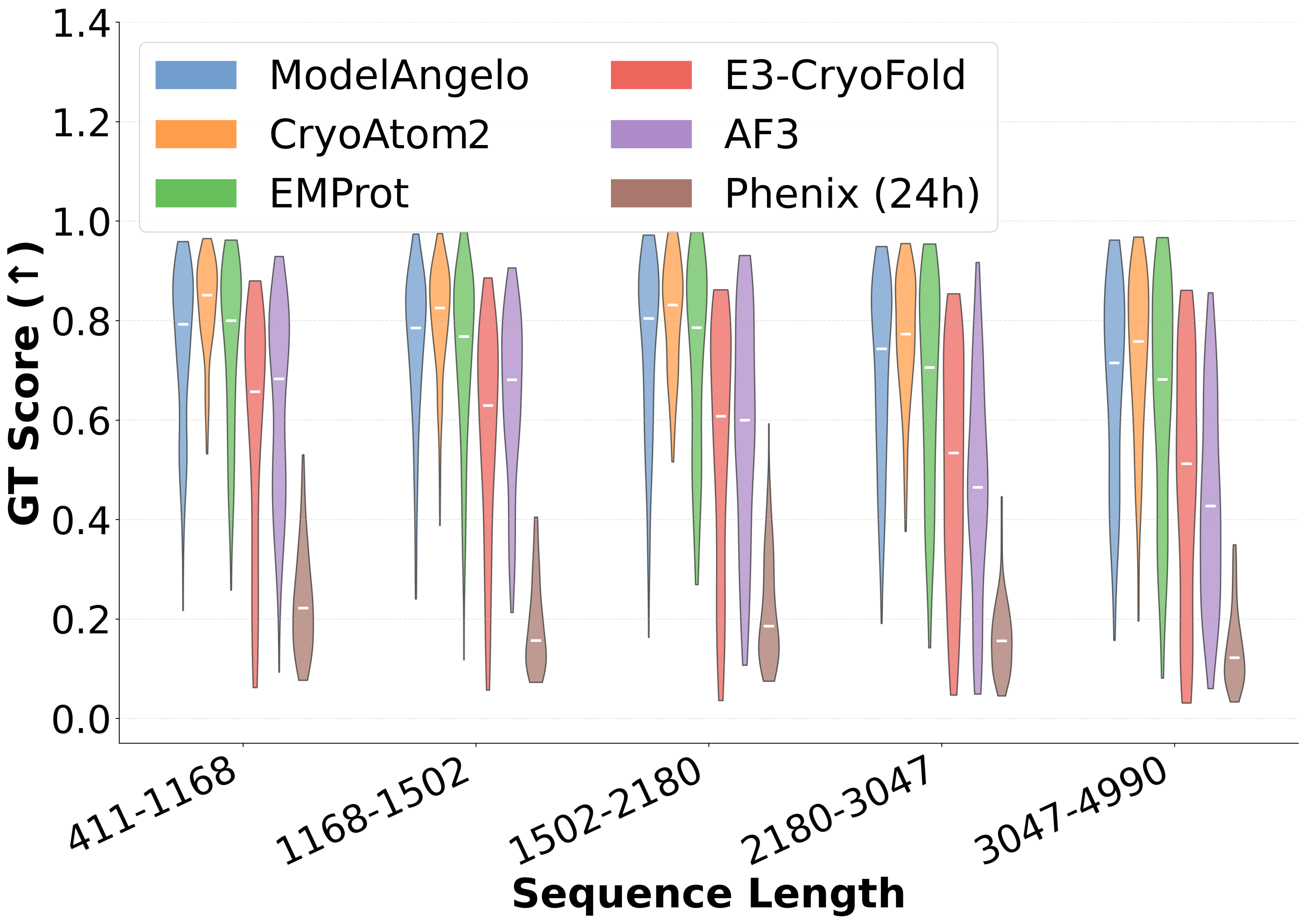}
\end{minipage}
\hfill
\begin{minipage}[t]{0.48\linewidth}
\centering
\includegraphics[width=\linewidth]{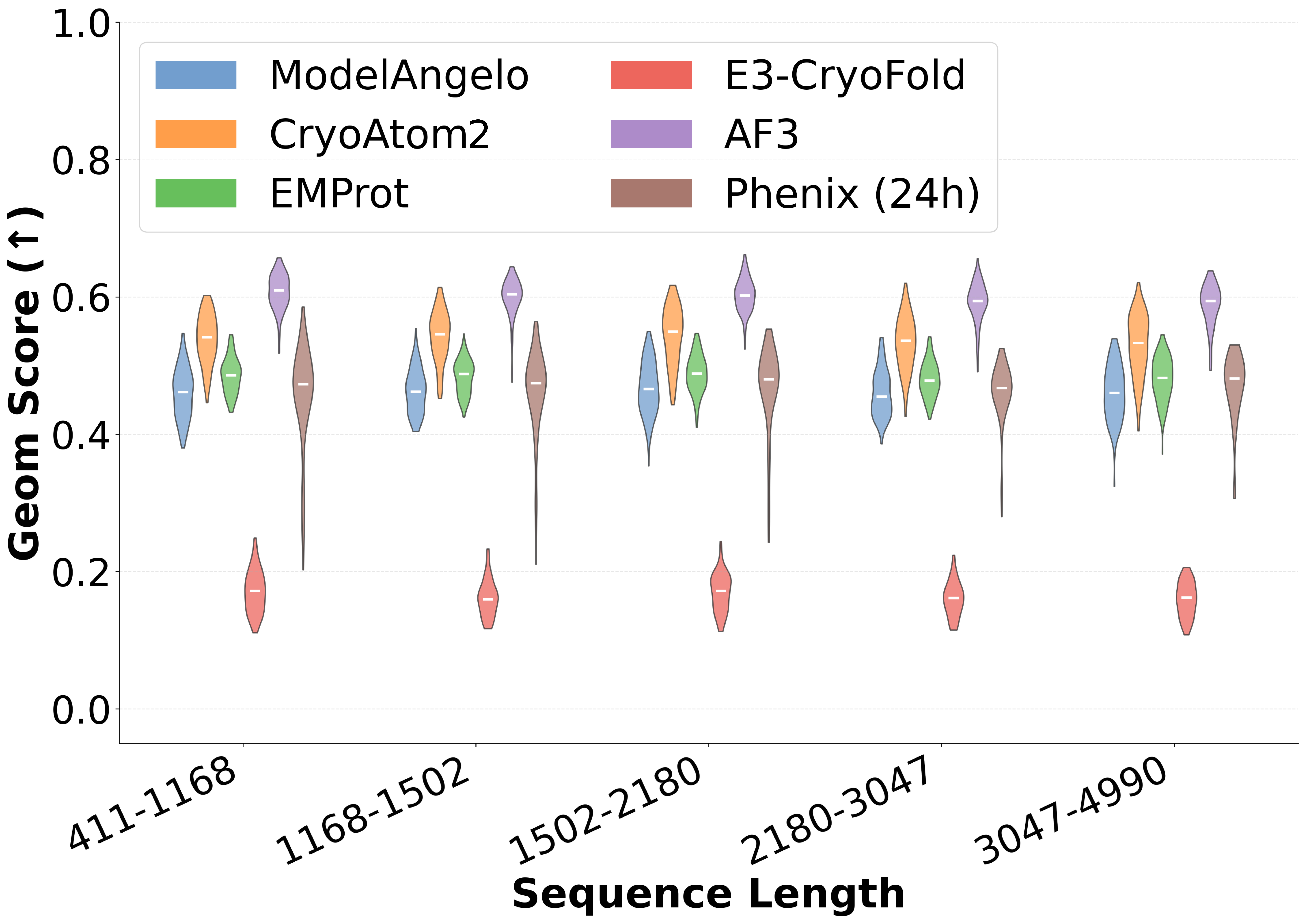}
\end{minipage}
\caption{GT Score and Geom Score by sequence length. Phenix is shown only for completed 24-hour targets.}
\label{fig:length_gt_geom}
\end{figure}

\subsection{Protein-category Effects}
\label{app:category}

\begin{figure}[H]
    \centering
    \includegraphics[width=0.65\textwidth]{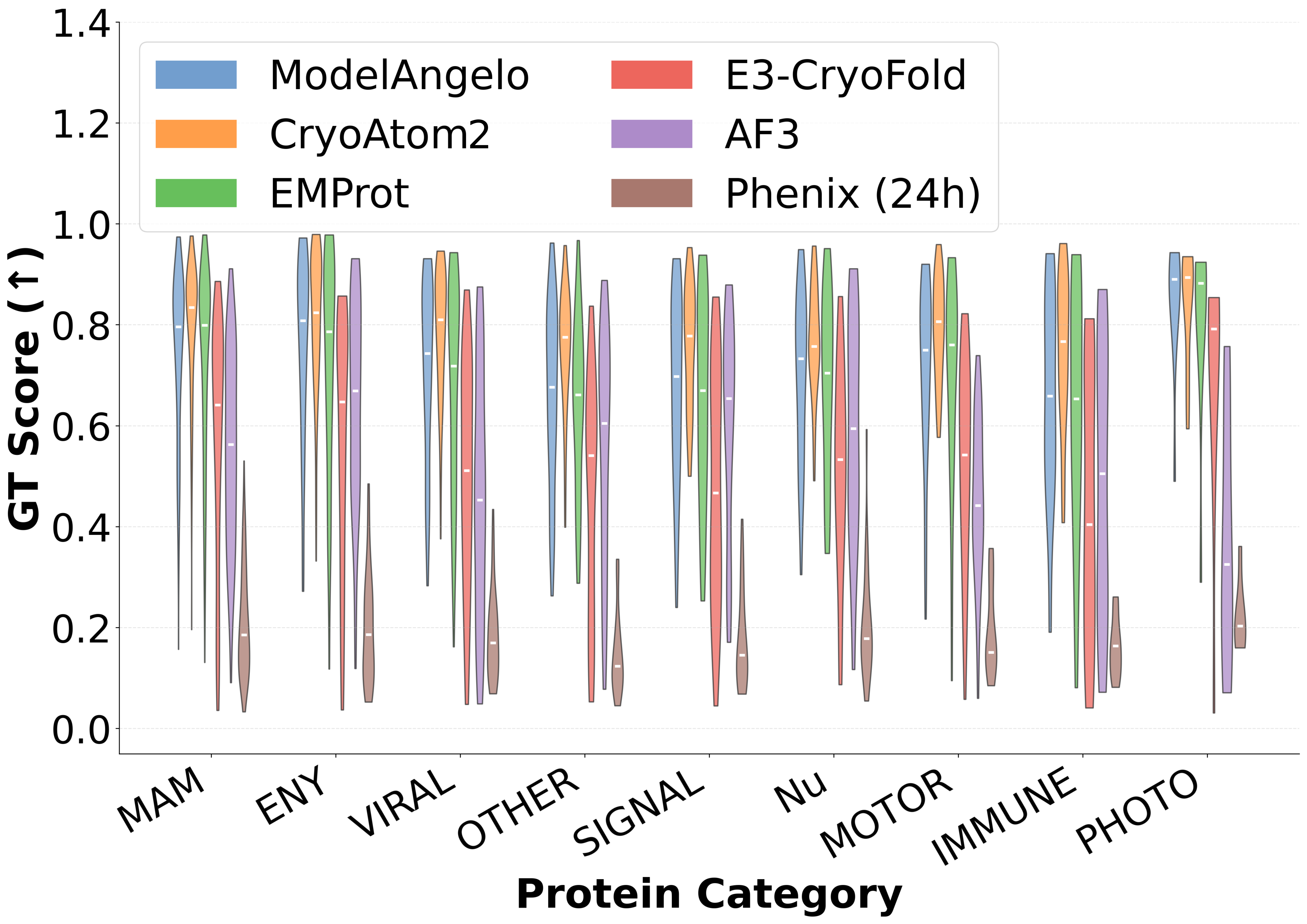}
    
    \vspace{1em} 
    \includegraphics[width=0.65\textwidth]{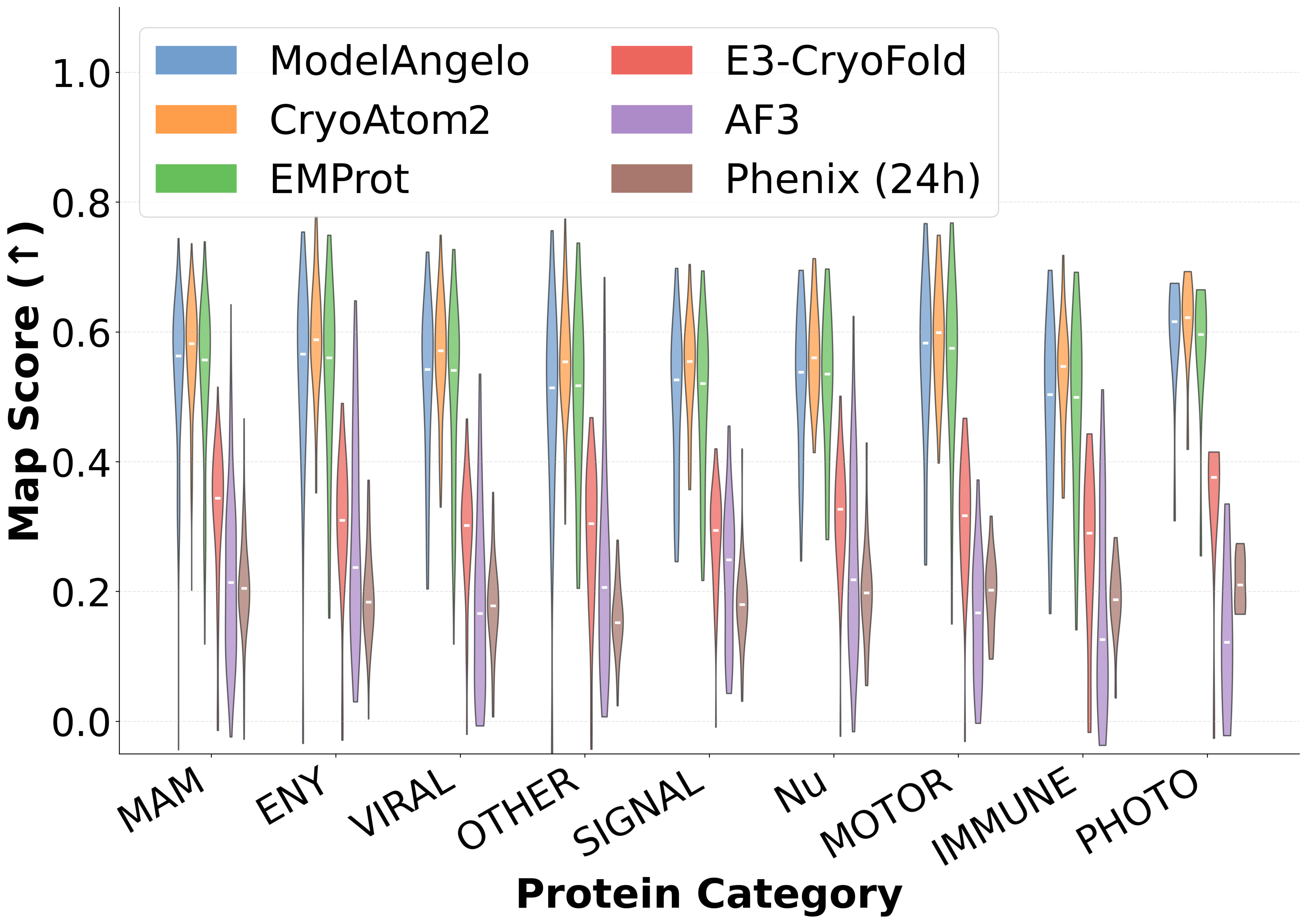}
    
    \vspace{1em}
    \includegraphics[width=0.65\textwidth]{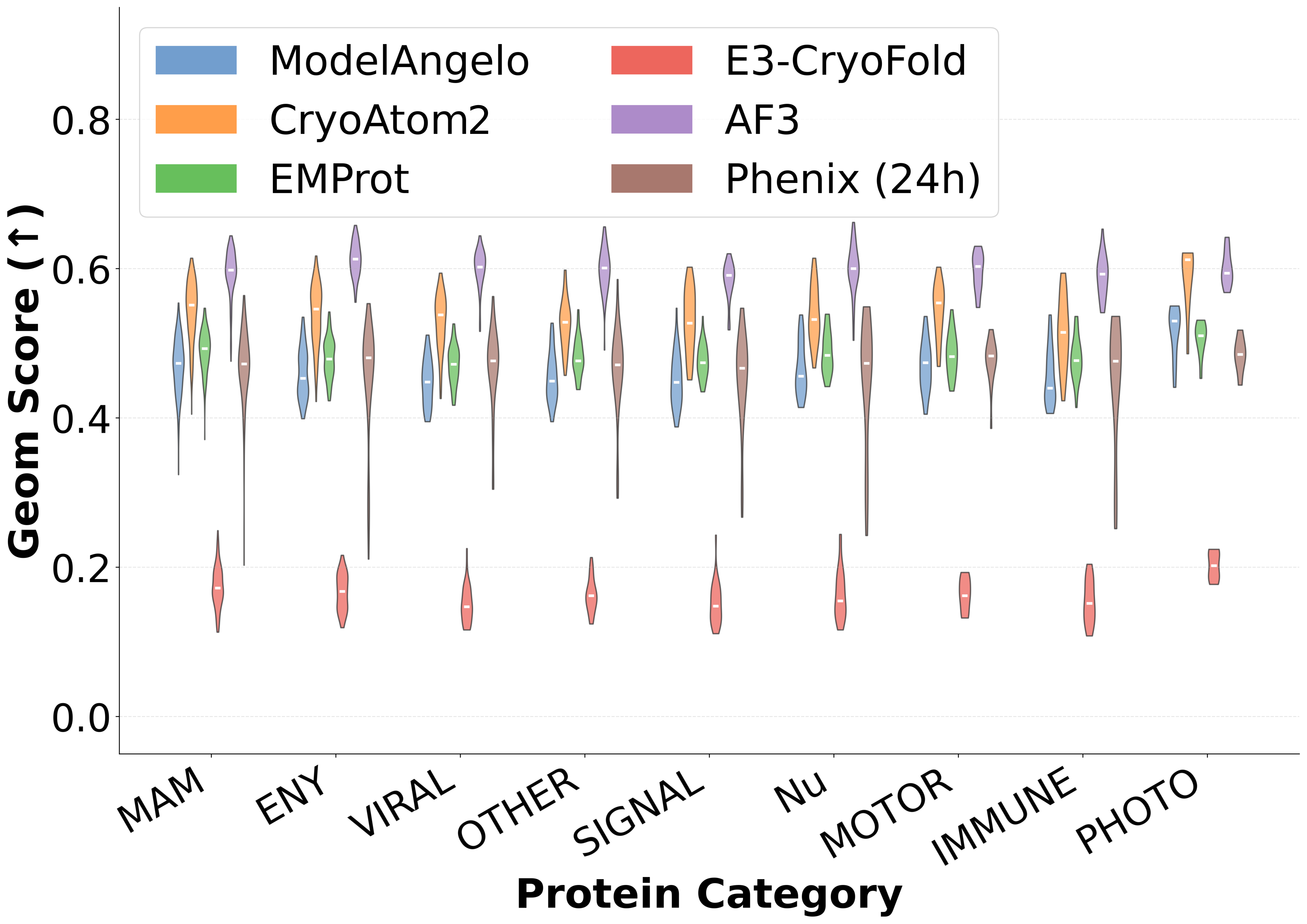}
    
    \caption{GT Score, Map Score, and Geom Score across nine PDB-keyword protein categories.}
    \label{fig:category_scores}
\end{figure}
\subsection{Test--Train Sequence Similarity}
\label{app:identity}

\begin{figure}[H]
    \centering
    \includegraphics[width=0.65\textwidth]{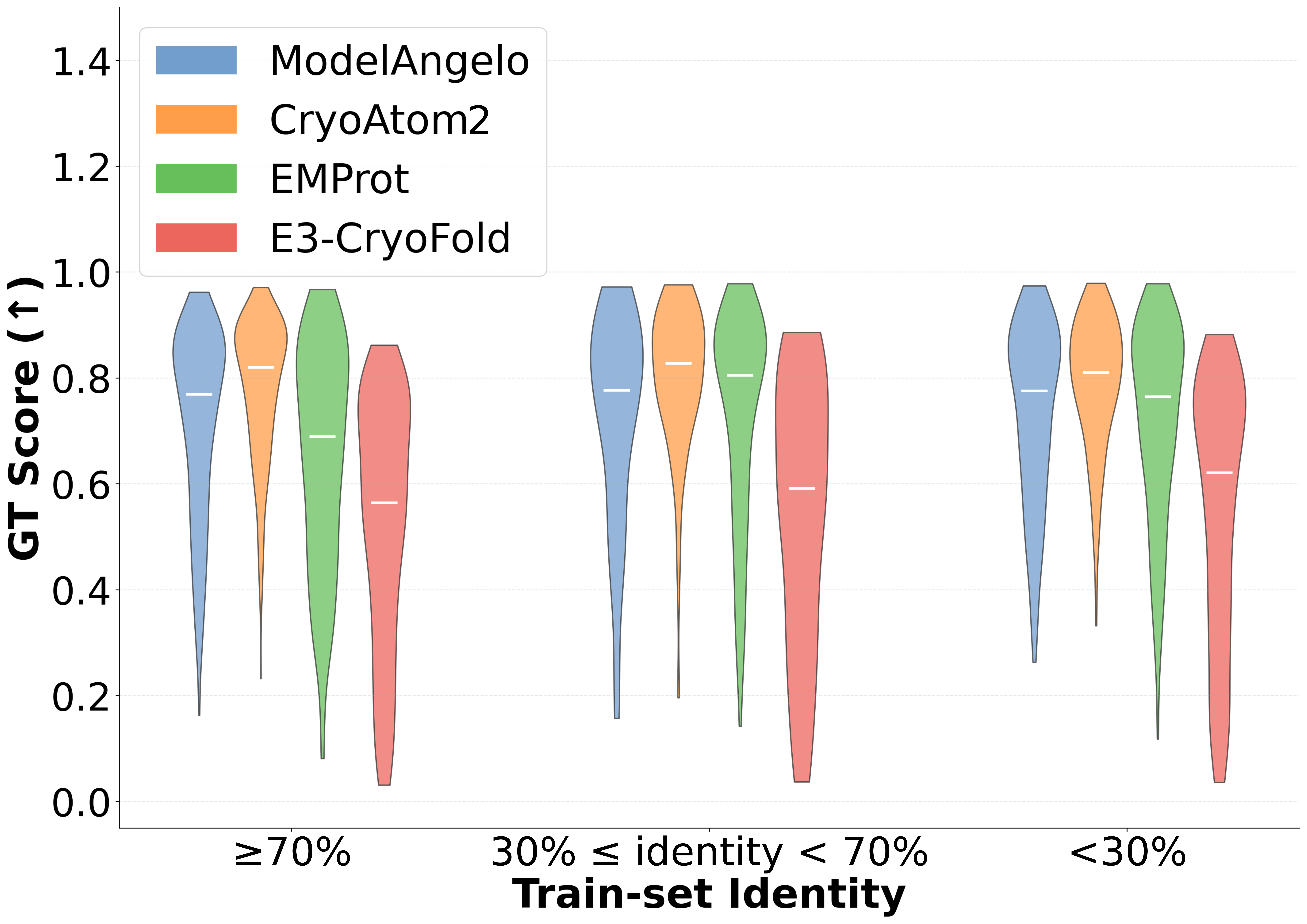}
    
    \vspace{1em}
    \includegraphics[width=0.65\textwidth]{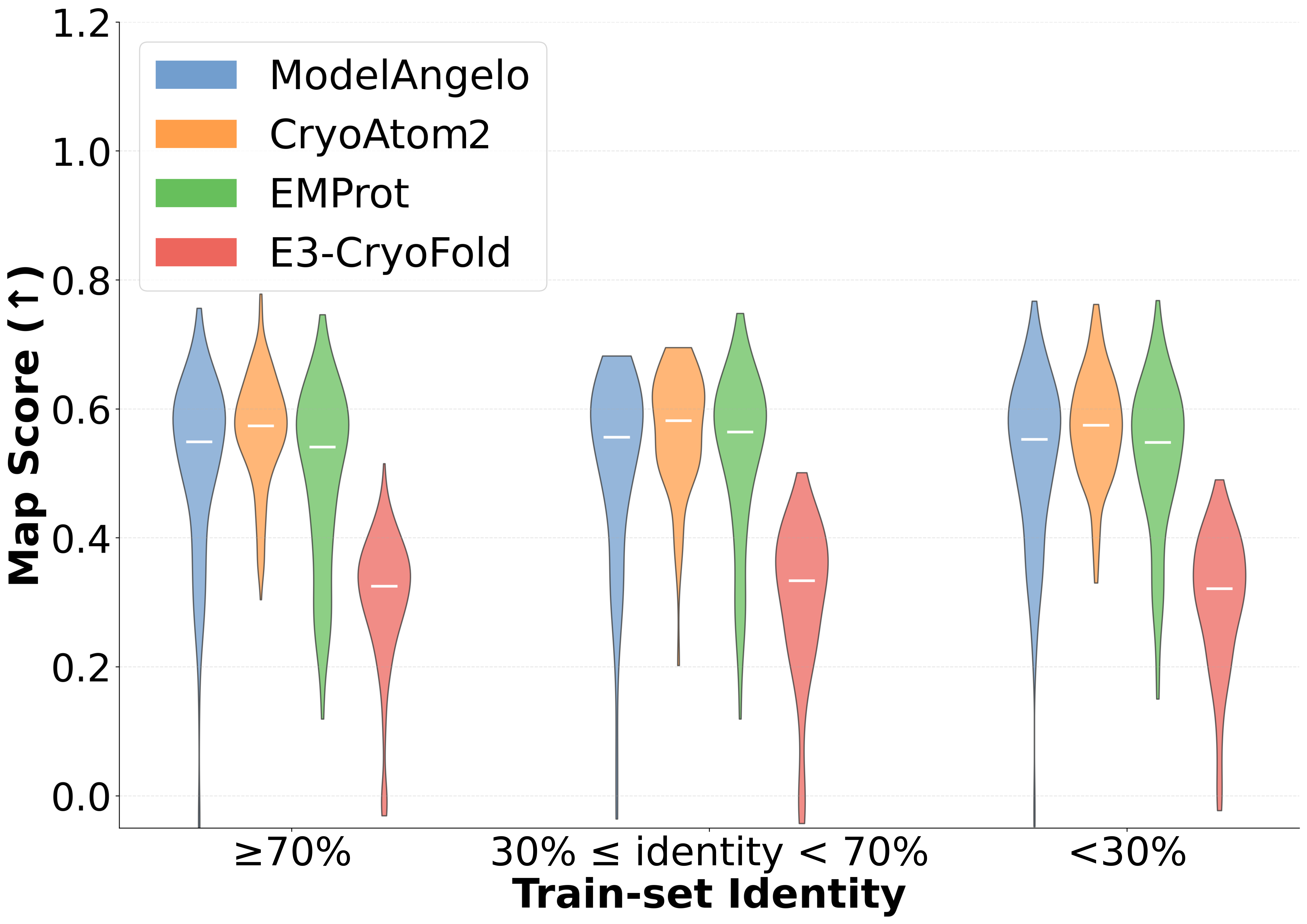}
    
    \vspace{1em}
    \includegraphics[width=0.65\textwidth]{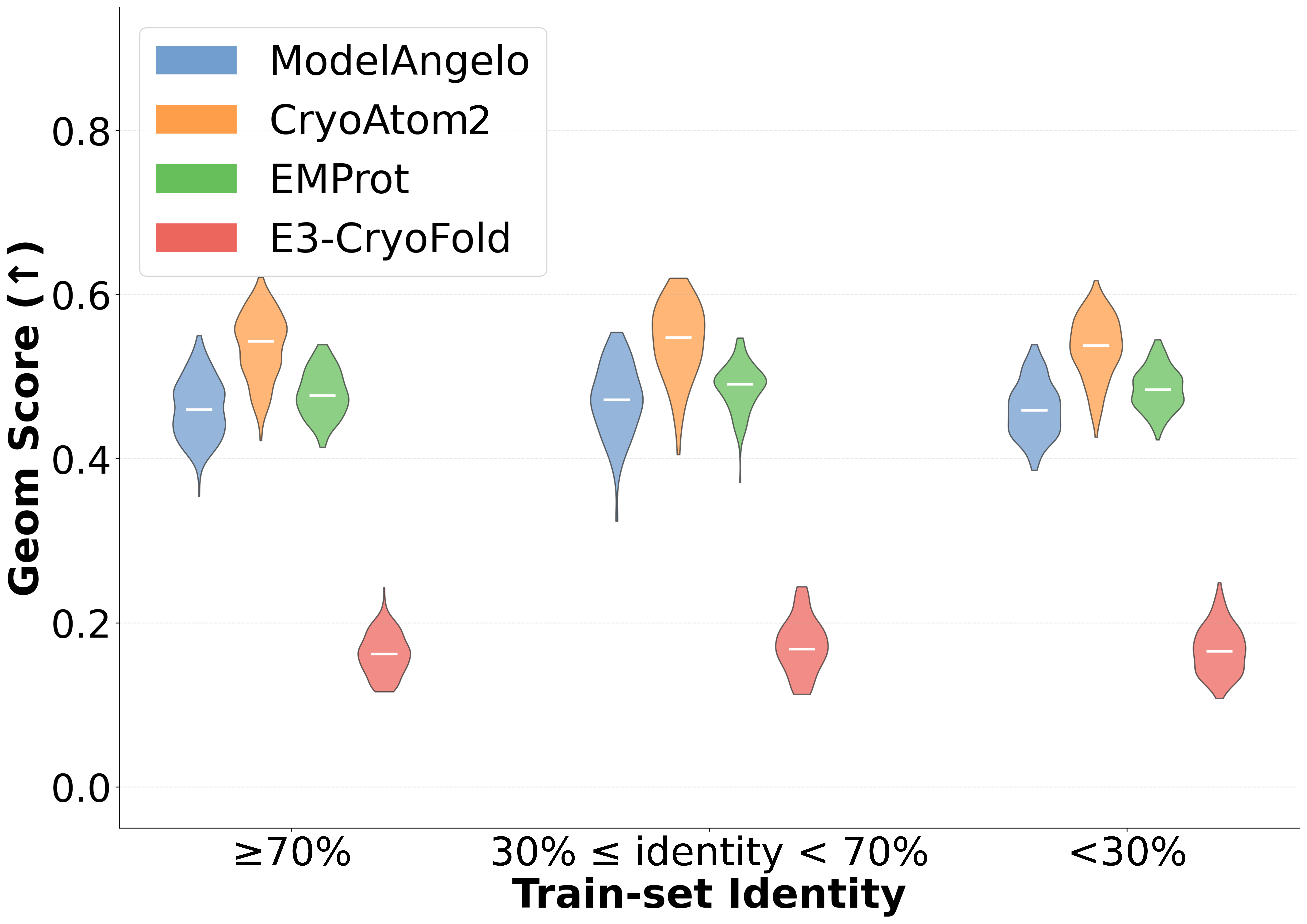}
    
    \caption{GT Score, Map Score, and Geom Score as a function of maximum test--train sequence identity. AlphaFold~3 and Phenix are excluded from this analysis.}
    \label{fig:identity_scores}
\end{figure}

\section{Failure Cases of AlphaFold~3}
\label{sec:AF3_failure}

\begin{figure}[H]
\centering

\begin{minipage}[t]{0.48\linewidth}
    \centering
    \includegraphics[width=\linewidth]{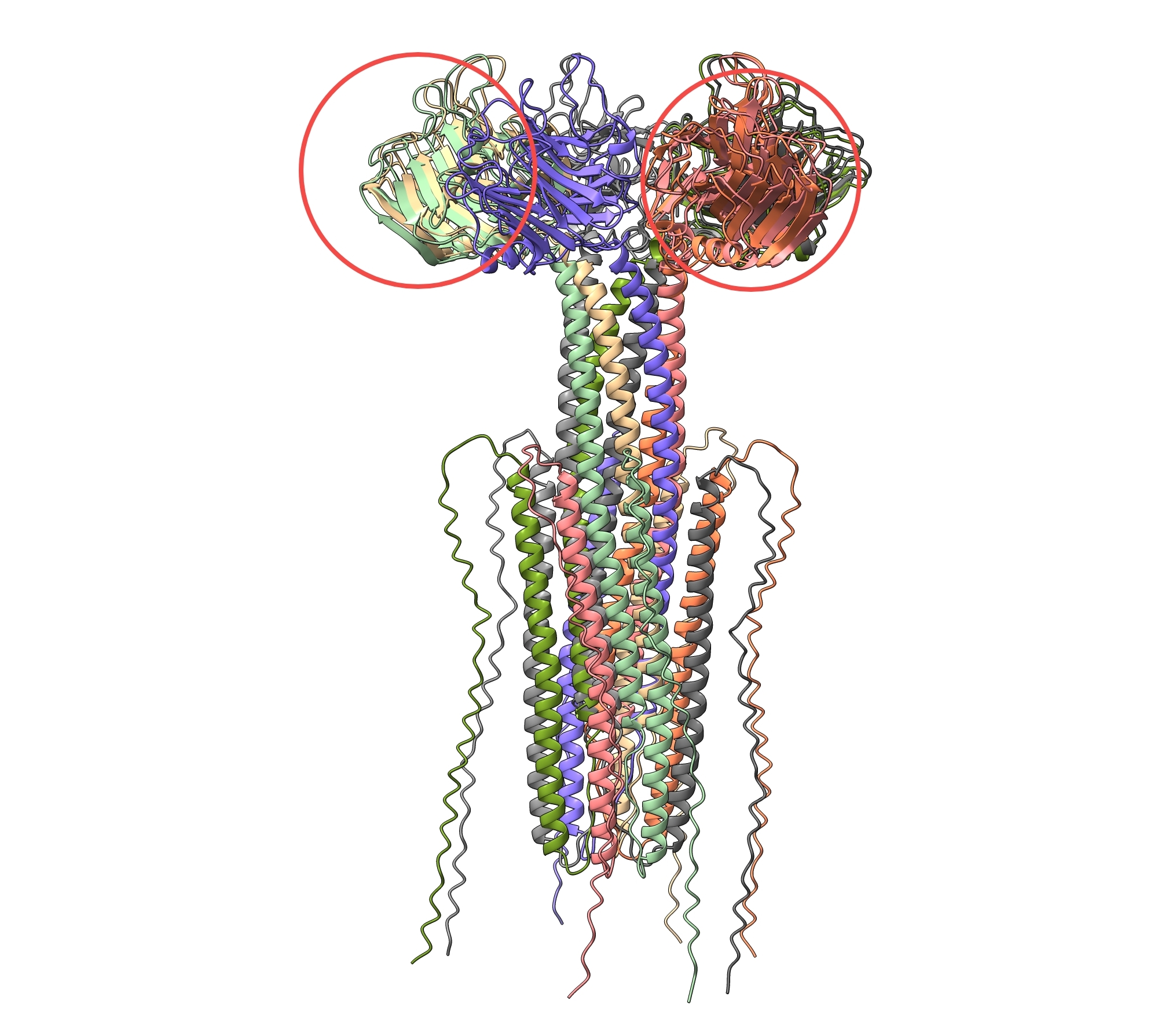}
\end{minipage}
\hfill
\begin{minipage}[t]{0.48\linewidth}
    \centering
    \includegraphics[width=\linewidth]{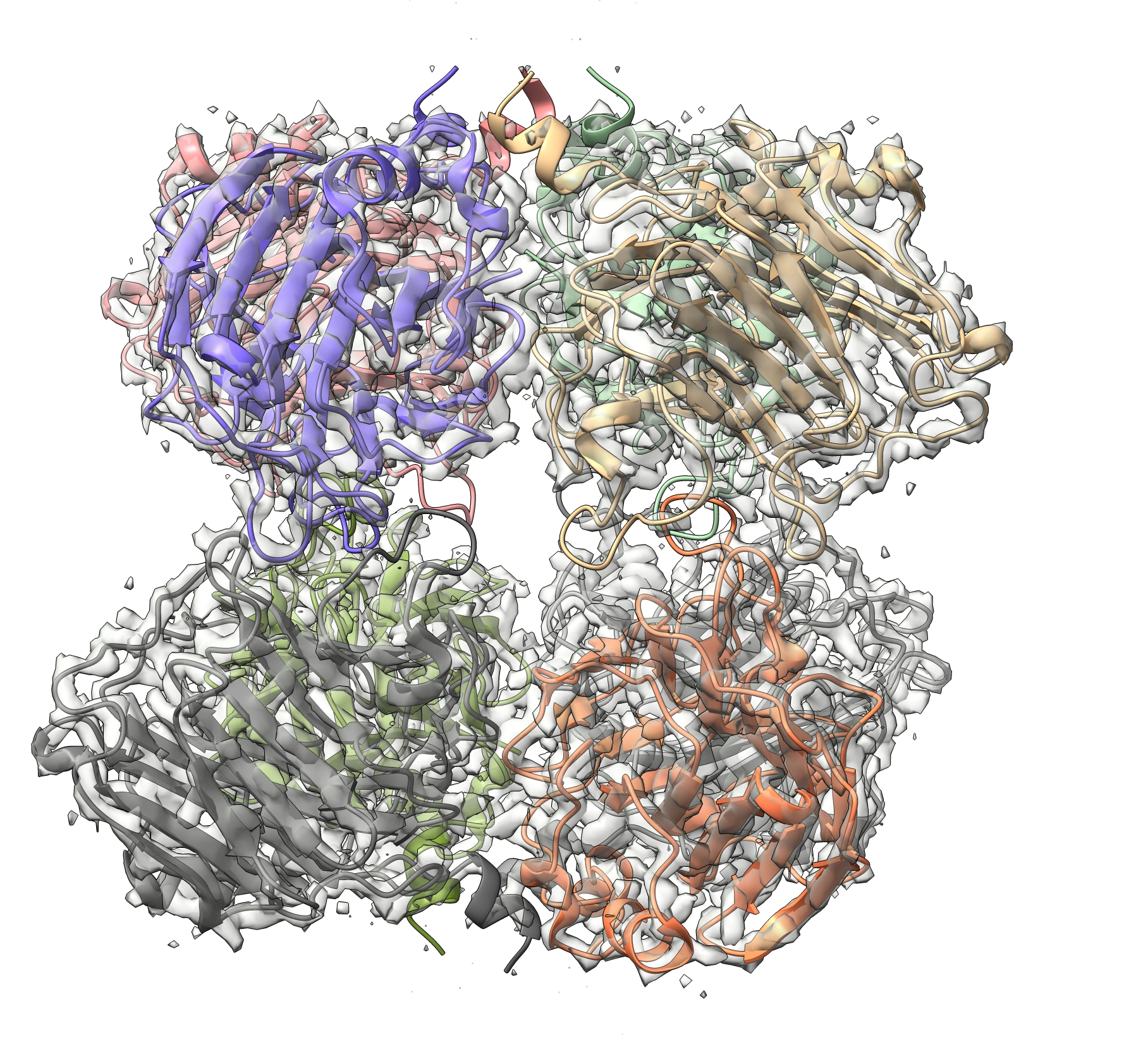}
\end{minipage}

\vspace{2mm}

\begin{minipage}[t]{0.48\linewidth}
    \centering
    \includegraphics[width=\linewidth]{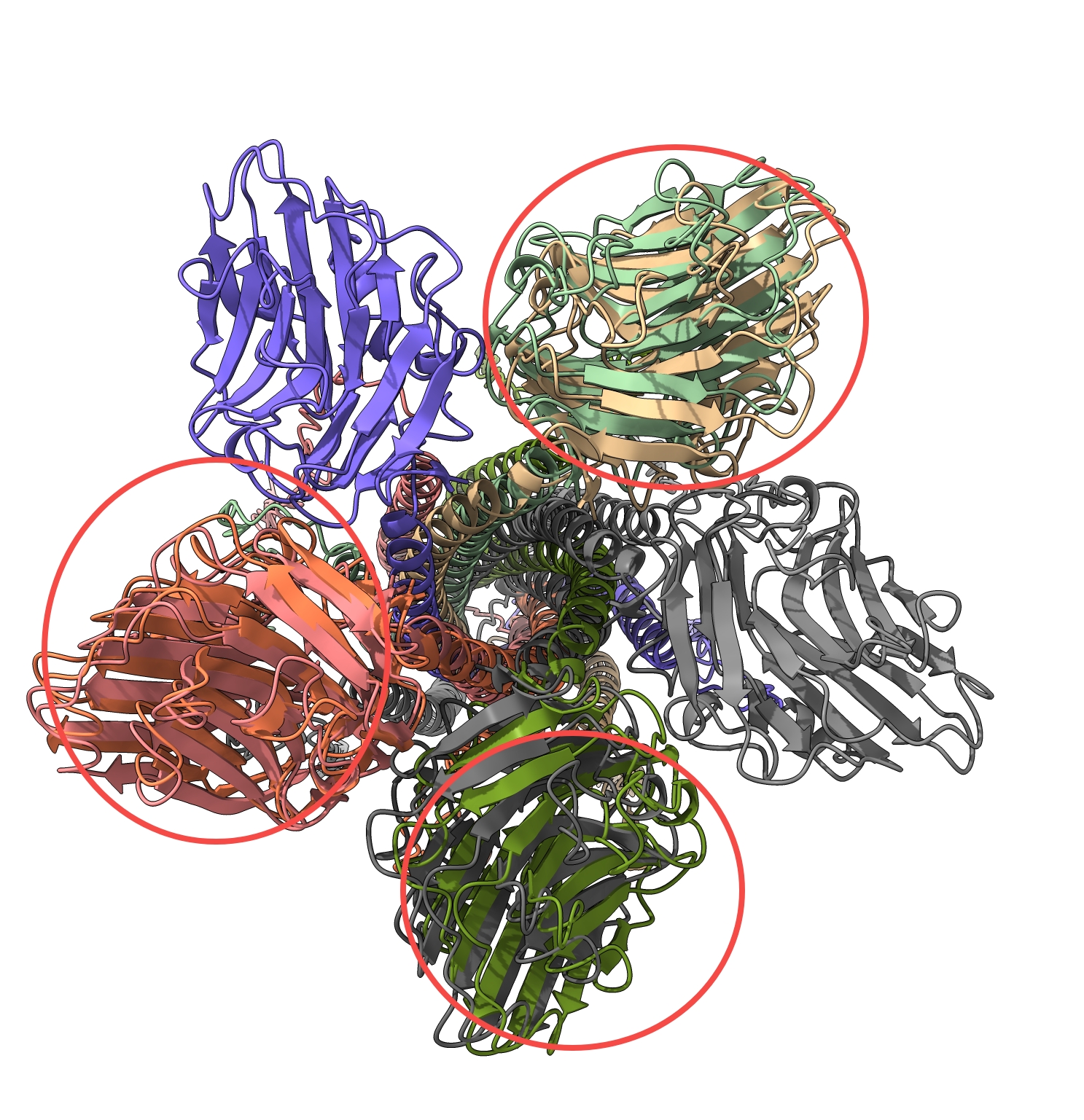}
\end{minipage}
\hfill
\begin{minipage}[t]{0.48\linewidth}
    \centering
    \includegraphics[width=\linewidth]{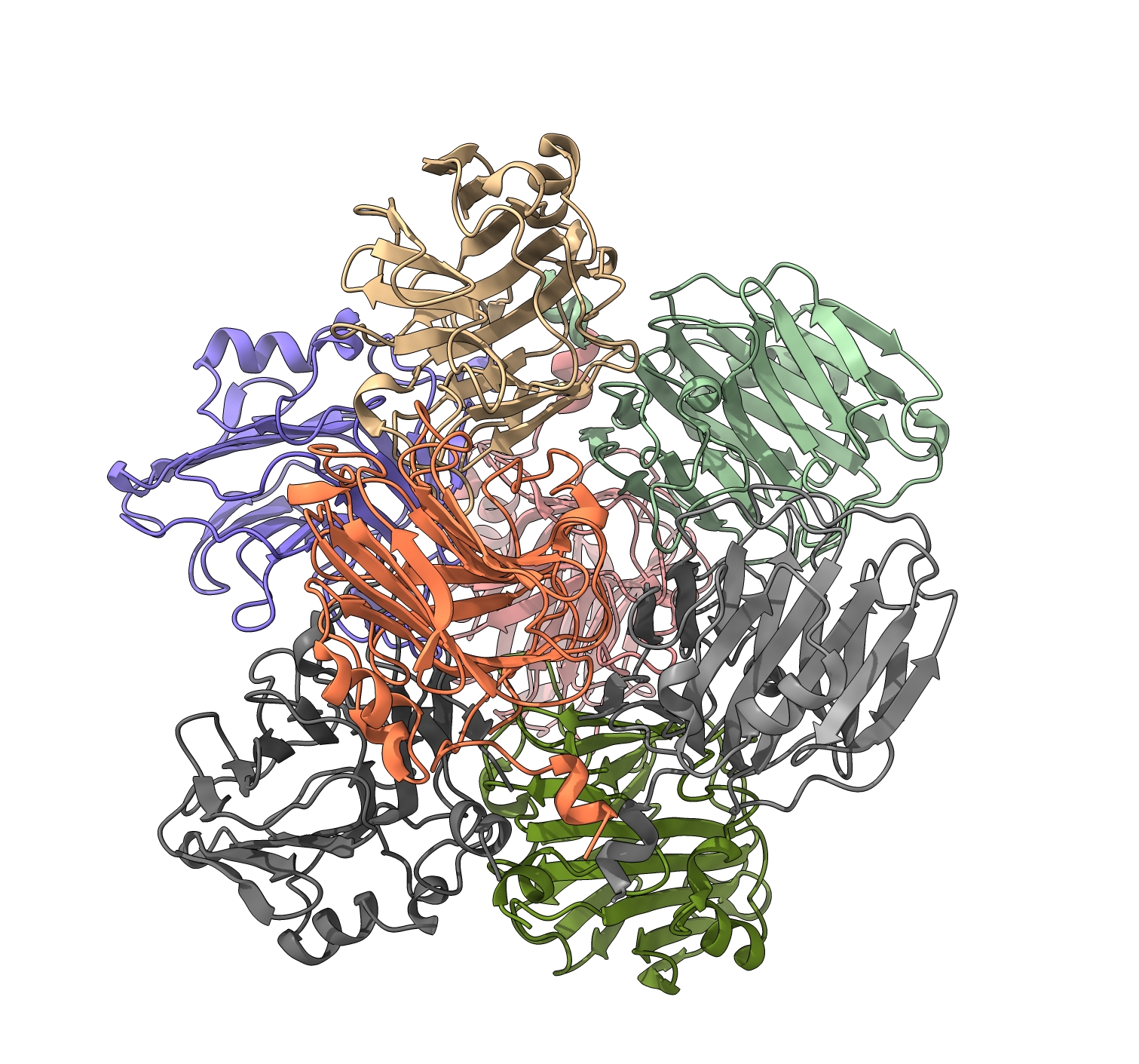}
\end{minipage}

\caption{
Assembly failure of AlphaFold~3 on EMD-14774 / PDB 7ZL1. Left: the prediction in two views. Right: corresponding views of the deposited structure. The red circle highlights overlapping predicted subunits.
}
\label{fig:af3_failure_examples}
\end{figure}

\paragraph{Observed failures.}
Of the 947 targets with fewer than 5,000 amino acids in the 1,152-map test pool, 935 met our AF3 completion criterion. Seven targets failed on the server, and five yielded structurally incorrect outputs: PDB 7SGD / EMD-25107, PDB 7ZL1 / EMD-14774, PDB 8IHN / EMD-35450, PDB 8ZK0 / EMD-60162, and PDB 9LHO / EMD-63103. In 7ZL1, the predicted subunits form a distorted, pentamer-like arrangement with substantial overlap, differing from the deposited octamer (Figure~\ref{fig:af3_failure_examples}). The resulting steric clashes trigger the clash penalty in the AF3 ranking score, yielding a negative value~\cite{AF3}.

\paragraph{Sequence scope and chain correspondence.}
The full-length FASTA for 7ZL1 includes residues absent from the deposited coordinate model. Reconciling the input sequences with experimentally resolved regions requires explicit sequence--structure mapping, including chain multiplicity and residue coverage. This correspondence becomes particularly important for assemblies containing identical or closely homologous subunits, where similar sequences must be organized into the correct spatial arrangement. Structural validation should account for permutations of equivalent chains while assessing subunit placement, interchain contacts, and chain connectivity.

\paragraph{Training context and assembly priors.}
AF3 uses 384-token training crops followed by fine-tuning with 640- and 768-token crops, with larger crops improving interface accuracy~\cite{AF3}. For assemblies substantially larger than these windows, individual training examples provide only partial context for learning global interchain organization. Limited coverage of the relevant large oligomeric architectures may further contribute to assembly errors. Repeated subunits pose a particular challenge: accurate local folds must be combined with consistent interfaces and long-range assembly geometry. The whole-chain overlaps reported for AF3 homomers illustrate this distinction between local structural quality and global assembly accuracy~\cite{AF3}.

\paragraph{Implications for density-guided modeling.}
A plausible fold for each subunit does not ensure that the complete assembly explains the map. For repeated subunits, similar local density can accommodate several placements, while sequence priors may favor contacts or conformations that differ from the observed structure. Correcting these errors requires map information to influence interchain organization as well as local coordinates. The density-conditioned methods discussed in Section~\ref{subsec:sequence_and_model_building} approach this problem through sampling guidance, embedding optimization, or learned map features. For the failure modes illustrated here, the key tests are whether a method recovers the spatial arrangement and sequence correspondence of the subunits, resolves clashes, and fits the density without introducing incorrect chain connections.

\clearpage
\section{Dataset Composition and Curation Analysis}
\label{app:dataset_composition}

\subsection{Dataset groups and composition}
\label{app:dataset_groups}
We compare candidate, curated, and rejected entries, together with the test pool and common benchmark. Of the 21,229 candidate entries, 17,809 are retained and 3,420 rejected; the retained entries comprise 16,657 training entries and a 1,152-map test pool, from which the 618-target common benchmark is drawn. Tables~\ref{tab:dataset_resolution}--\ref{tab:dataset_length} summarize their resolution, taxonomy, protein chain count, map size, and total protein sequence length.

\begin{table}[H]
\centering
\footnotesize
\setlength{\tabcolsep}{3pt}
\renewcommand{\arraystretch}{1.12}
\begin{tabular*}{0.94\textwidth}{@{\extracolsep{\fill}}ccccccc@{}}
\toprule
\textbf{Resolution (\AA)} & \textbf{Candidate} & \textbf{Curated} & \textbf{Test pool} & \textbf{Benchmark} & \textbf{Rejected} & \shortstack{\textbf{Q-score} \textbf{removal}} \\
\midrule
$\leq2.5$ & 8.0 & 8.3 & 6.2 & 4.4 & 6.5 & 1.6 \\
2.5--3.0 & 25.8 & 27.3 & 24.0 & 28.6 & 18.0 & 1.3 \\
3.0--3.5 & 35.9 & 37.3 & 35.2 & 38.8 & 29.0 & 1.5 \\
3.5--4.0 & 19.6 & 19.2 & 22.1 & 15.9 & 21.6 & 3.8 \\
4.0--4.5 & 7.0 & 6.1 & 10.8 & 11.8 & 11.7 & 9.1 \\
4.5--5.0 & 2.1 & 1.2 & 1.1 & 0.2 & 6.4 & 21.6 \\
5.0--5.5 & 0.8 & 0.3 & 0.4 & 0.2 & 3.2 & 27.0 \\
5.5--6.0 & 0.8 & 0.2 & 0.3 & 0.2 & 3.7 & 40.5 \\
\bottomrule
\end{tabular*}
\vspace{2mm}
\caption{Resolution distributions (\%) across the five dataset groups. The final column reports the percentage of candidates removed by Q-score within each resolution bin.}
\label{tab:dataset_resolution}
\end{table}

\begin{table}[H]
\centering
\footnotesize
\setlength{\tabcolsep}{3pt}
\renewcommand{\arraystretch}{1.12}
\begin{tabular*}{0.94\textwidth}{@{\extracolsep{\fill}}cccccc@{}}
\toprule
\textbf{Taxonomic group} & \textbf{Candidate} & \textbf{Curated} & \textbf{Test pool} & \textbf{Benchmark} & \textbf{Rejected} \\
\midrule
Eukaryota & 53.3 & 56.9 & 55.4 & 56.1 & 34.8 \\
Bacteria & 16.6 & 17.0 & 21.3 & 18.8 & 14.2 \\
Viruses & 10.9 & 7.2 & 7.0 & 5.7 & 30.5 \\
Mixed & 17.8 & 17.7 & 14.8 & 18.4 & 18.1 \\
Archaea & 1.0 & 0.9 & 0.9 & 0.3 & 1.6 \\
Other & 0.4 & 0.3 & 0.7 & 0.6 & 0.8 \\
\bottomrule
\end{tabular*}
\vspace{2mm}
\caption{Taxonomic composition of the five dataset groups.}
\label{tab:dataset_taxonomy}
\end{table}

\begin{table}[H]
\centering
\footnotesize
\setlength{\tabcolsep}{3pt}
\renewcommand{\arraystretch}{1.12}
\begin{tabular*}{0.94\textwidth}{@{\extracolsep{\fill}}cccccc@{}}
\toprule
\textbf{Protein chains} & \textbf{Candidate} & \textbf{Curated} & \textbf{Test pool} & \textbf{Benchmark} & \textbf{Rejected} \\
\midrule
1 & 8.7 & 7.5 & 6.1 & 6.7 & 14.8 \\
2 & 13.8 & 14.6 & 12.3 & 15.7 & 10.0 \\
3--4 & 27.0 & 26.9 & 25.2 & 30.8 & 27.4 \\
5--8 & 27.0 & 28.1 & 28.4 & 31.8 & 21.3 \\
9--16 & 12.9 & 12.6 & 14.3 & 11.2 & 14.1 \\
$>16$ & 10.7 & 10.3 & 13.7 & 3.7 & 12.4 \\
\bottomrule
\end{tabular*}
\vspace{2mm}
\caption{Distribution of protein chain counts per entry.}
\label{tab:dataset_chains}
\end{table}

\begin{table}[H]
\centering
\footnotesize
\setlength{\tabcolsep}{3pt}
\renewcommand{\arraystretch}{1.12}
\begin{tabular*}{0.94\textwidth}{@{\extracolsep{\fill}}cccccc@{}}
\toprule
\textbf{Map size (MiB)} & \textbf{Candidate} & \textbf{Curated} & \textbf{Test pool} & \textbf{Benchmark} & \textbf{Rejected} \\
\midrule
$\leq64$ & 15.7 & 15.7 & 19.2 & 17.6 & 15.8 \\
64--128 & 29.7 & 31.8 & 32.8 & 34.3 & 18.7 \\
128--256 & 34.8 & 35.5 & 31.0 & 31.9 & 31.1 \\
256--512 & 10.1 & 9.5 & 9.8 & 9.9 & 12.9 \\
512--1{,}024 & 7.8 & 6.5 & 5.9 & 5.0 & 14.8 \\
$>1{,}024$ & 1.9 & 1.0 & 1.3 & 1.3 & 6.8 \\
\bottomrule
\end{tabular*}
\vspace{2mm}
\caption{Distribution of map file sizes.}
\label{tab:dataset_map_size}
\end{table}

\begin{table}[H]
\centering
\footnotesize
\setlength{\tabcolsep}{3pt}
\renewcommand{\arraystretch}{1.12}
\begin{tabular*}{0.94\textwidth}{@{\extracolsep{\fill}}cccccc@{}}
\toprule
\shortstack{\textbf{Sequence length} \textbf{(residues)}} & \textbf{Candidate} & \textbf{Curated} & \textbf{Test pool} & \textbf{Benchmark} & \textbf{Rejected} \\
\midrule
$\leq300$ & 0.7 & 0.1 & 0.1 & 0 & 3.8 \\
301--600 & 4.2 & 3.0 & 2.3 & 2.8 & 10.7 \\
601--1{,}000 & 8.8 & 7.3 & 7.6 & 10.1 & 16.8 \\
1{,}001--2{,}000 & 32.1 & 33.6 & 34.8 & 43.5 & 24.1 \\
2{,}001--5{,}000 & 37.6 & 40.1 & 37.4 & 43.7 & 24.7 \\
$>5{,}000$ & 16.6 & 15.9 & 17.8 & 0 & 19.9 \\
\bottomrule
\end{tabular*}
\vspace{2mm}
\caption{Distribution of total protein sequence length per entry.}
\label{tab:dataset_length}
\end{table}

\subsection{Filtering and benchmark selection}
\label{app:filtering_selection}
Of the 3,420 rejected candidates, 994 (29.1\%) were removed by Q-score filtering and 2,426 by unmodeled-density filtering. Q-score removal increases from 1.6\% at $\leq2.5$\,\AA{} to 40.5\% at 5.5--6.0\,\AA{} (Table~\ref{tab:dataset_resolution}), whereas rates across taxonomic groups range from 2.2\% to 5.8\%. This resolution dependence reflects atomic resolvability: filtering improves the reliability of atom-level supervision but reduces coverage of lower-resolution maps within the 6\,\AA{} inclusion limit.

The common benchmark further favors smaller assemblies: targets with more than 16 protein chains represent 3.7\% of its entries, compared with 13.7\% of the full test pool, and none exceeds 5,000 residues. Its results should therefore be considered alongside test-pool composition and method completion rates.

\paragraph{Comparison with previous benchmarks.}
Table~\ref{tab:benchmark_resolution_comparison} compares resolution distributions using the same bins. The CryoAtlas test pool spans the full reported range, whereas Cryo2StructData and ModelAngelo concentrate on narrower intervals.

\begin{table}[H]
\centering
\footnotesize
\setlength{\tabcolsep}{6pt}
\renewcommand{\arraystretch}{1.12}
\begin{tabular}{@{}cccc@{}}
\toprule
\textbf{Resolution (\AA)} & \textbf{CryoAtlas test} & \textbf{Cryo2StructData} & \textbf{ModelAngelo} \\
\midrule
$\leq2.5$ & 6.2 & 1.1 & 11.3 \\
2.5--3.0 & 24.0 & 3.2 & 27.1 \\
3.0--3.5 & 35.2 & 26.7 & 39.0 \\
3.5--4.0 & 22.1 & 43.9 & 22.6 \\
4.0--4.5 & 10.8 & 24.5 & 0 \\
4.5--5.0 & 1.1 & 0 & 0 \\
5.0--5.5 & 0.4 & 0 & 0 \\
5.5--6.0 & 0.3 & 0.3 & 0 \\
\bottomrule
\end{tabular}
\vspace{2mm}
\caption{Resolution distributions (\%) in the displayed bins for the CryoAtlas test pool and earlier Cryo2StructData and ModelAngelo benchmarks. Percentages are rounded independently.}
\label{tab:benchmark_resolution_comparison}
\end{table}

\subsection{Comparison of curation workflows}
\label{app:curation_comparison}
Table~\ref{tab:curation_workflows} compares the processing steps and outputs of existing resources. ModelAngelo and CryoAtom2 use manual and model-assisted screening~\cite{ModelAngelo_2024,CryoAtom,CryoAtom2}, while Cryo2StructData releases a processed corpus whose alignment issues are examined in Appendix~\ref{sec:c2s_issues}. CryoDataBot provides configurable collection with Q-score/volume-overlap filtering, redundancy reduction, and resampling~\cite{CryoDataBot_2025}. CryoAtlas combines explicit map--structure correction, sequence mapping, and Q-score and 3D unmodeled-density filtering to produce a fixed corpus with recorded curation decisions and a common benchmark.

\begin{table}[H]
\centering
\footnotesize
\setlength{\tabcolsep}{3pt}
\renewcommand{\arraystretch}{1.2}
\begin{tabular}{@{}ccccc@{}}
\toprule
\textbf{Curation/design choice} & \textbf{MA /}\textbf{CA} & \textbf{Cryo2StructData} & \textbf{CryoDataBot} & \textbf{CryoAtlas} \\
\midrule
Map--model misalignment & Manual/model screening & No & VOF screening & Fitmap correction \\
\addlinespace[3pt]
Q-score filtering & No & No & Yes & Yes \\
\addlinespace[3pt]
Unmodeled-density filtering & Manual/model screening & No & VOF & 3D ratio \\
\addlinespace[3pt]
Sequence--structure alignment & No & Yes & No & Yes \\
\addlinespace[3pt]
Per-entry removal reasons & No & No & Partial & Yes \\
\midrule
Fixed ML-ready corpus & No & Yes & Custom generation & Yes \\
\addlinespace[3pt]
Benchmark metrics and outputs & Partial & Partial & No & Yes \\
\bottomrule
\end{tabular}
\vspace{2mm}
\caption{Comparison of reported curation steps and resource outputs. Manual/model-assisted and volume-overlap (VOF) screening detect potentially unsuitable pairs; CryoAtlas additionally corrects map--model alignment. MA: ModelAngelo; CA: CryoAtom2. Cryo2StructData reports sequence--structure alignment; observed alignment errors are discussed in Appendix~\ref{sec:c2s_issues}. ``No'' denotes a step or output not reported in the compared workflow.}
\label{tab:curation_workflows}
\end{table}

\newpage
\section*{NeurIPS Paper Checklist}

\begin{enumerate}

\item {\bf Claims}
    \item[] Question: Do the main claims made in the abstract and introduction accurately reflect the paper's contributions and scope?
    \item[] Answer: \answerYes{} 
    \item[] Justification: The abstract and introduction clearly summarize the paper's contributions, including the curated CryoAtlas dataset, unified benchmark, baseline evaluation, stratified analysis, and controlled retraining study. These claims are supported by the reported dataset statistics, evaluation results, and training experiments. The paper also states key limitations and distinguishes controlled comparisons from external references.

\item {\bf Limitations}
    \item[] Question: Does the paper discuss the limitations of the work performed by the authors?
    \item[] Answer: \answerYes{} 
    \item[] Justification: The paper includes a dedicated limitations section. While the dataset and benchmark are designed to be broad, carefully curated, and substantially more standardized than existing resources, we also discuss remaining scope and evaluation limitations, including the current focus on protein chains, residual biases in composite metrics, the need for more region-aware functional evaluation, and the single-state treatment of heterogeneous cryo-EM densities. We further state caveats for the controlled retraining experiments.

\item {\bf Theory assumptions and proofs}
    \item[] Question: For each theoretical result, does the paper provide the full set of assumptions and a complete (and correct) proof?
    \item[] Answer: \answerNA{} 
    \item[] Justification: The paper does not present theoretical results, theorems, or formal proofs. The mathematical expressions are used only to define benchmark metrics and composite scores, not to establish theoretical guarantees. These definitions are numbered and described in the evaluation framework section and appendix.

    \item {\bf Experimental result reproducibility}
    \item[] Question: Does the paper fully disclose all the information needed to reproduce the main experimental results of the paper to the extent that it affects the main claims and/or conclusions of the paper (regardless of whether the code and data are provided or not)?
    \item[] Answer: \answerYes{} 
    \item[] Justification: The paper discloses the procedures needed to reproduce or verify the main results, including dataset collection and curation, quality filtering, train/test splitting, benchmark metric definitions, composite scores, baseline evaluation, and controlled retraining settings. We also report implementation caveats for the ModelAngelo reimplementation and specify the use of the official GNN checkpoint during inference. The curated dataset, metadata, benchmark outputs, and evaluation protocols are publicly available through the links in the Data and Code Availability paragraph.

\item {\bf Open access to data and code}
    \item[] Question: Does the paper provide open access to the data and code, with sufficient instructions to faithfully reproduce the main experimental results, as described in supplemental material?
    \item[] Answer: \answerYes{} 
    \item[] Justification: The curated dataset, benchmark metadata, baseline and retraining outputs, and release documentation are publicly available on \href{https://huggingface.co/datasets/markfaa/CryoAtlas}{Hugging Face}. Evaluation, CNN training, and sequence-identity analysis code are available in \href{https://github.com/oulling/CryoAtlas-Code}{CryoAtlas-Code}. Training settings and the evaluation protocol are described in the main paper and appendices.

\item {\bf Experimental setting/details}
    \item[] Question: Does the paper specify all the training and test details (e.g., data splits, hyperparameters, how they were chosen, type of optimizer) necessary to understand the results?
    \item[] Answer: \answerYes{} 
    \item[] Justification: The paper reports the necessary experimental details, including data splits, benchmark construction, evaluation metrics, baseline protocols, and controlled retraining settings. Full training details, hyperparameters, checkpoint usage, and implementation caveats are provided in the appendix and supplemental material.

\item {\bf Experiment statistical significance}
    \item[] Question: Does the paper report error bars suitably and correctly defined or other appropriate information about the statistical significance of the experiments?
    \item[] Answer: \answerYes{} 
    \item[] Justification: Benchmark means and target-level sample SDs are reported in Appendix Tables~\ref{tab:main_results_sd} and~\ref{tab:training_utility_sd}; these SDs describe variation across targets. Table~\ref{tab:rebuttal_training_stability} separately reports mean $\pm$ sample SD over three independent CryoAtlas seeds for the first $5{,}000$ steps; the other stability controls are single runs, and the downstream checkpoint comparisons do not estimate seed variability.

\item {\bf Experiments compute resources}
    \item[] Question: For each experiment, does the paper provide sufficient information on the computer resources (type of compute workers, memory, time of execution) needed to reproduce the experiments?
    \item[] Answer: \answerYes{} 
    \item[] Justification: The paper provides compute details for the main controlled training experiments, including GPU type, number of GPUs, memory, batch size, number of steps, and wall-clock time.
    
\item {\bf Code of ethics}
    \item[] Question: Does the research conducted in the paper conform, in every respect, with the NeurIPS Code of Ethics \url{https://neurips.cc/public/EthicsGuidelines}?
    \item[] Answer: \answerYes{} 
    \item[] Justification: The work uses public EMDB/PDB structural data and does not collect human-subject or private personal data. Data provenance, scope, and methodological limitations are described in Section~\ref{sec:dataset}, the Discussion, and Appendix~\ref{app:dataset_composition}.

\item {\bf Broader impacts}
    \item[] Question: Does the paper discuss both potential positive societal impacts and negative societal impacts of the work performed?
    \item[] Answer: \answerYes{} 
    \item[] Justification: The Impact Statement discusses benefits for reproducible structural biology and the risk of propagating errors from incorrect atomic models. It also addresses limitations from curation and benchmark selection, recommends validation against experimental density before downstream use, and considers potential dual-use risks.
    
\item {\bf Safeguards}
    \item[] Question: Does the paper describe safeguards that have been put in place for responsible release of data or models that have a high risk for misuse (e.g., pre-trained language models, image generators, or scraped datasets)?
    \item[] Answer: \answerNA{} 
    \item[] Justification: The released resources are curated cryo-EM structure-modeling data and evaluation artifacts derived from public EMDB/PDB entries, rather than high-risk generative models or scraped datasets containing personal or unsafe content. As such, no special misuse safeguards of the type described in this question are required. We nevertheless provide documentation, evaluation protocols, and limitations to support responsible research use.

\item {\bf Licenses for existing assets}
    \item[] Question: Are the creators or original owners of assets (e.g., code, data, models), used in the paper, properly credited and are the license and terms of use explicitly mentioned and properly respected?
    \item[] Answer: \answerYes{} 
    \item[] Justification: We cite and document all existing datasets, tools, models, checkpoints, and evaluation software used in the paper, including their sources, access dates or versions where applicable, and license or terms-of-use information. The CryoAtlas dataset and original code are licensed under CC BY-NC 4.0; third-party assets retain their original licenses and terms of use.

\item {\bf New assets}
    \item[] Question: Are new assets introduced in the paper well documented and is the documentation provided alongside the assets?
    \item[] Answer: \answerYes{} 
    \item[] Justification: The paper introduces new assets, including the CryoAtlas dataset, curated train/test splits, benchmark metadata, evaluation protocols, and baseline outputs. These assets are documented alongside the release, including dataset construction procedures, file organization, preprocessing and alignment steps, quality-control criteria, metric definitions, intended use, limitations, and licensing. The dataset and original CryoAtlas code are released under CC BY-NC 4.0, with documentation and access links provided on \href{https://huggingface.co/datasets/markfaa/CryoAtlas}{Hugging Face} and \href{https://github.com/oulling/CryoAtlas-Code}{GitHub}.

\item {\bf Crowdsourcing and research with human subjects}
    \item[] Question: For crowdsourcing experiments and research with human subjects, does the paper include the full text of instructions given to participants and screenshots, if applicable, as well as details about compensation (if any)? 
    \item[] Answer: \answerNA{} 
    \item[]  Justification: The paper does not involve crowdsourcing experiments or research with human subjects. The dataset is constructed from publicly available structural biology resources, and no participant instructions, screenshots, or compensation details are applicable.

\item {\bf Institutional review board (IRB) approvals or equivalent for research with human subjects}
    \item[] Question: Does the paper describe potential risks incurred by study participants, whether such risks were disclosed to the subjects, and whether Institutional Review Board (IRB) approvals (or an equivalent approval/review based on the requirements of your country or institution) were obtained?
    \item[] Answer: \answerNA{} 
    \item[] Justification: The paper does not involve crowdsourcing or research with human subjects. It uses publicly available structural biology resources and does not collect data from study participants; therefore, IRB approval or equivalent human-subjects review is not applicable.

\item {\bf Declaration of LLM usage}
    \item[] Question: Does the paper describe the usage of LLMs if it is an important, original, or non-standard component of the core methods in this research? Note that if the LLM is used only for writing, editing, or formatting purposes and does \emph{not} impact the core methodology, scientific rigor, or originality of the research, declaration is not required.
    \item[] Answer: \answerNA{} 
    \item[] Justification: The core methodology of this work does not use LLMs as an important, original, or non-standard component. The paper focuses on cryo-EM dataset curation, benchmarking, and empirical evaluation; any LLM use was limited to writing, editing, or formatting assistance and did not affect the scientific methodology, experiments, or conclusions.

\end{enumerate}

\end{document}